\documentclass[journal]{IEEEtran}
\ifCLASSINFOpdf
\else
\fi
\usepackage{rotating}
\usepackage{graphics}
\usepackage{makecell}
\usepackage{tabularx}
\usepackage{multirow} 
\usepackage{subfigure}
\usepackage{amssymb}
\usepackage{amsmath}
\usepackage{array}
\usepackage{wasysym}
\usepackage{threeparttable}
\usepackage{arydshln}
\usepackage{hyperref}
\usepackage{xcolor}

\newtheorem{myDef}{\textbf{Definition}}

\begin{document}
%
\title{Zero-Knowledge Proof-based Verifiable Decentralized Machine Learning in Communication Network: A Comprehensive Survey}
%
%
%

\author{Zhibo~Xing,
        Zijian~Zhang,~\IEEEmembership{Member,~IEEE,}
        Ziang~Zhang,
        Zhen~Li,
        Meng~Li,~\IEEEmembership{Senior Member,~IEEE,}
        Jiamou~Liu,
        Zongyang~Zhang,
        Yi~Zhao,
        Qi~Sun,
        Liehuang~Zhu,~\IEEEmembership{Senior Member,~IEEE,}
        Giovanni~Russello,~\IEEEmembership{Member,~IEEE.}
\thanks{Zhibo Xing is with the School of Cyberspace Science and Technology, Beijing Institute of Technology, Beijing, 100081, China, and the School of Computer Science, The University of Auckland, Auckland, 1010, New Zealand. E-mail: 3120215670@bit.edu.cn.}
\thanks{Zijian Zhang is with the School of Cyberspace Science and Technology, Beijing Institute of Technology, Beijing, 100081, China, and Southeast Institute of Information Technology, Beijing Institute of Technology, Fujian, 351100, China. E-mail: zhangzijian@bit.edu.cn.}%
\thanks{Jiamou Liu and Giovanni Russello are with the School of Computer Science, The University of Auckland, Auckland, 1010, New Zealand. Email: \{jiamou.liu, g.russello\}@auckland.ac.nz.}%
\thanks{Ziang Zhang, Zhen Li, Yi Zhao, Liehuang Zhu are with the School of Cyberspace Science and Technology, Beijing Institute of Technology, Beijing, 100081, China. E-mail: \{3220231794, zhen.li, zhaoyi, liehuangz\}@bit.edu.cn.}%
\thanks{Meng Li is with Key Laboratory of Knowledge Engineering with Big Data (Hefei University of Technology), Ministry of Education; School of Computer Science and Information Engineering, Hefei University of Technology, Anhui, 230601, China; Anhui Province Key Laboratory of Industry Safety and Emergency Technology; and Intelligent Interconnected Systems Laboratory of Anhui Province (Hefei University of Technology). Email: mengli@hfut.edu.cn.}%
\thanks{Zongyang Zhang is with the School of Cyber Science and Technology, Beihang University, Beijing, 100191, China. E-mail: zongyangzhang@buaa.edu.cn.}%
\thanks{Qi Sun is with the Department of Bioinformatics, Hangzhou Nuowei Information Technology Co.,Ltd, Zhejiang, 310053, China. E-mail: sunq0810@gmail.com.}%
\thanks{\textit{(Corresponding Author: Zijian Zhang, Meng Li.)}}}%

%
%

\markboth{Journal of \LaTeX\ Class Files,~Vol.~14, No.~8, August~2015}%
{Shell \MakeLowercase{\textit{et al.}}: Bare Demo of IEEEtran.cls for IEEE Journals}
%



\maketitle

\begin{abstract}


\textcolor{black}{Over recent decades, machine learning has significantly advanced network communication, enabling improved decision-making, user behavior analysis, and fault detection. 
Simultaneously, the growth of communication networks has facilitated the efficient collection of large-scale training data. 
Traditional centralized machine learning, however, requires data collection from users, raising significant concerns about privacy and security. 
Decentralized approaches, where participants exchange computation results instead of raw private data, mitigate these risks but introduce challenges related to trust and verifiability. 
A critical issue arises: \textit{How can one ensure the integrity and validity of computation results shared by other participants? }
Existing survey articles predominantly address security and privacy concerns in decentralized machine learning, whereas this survey uniquely highlights the emerging issue of \textit{verifiability}. 
Recognizing the critical role of zero-knowledge proofs in ensuring verifiability, we present a comprehensive review of Zero-Knowledge Proof-based Verifiable Machine Learning (ZKP-VML). 
To clarify the research problem, we present a definition of ZKP-VML consisting of four algorithms, along with several corresponding key security properties. 
Besides, we provide an overview of the current research landscape by systematically organizing the research timeline and categorizing existing schemes based on their security properties. 
Furthermore, through an in-depth analysis of each existing scheme, we summarize their technical contributions and optimization strategies, aiming to uncover common design principles underlying ZKP-VML schemes.
Building on the reviews and analysis presented, we identify current research challenges and suggest future research directions.
To the best of our knowledge, this is the most comprehensive survey to date on verifiable decentralized machine learning and ZKP-VML. }

\end{abstract}


%
\IEEEpeerreviewmaketitle

\section{Introduction\label{sec1}}

\IEEEPARstart{I}n recent years, artificial intelligence (AI) has seen widespread adoption in our daily life, receiving significant attention from both academia and industry. 
The complexity of problems that AI can tackle grows in tandem with advancements in machine learning (ML) methodologies. 
However, the development of ML also increases the communication and computation overheads, as larger ML models are required to be trained on larger datasets.
On the one hand, the exponentially growing demand for computing power and data scale in machine learning tasks making it difficult for personal computers to accomplish machine learning tasks~\cite{canziani2016analysis,li2016evaluating}.
On the other hand, concerns about data privacy increases with time, which damages people's enthusiasm in sharing their data and participating in machine learning. 
The demand for computing power and concerns on data privacy restrain the application of centralized machine learning (CML) while encourage the development of decentralized machine learning (DML). 
DML allows participants to accomplish the ML tasks with multiple rounds of local computation and global communication, instead of simply gathering needed data without privacy guarantee. 
A series of secure computation protocols and DML frameworks have been proposed to safeguard the data privacy. 
Federated learning (FL)~\cite{mcmahan2017communication} is a popular DML framework, which allows several participants to collaboratively train a ML model without sharing their own private datasets, thus achieving data privacy protection. 
These works are dedicated to perform ML tasks while protecting data privacy and minimizing computational and communication burdens. 
Although the DML efficiently solves the data privacy concerns, the verifiability issue is still highly neglected. 

The concern over \textit{verifiability} is rooted in DML framework, as we have no access to the private data and computation process of others. 
A key focus will be on how requesters can verify that the computation results provided by performers are accurate through faithful local computation. 
For example, in ML training tasks, a malicious performer may implant backdoors into the model via poisoning attacks~\cite{sun2021data,tian2022comprehensive}, leading to the misclassification. 
In ML inference tasks, a malicious performer may also give a false prediction, misguiding the requester to make wrong decisions. 
These security issues and malicious attacks derive from the difficulty in verifying the correctness of the computation process and results. 

Intuitively, the computation performer may send all the input data to the requester for the re-execution and verification, while this is not feasible in practice yet. 
This verification method would severely damage the basic data privacy that DML pursue.
Notably, sending all the input data and re-executing the ML computation would put an unbearable workload on communication and computation. 
Thus, the problem is, \textit{how can we verify the results submitted by others without access to their private data while maintaining low communication and computational overhead.}
Fortunately, zero-knowledge proof provides a possible solution for this tricky problem. 

Zero-knowledge proof (ZKP) is a powerful cryptographic technique for addressing the verifiability issues in DML, especially when it comes to the privacy and low overhead. 
ZKPs allow one party to prove the correctness of a statement without revealing additional information to another party. 
Within DML, ZKP aligns well with the need of verifying the correctness and integrity of local computation results.
By presenting the corresponding DML algorithm and the inputs and outputs as a statement that “The output is honestly computed with the given algorithm, given public inputs and some specific private inputs”, the performer can generate a proof, arguing for the correctness of the computational process and results, without revealing any additional information about the private inputs. 
This proof can be verified by other participants, and once the verification passes, they can trust the statement without accessing to additional details, such as private inputs to the algorithm.

\begin{figure*}
    \centering
    \subfigure[Artificial intelligence diagnosis based on case privacy data]{
        \includegraphics[width=0.4\textwidth]{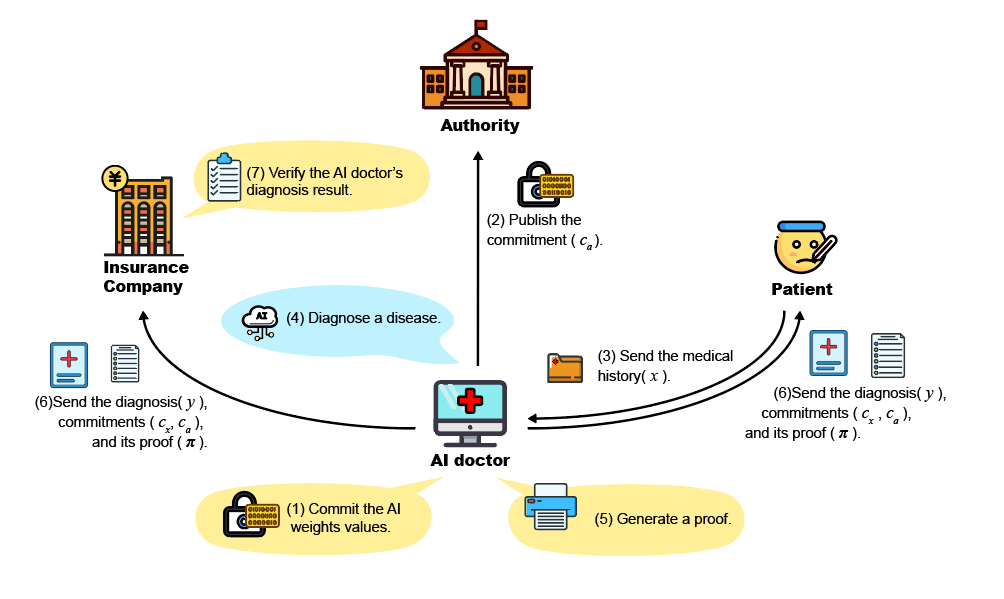}
        \label{hospital}
    }
    \subfigure[Federated training of anti-money laundering models based on private transaction behavior]{
        \includegraphics[width=0.4\textwidth]{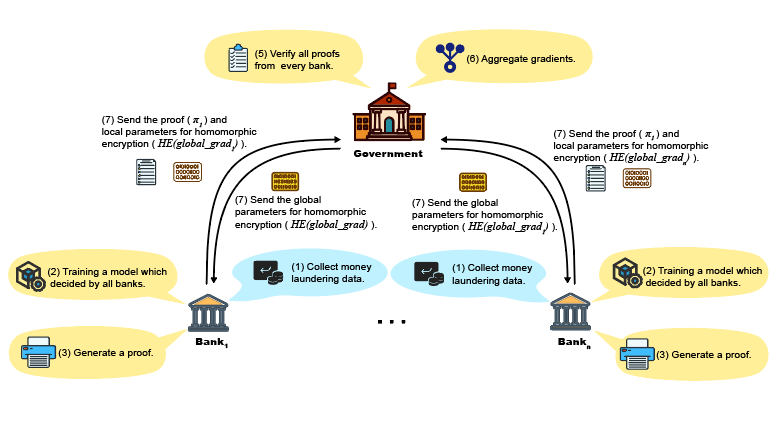}
        \label{bank}
    }
    \caption{Real-life applications of zero-knowledge proof-based verifiable machine learning}
    \label{real-life application}
\end{figure*}

\textcolor{black}{Considering that verifiable machine learning is a relatively new research area, in order to provide a clearer understanding at how zero-knowledge proofs ensuring the verifiability in decentralized machine learning and why this is important and necessary, we describe the following applications of ZKP-VML in real-life scenarios as examples in Fig.~\ref{real-life application}. }

\textcolor{black}{
We illustrate outsourced inference in an AI-assisted diagnosis scenario. Machine learning enables hospitals to diagnose diseases more efficiently and accurately. However, hospitals must provide proofs of the inference process to ensure verifiability. Such proofs allow patients to verify the correctness of their diagnoses and enable insurance companies to process claims accordingly. Moreover, both the diagnostic model and patient data are sensitive—hospitals must protect their proprietary models, while patient privacy must be preserved. Since the proof is shared with both the patient and the insurer, it should reveal no information beyond the diagnosis result. Hospitals can generate ZKPs for the inference process, allowing verification without exposing the model or input data. Additionally, commitment schemes can be integrated with ZKPs to guarantee the integrity of the model and medical data used in inference.
The whole process is shown in Fig.~\ref{hospital}.}

\textcolor{black}{
The second example concerns federated learning in training models for money laundering detection. By analyzing user information and transaction records, AI-driven machine learning models can identify potential financial crimes more efficiently. Governments can facilitate federated learning among multiple banks to collaboratively train a global model while preserving data privacy. Federated learning ensures that each bank's data remains confidential and is not exposed to other banks or the government. To guarantee the integrity of local training, banks must provide proofs of correctness. ZKPs enable each bank to demonstrate the validity of its training process without revealing sensitive data. This allows the government to verify training correctness while maintaining privacy.
The whole process is shown in Fig.~\ref{bank}.}

\subsection{Related Work}

\textcolor{black}{
In related work, we mainly compare surveys focusing on machine learning verifiability, in particular achieved through zero-knowledge proofs. 
Considering that verifiability is a key component of security in machine learning, we further cover some surveys related to security and privacy as related work to provide a more comprehensive comparison. }
\textcolor{black}{We categorize existing work into three progressive layers: machine learning security and privacy, machine learning verifiability, and ZKP-based machine learning verifiability. Table \ref{relatedtable} analyzes and compares related work based on the above categorization. }

\begin{table*}[t]
\centering
\caption{\textcolor{black}{Comparison with related work.}}
\label{relatedtable}
\begin{threeparttable}
\begin{tabular}{cc cc ccccc}
\multicolumn{2}{c}{\multirow{2}{*}{\thead{Related\\Survey}}} & \multicolumn{2}{c}{\makecell{Secure ML \\ \rule{2cm}{0.4pt}}} & \multicolumn{5}{c}{\makecell{\textbf{Zero-Knowledge Proof-based} Verifiable ML \\ \rule{9cm}{0.4pt}}} \\
& & Privacy & Verifiability & \thead{Scheme\\Coverage\\\textbf{of ZKP-VML}} & \thead{Technical\\Analysis\\\textbf{of ZKP-VML}} & \thead{Scheme\\Categorization\\\textbf{of ZKP-VML}} & \thead{Future Research\\Directions\\\textbf{of ZKP-VML}} & \thead{Problem\\Definition\\\textbf{of ZKP-VML}} \\ \hline
\multirow{4}{*}{\thead{Security\\and Privacy}} &\thead{Ma et al.~\cite{ma2021survey}} & $\CIRCLE$ & $\LEFTcircle$ & $\Circle$(0) & $\Circle$ & $\Circle$ & $\Circle$ & $\Circle$  \\ 
&\textcolor{black}{\thead{Qin et al.~\cite{qin2023cryptographic}}} & \textcolor{black}{$\CIRCLE$} & \textcolor{black}{$\LEFTcircle$} & \textcolor{black}{$\LEFTcircle$(5)} & \textcolor{black}{$\Circle$} & \textcolor{black}{$\LEFTcircle$} & \textcolor{black}{$\Circle$} & \textcolor{black}{$\Circle$}  \\ 
&\textcolor{black}{\thead{Chen et al.~\cite{chen2024federated}}} & \textcolor{black}{$\CIRCLE$} & \textcolor{black}{$\LEFTcircle$} & \textcolor{black}{$\LEFTcircle$(2)} & \textcolor{black}{$\LEFTcircle$} & \textcolor{black}{$\Circle$} & \textcolor{black}{$\Circle$} & \textcolor{black}{$\Circle$}  \\ 
&\thead{Hallaji et al.~\cite{hallaji2024decentralized}} & $\CIRCLE$ & $\LEFTcircle$ & $\LEFTcircle$(1) & $\Circle$ & $\Circle$ & $\Circle$ & $\Circle$ \\ \hdashline

\multirow{2}{*}{\thead{Verifiability}} &\thead{Zhang et al.~\cite{zhang2022towards}} & $\CIRCLE$ & $\CIRCLE$ & $\LEFTcircle$(1) & $\Circle$ & $\Circle$ & $\Circle$ & $\Circle$  \\ 
&\thead{Tariq et al.~\cite{tariq2024trustworthy}} & $\CIRCLE$ & $\CIRCLE$ & $\LEFTcircle$(2) & $\Circle$ & $\Circle$ & $\Circle$ & $\Circle$ \\ \hdashline

\multirow{4}{*}{\thead{ZKP-based\\Verifiability}} &\thead{Modulus Lab~\cite{modulus2023cost}} & $\CIRCLE$ & $\CIRCLE$ & $\LEFTcircle$(9) & $\Circle$ & $\LEFTcircle$ & $\LEFTcircle$ & $\Circle$ \\ 
&\thead{Sathe et al.~\cite{sathe2023state}} & $\CIRCLE$ & $\CIRCLE$ & $\LEFTcircle$(5) & $\Circle$ & $\Circle$ & $\Circle$ & $\Circle$ \\
&\thead{Zhang et al.~\cite{zhang2024research}} & $\CIRCLE$ & $\CIRCLE$ & $\Circle$(0) & $\LEFTcircle$ & $\Circle$ & $\LEFTcircle$ & $\Circle$ \\ 
&\textcolor{black}{\thead{Kersic et al.~\cite{kervsivc2024chain}}} & \textcolor{black}{$\CIRCLE$} & \textcolor{black}{$\CIRCLE$} & \textcolor{black}{$\Circle$(0)} & \textcolor{black}{$\LEFTcircle$} & \textcolor{black}{$\Circle$} & \textcolor{black}{$\LEFTcircle$} & \textcolor{black}{$\LEFTcircle$} \\ \hline \hline
&\thead{Our Work} & $\CIRCLE$ & $\CIRCLE$ & $\CIRCLE$(53+) & $\CIRCLE$ & $\CIRCLE$ & $\CIRCLE$ & $\CIRCLE$\\ 
\hline
\end{tabular}
\begin{tablenotes}
    \footnotesize
    \item[1] $\Circle$ denotes the requirement has not been met; 
    \item[2] $\LEFTcircle$ denotes the requirement is partially met; 
    \item[3] $\CIRCLE$ denotes the requirement is fully met;
\end{tablenotes}
\end{threeparttable}
\end{table*}

\subsubsection{Survey on Secure Machine Learning}

\textcolor{black}{
Several studies focus on the security issues of decentralized machine learning while also considering verifiability. For example, Ma et al.~\cite{ma2021survey} explore the security challenges of outsourced deep learning, presenting a system model and security requirements. They provide a comprehensive analysis of techniques for secure outsourcing, including a detailed taxonomy of privacy-preserving strategies for training and inference. The work also offers comparative analyses of cryptographic methods such as homomorphic encryption (HE) and secret sharing (SS), while examining trade-offs between privacy, verifiability, efficiency, and non-interactivity. However, their focus is mainly on privacy, with limited attention to verifiability. The discussion on verifiability remains conceptual, lacking empirical evaluations or practical implementations. Additionally, the key cryptographic technique, ZKPs, is not addressed.
}
\textcolor{black}{
Qin et al.~\cite{qin2023cryptographic} systematically examine cryptographic methods used in privacy-preserving machine learning (PPML). They classify existing PPML protocols based on cryptographic primitives, including secure multiparty computation (MPC), ZKPs, HE, and differential privacy (DP). Emerging trends, such as quantum-resistant cryptographic methods and efficient key-sharing for multiparty learning, are also highlighted. The discussion on verifiability primarily focuses on ZKPs and trusted execution environments (TEEs), outlining their theoretical frameworks and potential applications in PPML. However, the paper lacks a detailed exploration of practical mechanisms for ensuring verifiable correctness in complex ML models, particularly in decentralized settings. This leaves a gap in addressing real-world challenges like efficient proof generation and deployment, limiting the survey's practical relevance for verifiability.
}
\textcolor{black}{
Chen et al.~\cite{chen2024federated} offers a comprehensive analysis of privacy-preserving computation techniques in federated learning (FL), categorizing protocols based on cryptographic methods such as MPC, HE, DP, TEEs, and ZKPs. The paper provides detailed instantiation processes for each protocol and compares them in terms of security and computational efficiency. While it addresses verifiability, the focus is primarily on the global model aggregation process in FL, neglecting the more computationally intensive training process in DML. This leaves a gap in fully verifiable machine learning.
}
\textcolor{black}{
Hallaji et al.~\cite{hallaji2024decentralized} examine security and privacy challenges in decentralized federated learning (DFL) systems, particularly those using blockchain technology. The paper reviews state-of-the-art DFL methods, identifies threats to privacy and performance, and evaluates defense mechanisms, including DP, HE, and MPC. It also addresses DFL verifiability by introducing frameworks with trusted trainers and workers to ensure system reliability. However, the analysis of verifiability is limited, as the discussed methods are neither comprehensive nor sufficient, and the role of ZKPs in ensuring verifiability is overlooked.
}

\textcolor{black}{
These works offer comprehensive reviews of security and privacy issues across various machine learning scenarios. However, their analysis of verifiability is limited. Verifiability is a critical aspect of security, as it enhances trustworthiness by enabling the verification of transparent computations in decentralized machine learning. As a result, their discussions of security are not fully comprehensive.
}

\subsubsection{Survey on Verifiable Machine Learning}

\textcolor{black}{Some studies focus on trust and verifiability issues in decentralized machine learning. }
\textcolor{black}{
Zhang et al.~\cite{zhang2022towards} focus on verifiable FL, defining it as the ability for one party to prove to others in an FL protocol that it has correctly performed the intended task. They propose a novel taxonomy for verifiable FL, categorizing verifiability based on centralized and decentralized FL settings. In each setting, verifiability is further divided by participant security needs, covering model aggregation, local information, and model updates. Techniques such as TEEs, reputation mechanisms, and contract theory are discussed. However, the analysis is limited by a narrow view of attacker behavior, leaving the verifiability framework incomplete. Notably, little attention is given to verifying the correctness of the local training process, and the role of ZKPs in verifiability is underexplored.
}
\textcolor{black}{
Tariq et al.~\cite{tariq2024trustworthy} provide a comprehensive exploration of trustworthy FL, addressing its principles, challenges, and future directions. They introduce a novel taxonomy with five key pillars: interpretability \& explainability, transparency, privacy \& robustness, fairness, and accountability. Each pillar is further divided into subcategories, offering a multidimensional perspective on trust. A Trustworthy FL architecture is proposed, incorporating secure aggregation, incentive mechanisms, and verifiability at both the client and server levels. While the paper emphasizes communication efficiency, it offers limited discussion on verifiability. Though acknowledging its importance, the work does not delve into the technical challenges of verifying the integrity of the local training process, the most computationally intensive part of FL. Additionally, the potential of cryptographic methods, such as ZKPs, to enhance verifiability is underexplored.
}

\textcolor{black}{
These surveys recognize the importance of verifiability in ML but focus less on verifying the model training or inference processes, prioritizing other computational aspects or alternative methods. As a result, their discussion of verifiability is insufficiently robust. Additionally, while ZKPs are crucial for ensuring verifiability, they are rarely addressed in these works, limiting the coverage of existing research.
}

 \subsubsection{Survey on ZKP-Based Verifiable Machine Learning}

\textcolor{black}{Some reviews have recognized the importance of zero-knowledge proofs (ZKPs) in addressing verifiability issues in machine learning. }
\textcolor{black}{
Modulus Labs' white paper~\cite{modulus2023cost} explores the intersection of ZKPs and ML inference, benchmarking six ZKP systems across various models, especially multilayer perceptrons. It compares proof generation time and memory usage, highlighting trade-offs like Plonky2's speed versus high memory consumption, and zkCNN's efficiency for large models. The paper emphasizes potential in ZKPs to enable secure AI inference on decentralized systems while showcasing real-world use cases. While the paper effectively benchmarks ZKP systems in technical performance, its analysis of verifiability lacks a broader exploration of existing work. It focuses more on the ZKP systems themselves, with only one ZKP-VML scheme discussed, leaving out key aspects of ZKP-VML design. Moreover, it overlooks scalability concerns related to high computational and memory costs, particularly for low-power devices. This omission limits the development of efficient ZKP-VML schemes, constraining the adoption of ZKP-based solutions in lightweight environments.
}
\textcolor{black}{
Sathe et al.~\cite{sathe2023state} review the integration of ZKPs into ML to address privacy and security concerns, highlighting their role in enabling verifiable computations without compromising data privacy. The paper discusses five ZKP-based machine learning schemes: vCNN~\cite{lee2024vcnn}, zkCNN~\cite{liu2021zkcnn}, ZEN~\cite{feng2021zen}, ZKP-FL~\cite{xing2023zero}, and Mystique~\cite{weng2021mystique}. However, the analysis is limited, covering only a narrow range of schemes without a systematic comparison or categorization. Additionally, the discussion lacks an in-depth exploration of current research priorities and future directions, limiting the paper's overall contribution. 
}
\textcolor{black}{
Zhang et al.~\cite{zhang2024research} explore how ZKPs can enhance privacy, security, and data integrity in machine learning by enabling privacy-preserving data sharing and secure multi-party computation. However, the paper does not cover or analyze any existing ZKP-ML schemes, weakening its classification of ZKP-ML applications. The analysis of future research directions lacks grounding in prior work, and the brief length of the paper constrains its contribution.
}
\textcolor{black}{
Kersic et al.~\cite{kervsivc2024chain} review the integration of ZKPs into ML for on-chain decentralized applications. The paper examines two on-chain ZKP-ML frameworks, EZKL and Orion, focusing on their ZKP systems, scalability, and privacy features, as well as exploring practical use cases like DAO treasury management. However, the review emphasizes on-chain applications rather than deeply analyzing the ZKP-ML schemes themselves. Given the emerging nature of on-chain ZKP-ML, the analysis is limited, covering only two schemes, neither of which has formal publications.
}

\textcolor{black}{While these surveys review ZKP-VML as a separate issue, they generally suffer from significant limitations in both the breadth of coverage and the depth of analysis of ZKP-VML schemes. This not only leaves room for improvement in classifying existing research and identifying future development directions, but also, to some extent, underestimates the contributions of current studies in the field.
The existing work covers no more than 10 different ZKP-VML schemes in total, yet in fact there are more than 50 ZKP-VML schemes. The limited coverage of existing work leads to the insufficient categorization and analysis of ZKP-VML issue. }

To summarize, Table \ref{relatedtable} shows that existing work has not yet provided a broad and in-depth review of the problem of \textbf{zero-knowledge proof-based verifiable machine learning}. 
On the one hand, there are only a few works that review the verifiability issues in machine learning, and even fewer of them consider zero-knowledge proofs. 
On the other hand, existing work does not provide a comprehensive and in-depth review of the ZKP-VML, lacking summary, coverage, analysis, and categorization of existing schemes.
Meanwhile, as for a relatively new research area, these existing works are limited in summarizing the problems that still exist and the future research directions, making it difficult to provide assistance to subsequent researchers.
Our work not only overcomes the above shortcomings, but also provides the definition of algorithms and properties to ZKP-VML, further paving the way for subsequent researchers. 

\subsection{\textcolor{black}{Motivation}}

\textcolor{black}{Decentralized machine learning protects data privacy, but the transparent local computation process introduces additional security challenges. 
Considering the potential active attacks during transparent computation, such as poisoning attacks and free-rider attacks, ensuring the verifiability of local computations is critical. 
Numerous reviews and surveys have explored security and privacy issues in decentralized machine learning, making valuable contributions to addressing various gaps in the literature.
However, only a small fraction specifically address verifiability concerns or treat verifiability as a distinct issue. 
In parallel, various protocols and solutions have been proposed to tackle verifiability challenges in decentralized machine learning. 
Notably, most existing solutions leverage zero-knowledge proofs as a fundamental technique, yet this topic has been relatively overlooked in existing surveys that include verifiability. 
Meanwhile, existing surveys on zero-knowledge proof-based verifiable machine learning lack both depth of analysis and breadth of scheme coverage, leading to an absence of a comprehensive review in this area. 
To bridge the gap between the limitations of existing surveys and the breadth of research outcomes in ZKP-VML, we present this survey. 
This survey aims to emphasize the importance of verifiability in machine learning and the role of zero-knowledge proofs as a foundational cryptographic primitive. 
Specifically, we seek to provide a more systematic and comprehensive analysis, along with a detailed categorization, of the existing research, thereby addressing the current gap in surveys and reviews within this field.
}

\subsection{\textcolor{black}{Contributions}}

\textcolor{black}{
In this paper, we first present the definition of ZKP-VML algorithms and their corresponding properties, aiming not only to address gaps in the problem's definition but also to establish evaluation criteria for existing work and provide a reference for future research. 
Based on the definition and the basic process of ZKP-VML, we analyze the challenges and difficulties it faces. 
In analyzing the existing schemes, we first outline the timeline of ZKP-VML research development to provide a clear overview of the current state of research. 
Then we list and compare the existing schemes based on their security properties, as described previously. 
We identify three main technical route and some sub-routes, each main route addressing a specific research challenge faced by ZKP-VML as mentioned before. 
Each scheme is categorized according to its technical route, accompanied by an in-depth analysis of the scheme itself and how it addresses the corresponding research challenge. 
Furthermore, considering the computational cost as one of the major disadvantages of ZKP-VML, we also analyze how existing schemes optimize the additional computational and communication burdens. 
The existing optimization methods can be broadly classified into two main categories and several sub-categories. 
Finally, based on the analysis and categorization of existing ZKP-VML schemes, we identify the remaining research challenges and propose several promising directions for future research.
The main contributions are listed as follows:
}

\begin{enumerate} 

    \item \textcolor{black}{\textbf{Bring ZKP-VML to the stage.} 
    To the best of our knowledge, this is the first comprehensive review of ZKP-VML, accompanied by formal definitions and security properties. 
    }
    
    \item \textcolor{black}{\textbf{Cover 56 existing ZKP-VML schemes.} 
    We cover nearly all existing ZKP-VML schemes from the inception in 2017 to June 2024, outlining the development timeline of ZKP-VML. For each existing ZKP-VML scheme we provide a detailed analysis.
    }
    
    \item \textcolor{black}{\textbf{Provide two classifications from different perspectives.} 
    We classify existing ZKP-VML schemes from two novel perspectives: technical routes and optimization methods. Additionally, we provide a detailed analysis for each commonly employed technique and optimization in existing scheme. 
    }
    
    \item \textcolor{black}{\textbf{Present challenges and future directions.} 
    Based on the analysis of existing schemes, we summarize the current challenges faced by ZKP-VML and propose potential directions for future research, offering guidance for future investigations in this field.
    }
    
\end{enumerate}

\subsection{Scope of This Survey}


Section~\ref{sec2} illustrates the background knowledge of machine learning and zero-knowledge proof. 
\textcolor{black}{
Section~\ref{sec3} defines the zero-knowledge proof-based verifiable machine learning (ZKP-VML).
Section~\ref{sec4} provides the overview of existing ZKP-VML schemes. 
Section~\ref{sec5} analyzes the key technical route on scheme building.
Section~\ref{sec6} summarizes the optimization methods employed by existing schemes. 
}
Section~\ref{sec7} presents challenges and future directions.
Finally, Section~\ref{con} concludes the survey.

\section{Background\label{sec2}}

In this section, we provide background knowledge on decentralized machine learning and zero-knowledge proof, outlining how zero-knowledge proof addresses the security issues in decentralized machine learning.

\subsection{Decentralized machine learning}



\textcolor{black}{
Decentralized machine learning (DML) mitigates resource constraints and lowers participation barriers by distributing computational workloads. Traditionally, DML involves multiple workers sharing computation. We extend this definition to include single-worker scenarios, such as outsourced machine learning, due to their similar security concerns. The distribution of computation introduces additional security challenges, including increased participant interactions and potential malicious participants. To systematically analyze these security issues, we classify DML into single-worker and multi-worker paradigms based on workload distribution.
}
We will then discuss their workflow and the security issues they face respectively. 

\subsubsection{\textcolor{black}{Single-Worker DML}}

\textcolor{black}{
Single-Worker DML refers to a paradigm in which the computational workload of a machine learning task is handled by a single computing entity. Although the computation is solely performed by the central entity, the task itself is delegated by requesters who lack the capability to independently complete the machine learning process. Therefore, despite its centralized nature, this approach is still considered a form of DML. Common single-worker DML includes outsourced model training and inference serving.
}

\textcolor{black}{Due to the limited computing power and storage capacity, requesters can outsource the model training task to a performer, allowing them to perform the computations~\cite{duc2019machine}. 
This represents the fundamental concept of outsourced model training. }
The requester sends all the necessary data, including the model parameter $W$, the training dataset $D=(X,Y)$, and the hyper-parameters $\eta$ to the server for the training task. 
The performer then performs the training task $W'\leftarrow W - \eta \nabla \mathcal{L}(f(X,W), Y)$ according to the loss function $\mathcal{L}$ and returns the trained parameters $W'$. 
However, verifiability remains a challenge. To save computational resources, dishonest performers may return manipulated results instead of performing computations honestly. The requester must verify that the received model parameters are correctly computed, especially when lacking sufficient local resources. This challenge intensifies when the requester depends on the server’s private data for training, requiring verification without direct access to the complete input. For example, the performer may hold a private dataset used for model optimization or possess a pre-trained private model that the requester seeks to personalize with its own data.

\textcolor{black}{
With the rapid advancement of large language models, inference serving has become a widely used machine learning application. Unlike traditional settings where training and inference occur on the same device, inference serving offloads inference computation to a remote performer. Instead of running a model locally, a requester submits a query to a performer, which processes it using a trained model and returns the result. In this process, the requester seeks to obtain the inference result $\hat{Y} \leftarrow f(X, W)$ for input $X$ on model $f$ with parameters $W$. Verifiability is a key concern, as requesters must ensure the returned results are computed correctly using the specified input and model. However, dishonest performers may return arbitrary results or bypass computation to save resources. Additionally, verification mechanisms must preserve model privacy, preventing malicious requesters from reconstructing the model from the proofs.
}


\subsubsection{\textcolor{black}{Multi-Worker DML}}

\textcolor{black}{
Multi-worker DML distributes the computational workload of a machine learning task across multiple workers, enhancing efficiency and reducing single points of failure. However, effective task allocation is crucial for seamless collaboration. Common approaches include federated learning and crowdsourced inference.
}


Federated learning (FL)~\cite{mcmahan2017communication} enables multiple clients to collaboratively train a global model while preserving data privacy. Each client contributes to model training without sharing its private dataset. We focus on horizontal FL, where data samples are partitioned across participants, to illustrate the general process. 
A typical FL system consists of a central server $S$ and $n$ clients $C_1, ..., C_n$, each with a private dataset $D_u$. Training proceeds in multiple rounds until the global model converges. In the $j$-th training round, the server distributes the global model $M_{j-1}$ to clients. Each client $C_u$ trains the model on its dataset $D_u$ to obtain the local model $m^j_u$. Clients submit their local model to the server, which aggregates them to obtain the global model $M^j = M^{j-1} + \frac{gl}{n}\sum_{u=1}^n (M^{j-1}-m^j_u)$, where $gl$ is the global learning rate. 
Dishonest clients may submit incorrect local models to evade computation or conduct poisoning attacks. The server may also return wrong global model as the aggregation result to fool clients, raising concerns about trustworthiness~\cite{mothukuri2021survey}.
Despite its privacy-preserving design, FL remains vulnerable to privacy attacks. A curious server may reconstruct client datasets via data reconstruction attacks or infer training participation using membership inference attacks~\cite{mothukuri2021survey}. 
\textcolor{black}{
Crowdsourcing inference involves multiple independent workers collaboratively performing machine learning inference tasks. These workers, operating within a distributed network, submit local inference results or partial computations to a central server, which aggregates them to generate the final output towards solving a larger problem. 
Each worker $C_u$ generate the inference result $\hat{Y}_u \leftarrow f(X_u, W_u)$ based on their given data $X_u$ and parameter $W_u$. 
$\alpha_u$ is the trust level of worker $C_u$, which is used for generate a more reliable final result $\hat{Y} \leftarrow \sum_{u=1}^n \alpha_u \hat{Y}_u$. 
While crowdsourcing inference shares a similar architectural with federated learning, it faces additional verifiability challenges beyond the local computation and global aggregation. On one hand, the heterogeneity of workers increases the complexity for the server to verify the correctness of each result. On the other hand, the limited access the server has to the workers' local models further increases the difficulty of performing accurate verification. 
}



\subsubsection{Common Models}

\textcolor{black}{
Different ML models involve distinct computational processes, which influence their adaptation to ZKP-VML. We introduce several widely used models, emphasizing those of particular relevance to ZKP-VML. Our discussion focuses on their computational characteristics, their role in DML, and their implications for verifiable computation. Since ZKP-VML prioritizes computational complexity and workload types, we categorize models accordingly. 
}

For traditional models with low computational complexity, we consider linear regression and decision tree, which feature simple structures and computations.

\textcolor{black}{
\textbf{Linear regression (LR)} is a fundamental ML model that establishes a linear relationship between independent variables $X=\{x_1, x_2, ..., x_p\}$ and a dependent variable $y$. The goal is to find the optimal hyperplane that minimizes the discrepancy between predicted and actual values~\cite{maulud2020review}. The prediction is given by $\hat{y}=\sum_{i=1}^n \theta_i x_i + \theta_0$, where $\hat{y}$ is the predicted value, $\theta_0$ is the intercept, $\theta_i$ is the coefficient for the feature $x_i$. The model minimizes the mean squared error (MSE) to measure the discrepancy between predictions and actual values to update parameters iteratively $\mathcal{L}(\theta) = \frac{1}{n} \sum_{i=1}^n (y_i-\hat{y}_i)^2$, where $\mathcal{L}(\theta) = \frac{1}{n} \sum_{i=1}^n (y_i-\hat{y}_i)^2$ and $\eta$ is the learning rate. LR is well-suited for DML due to its simplicity and the scalability of optimization methods. In distributed settings, the dataset $X$ and $y$ are partitioned across multiple workers. Each worker computes the gradient of the loss function on its local data and transmits the results to a central server, which aggregates the gradients and updates the global parameters.
}


\textcolor{black}{
\textbf{Decision tree (DT)} is a non-parametric, interpretable model for classification and regression. It constructs a tree-like structure where internal nodes represent features, branches correspond to decision rules, and leaf nodes denote output values. The objective is to recursively partition the dataset into subsets that are as homogeneous as possible concerning the target variable~\cite{myles2004introduction}. Splits are determined by optimizing a criterion such as information gain or variance reduction. Starting from the root node, the model evaluates potential splits and selects the one that maximizes the chosen criterion, repeating this process recursively until a stopping condition is met. For instance, in variance reduction, splits aim to minimize variance $Var(D) = \frac{1}{n} \sum_{i=1}^n (y_i-\overline{y})^2$ within child nodes, where $D$ is the training dataset. 
}
\textcolor{black}{
The training process of DT models may not seamlessly integrate with DML. 
Nevertheless, DTs are widely utilized in large-scale applications such as credit scoring, fraud detection, and customer segmentation, owing to their simplicity, interpretability, and ability to handle both numerical and categorical data. 
Consequently, in DML, DTs are often employed to provide classification and regression services to users.
}


\textcolor{black}{For deep models, we consider neural networks, convolutional neural networks and recurrent neural networks.}
In general, deep models require more computational resources during training compared to traditional models. 

\textcolor{black}{
\textbf{Neural Network (NN)} is highly versatile, capable of modeling complex, non-linear relationships in data, and widely used for tasks such as classification, regression, and feature extraction~\cite{jain1996artificial}. 
At a high level, NNs process input data through weighted connections, introduce non-linearity via activation functions, and iteratively update weights based on the error between predicted and actual outcomes. They consist of three types of layers. \textbf{Input layer} receives input features $X=\{x_1,x_2,...,x_p\}$, where each feature is assigned a weight reflecting its importance. \textbf{Hidden Layers} perform transformations using linear operations $z^{(l)} = W^{(l)}a^{(l-1)}+b^{(l)}$ followed by non-linear activation functions $a^{(l)}=\sigma(z^{(l)})$ in layer $l$, where $z^{(l)}$ is the pre-activation value, $W^{(l)}$ and $b^{(l)}$ are weights and biases, $a^{(l)}$ is the activation output, and $\sigma()$ is the activation function. \textbf{Output layer} produces probabilities using the Softmax function $\hat{y}_i = \frac{\exp(z_i)}{\Sigma_j \exp(z_i)}$. Training is performed via backpropagation, which computes the gradient of the loss function with respect to each weight and bias $\frac{\partial\mathcal{L}}{\partial W^{(l)}} = \frac{\partial\mathcal{L}}{\partial a^{(l)}} \cdot \frac{\partial a^{(l)}}{\partial z^{(l)}} \cdot \frac{\partial z^{(l)}}{\partial W^{(l)}}$, where loss function $\mathcal{L}$ quantifies the error between predicted values $\hat{y}$ and true values $y$. Weights are updated iteratively using an optimization algorithm like gradient descent $W^{(l)} \leftarrow W^{(l)} - \eta \frac{\partial\mathcal{L}}{\partial W^{(l)}}$, where $\eta$ is the learning rate. NNs are computationally intensive, especially with large datasets and deep architectures. In DML, data or models can be distributed across multiple workers, each performing local training and sending updates to a central server for aggregation as federated learning or split learning. Trained models can also be deployed for ML inference services. However, training involves backpropagation, making it more complex and resource-intensive than inference. Consequently, applying ZKP-VML to training poses greater challenges than to inference.
}

\textcolor{black}{
\textbf{Convolutional Neural Network (CNN)} is a specialized NN designed for grid-like data, such as images. Unlike traditional NNs, CNNs employ convolutional layers to extract spatial features to build complex representations and pooling layers to preserve key features. \textbf{Convolutional layers} apply learnable filters that slide over the input to generate feature maps, capturing local structures with $z_{i,j,k} = \sum_{m,n,c}X_{i+m,j+n,c}\cdot W_{m,n,c,k} + b_k$, where $X$ is the input, $W$ is the filter, $b_k$ is the bias for the $k$-th filter, $z_{i,j,k}$ is the output at position $(i,j)$. \textbf{Pooling layers} reduce spatial dimensions, preserving essential features while lowering computational costs. Max pooling, for instance, selects the maximum value within each patch with $z_{i,j,k} = \max_{p,q} X_{i+p,j+q,k}$. 
}
\textcolor{black}{
Training CNNs is computationally demanding, particularly for large datasets and deep architectures. Additionally, representing high-dimensional data and convolutional computations within the ZKP-VML framework presents significant challenges.
}

\textcolor{black}{
\textbf{Recurrent Neural Network (RNN)} processes sequential data by modeling temporal dependencies through recurrent connections, enabling them to retain information from past inputs. This makes them effective for tasks such as time-series analysis, natural language processing, and speech recognition. RNNs maintain a hidden state that updates at each time step based on the current input and the previous hidden state as $h_t = f(W_hh_{t-1}+W_xx_t+b)$, where $W_h$ and $W_x$ are weight matrices, $b$ is the bias vector, $f()$ is an activation function. The output at time step $t$ is computed as $y_t=g(W_yh_t+c)$, where $W_y$ is the output weight matrix, $c$ is the bias vector, and $g()$ is typically a softmax or linear activation function. For sequence modeling, the total loss $\mathcal{L}=\frac{1}{T}\sum_{t=1}^T\ell(y_t,\hat{y}_t)$ is the sum of errors over all time steps, where $\ell$ is the loss function, and $T$ is the sequence length. Gradient computation involves unrolling the network and applying backpropagation through time (BPTT) as $\frac{\partial\mathcal{L}}{\partial W_h} = \sum_{t=1}^T \frac{\partial\mathcal{L}}{\partial h_t} \cdot \frac{\partial h_t}{\partial W_h}$.
}
\textcolor{black}{RNNs are computationally intensive, particularly for long sequences or large datasets, due to their sequential nature. While their core computations resemble those of standard NNs, their concept of state may offer novel optimizations within the ZKP-VML framework.}

\textcolor{black}{
For large models, we consider transformers. Large models are trained on vast datasets with billions of parameters, demand extensive computational resources and large-scale distributed training. 
}

\textcolor{black}{
\textbf{Transformers} revolutionized ML by replacing recurrent and convolutional structures with self-attention mechanisms, enabling parallel processing of entire input sequences. Inputs are tokenized and embedded into continuous vector representations, augmented with positional encoding to retain sequence order as $PE_{(pos,2i)}=sin\left(\frac{pos}{10000^{2i/d}}\right), \notag$ and $PE_{(pos,2i+1)}=cos\left(\frac{pos}{10000^{2i/d}}\right)$. Self-attention computes weighted sums of input tokens, allowing the model to focus on relevant sequence elements with $\mathrm{Attention}(Q,K,V)=\mathrm{Softmax}\left(\frac{QK^T}{\sqrt{d_k}}\right)V$, where $Q=XW_Q, K=XW_K, V=XW_V$, $X$ is the input sequence, $W_Q, W_K, W_V$ are learnable weight matrices, $d_k$ is the dimensionality of $K$. Multi-head attention applies multiple self-attention operations in parallel to capture different input aspects as $\mathrm{Multihead}(Q,K,V) = \mathrm{Concat}(\mathrm{head}_1,...,\mathrm{head}_h)W_O$. 
}
\textcolor{black}{
Transformers are much more computationally intensive. Billions of parameters introduces massive resource and computational overheads for ZKP-VML converting into arithmetic circuit or generating proofs. Besides, transformers rely heavily on non-linear functions and complex operations, which are challenging to represent in ZKP-friendly formats.
}



\subsection{Zero-Knowledge Proof}

Zero-knowledge proof (ZKP)~\cite{goldwasser1985knowledge} is a cryptographic protocol where a prover demonstrates the truth of a statement to a verifier without revealing any additional information. In simple terms, ZKP has three key properties:


\begin{itemize}

    \item \textbf{Completeness:} Given a statement of a witness, the prover can convince the verifier that the statement is correct if the protocol is executed honestly. 
    \item \textbf{Soundness:} Any malicious prover cannot fool the verifier into accepting a false statement.
    \item \textbf{Zero-Knowledge:} The prover does not leak any information else but the correctness of the statement during the protocol.
    
\end{itemize}

\subsubsection{Non-Interactive Zero-Knowledge Argument}

Zero-knowledge proofs can be classified as interactive or non-interactive based on the number of interaction rounds. Interactive protocols involve multiple rounds between the prover and verifier, whereas non-interactive zero-knowledge (NIZK)~\cite{blum1991noninteractive,brassard1988minimum} proofs require only a single message from the prover. Once generated, NIZK proofs can be distributed for independent verification, reducing the need for assumptions like synchronous communication and reliable channels. Due to their efficiency, scalability, and security benefits, NIZK proofs are widely adopted in areas like DML. This paper focuses on NIZKs, which are constructed using two main methods: the Common Reference String (CRS) model and the Random Oracle Model (ROM)~\cite{wei2022overview}.


In the CRS model, it is assumed that a common reference string is generated by a trusted third party and is available to both the prover and the verifier for proof generation and verification. 
Specifically, we provide the formal definition of CRS-based NIZK protocols~\cite{groth2016size}. 
Let $\mathcal{R}$ be a relation generator that given a security parameter $\lambda$ in binary returns a polynomial time decidable binary relation $R$. 
The relation generator may also output some auxiliary input $z$.
For pairs $(\phi, w)\in R$, we call $\phi$ the statement and $w$ the witness. 
We define $\mathcal{R}_\lambda$ to be the set of possible relation $\mathcal{R}$ may output given $1^\lambda$.
An efficient prover publicly verifiable non-interactive argument for $R$ is a quadruple of probabilistic polynomial algorithms $\mathrm{(Setup, Prove, Vfy, Sim)}$ such that

\begin{itemize}
    \item $(\sigma,\tau)\leftarrow\mathrm{Setup}(R):$ The setup take a security parameter $\lambda$ and a relation $R\in R_\lambda$ as input, outputs a common reference string $\sigma$ and a simulation trapdoor $\tau$ for the relation $R$.
    The algorithm serves as an initialization for the circuit and needs to be executed only once for the same circuit.
    \item $\pi\leftarrow\mathrm{Prove}(R,\sigma,\phi,w):$ The prove take a common reference string $\sigma$ and $(\phi,w)\in R$ as input, outputs argument $\pi$.
    \item $0/1\leftarrow\mathrm{Vfy}(R,\sigma,\phi,\pi):$ The vfy takes a common reference string $\sigma$, a statement $\phi$ and an argument $\pi$ as input, outputs 0 (reject) or 1 (accept).
    \item $\pi\leftarrow\mathrm{Sim}(R,\tau,\phi):$ The sim takes a simulation trapdoor $\tau$ and statement $\phi$ as input, outputs an argument $\pi$.
\end{itemize}

\begin{myDef}
    We say $\mathrm{(Setup, Prove, Vfy, Sim)}$ is a non-interactive zero-knowledge argument of knowledge for $R$ if it has perfect completeness, computational knowledge soundness and perfect zero-knowledge as defined below.
\end{myDef}

\textit{Perfect completeness} says that, given any true statement, an honest prover should be able to convince an honest verifier.
For all $\lambda \in \mathbf{N}, R \in \mathcal{R}_\lambda, (\phi,w)\in R$: 
\textcolor{black}{
\begin{equation}
\mathrm{Pr}\left[
\begin{aligned}
&(\sigma,\tau)\leftarrow\mathrm{Setup}(R);\\
&\pi\leftarrow\mathrm{Prove}(R,\sigma,\phi,w):\ \\
&\mathrm{Vfy}(R,\sigma,\phi,\pi)=1
\end{aligned}
\right]=1
\end{equation}
}

\textit{Computational knowledge soundness} says that there exist an extractor that can compute a witness whenever the adversary produces a valid argument.
Formally, we require that for all non-uniform polynomial time adversaries $\mathcal{A}$ there exists a non-uniform polynomial time extractor $\mathcal{X_A}$ such that:
\textcolor{black}{
\begin{equation}
\mathrm{Pr}\left[
\begin{aligned}
&(R,z)\leftarrow\mathcal{R}(1^\lambda);\ 
(\sigma,\tau)\leftarrow\mathrm{Setup}(R);\ \\
&((\phi,\pi);w)\leftarrow\mathcal{A||X_A}(R,z,\sigma):\\
&\phi \notin L_R \rm{\ and\ } \mathrm{Vfy}(R,\sigma,\phi,\pi)=1
\end{aligned}
\right]\approx 0
\end{equation}
}

\textit{Perfect zero-knowledge} says that it leaks no information besides the truth of the statement. 
For all $\lambda \in \mathbf{N}, (R, z) \in \mathcal{R}_\lambda, (\phi,w)\in R$ and all adversaries $\mathcal{A}$: 
\textcolor{black}{
\begin{equation}
\mathrm{Pr}\left[
\begin{aligned}
&(\sigma,\tau)\leftarrow\mathrm{Setup}(R);\\
&\pi\leftarrow\mathrm{Prove}(R,\sigma,\phi,w):\\
&\mathcal{A}(R,z,\sigma,\tau,\pi)=1
\end{aligned}
\right]
=\mathrm{Pr}\left[
\begin{aligned}
&(\sigma,\tau)\leftarrow\mathrm{Setup}(R);\ \\
&\pi\leftarrow\mathrm{Sim}(R,\tau,\phi):\\
&\mathcal{A}(R,z,\sigma,\tau,\pi)=1
\end{aligned}
\right]
\end{equation}
}

In the random oracle model, a hash function can replace the verifier's random challenge, allowing interactive zero-knowledge arguments to be transformed into non-interactive ones via the Fiat-Shamir heuristic~\cite{fiat1986prove}. We demonstrate this process using the Schnorr protocol, an interactive cryptographic method for proving knowledge of a discrete logarithm without revealing the actual value, preserving the zero-knowledge property. The protocol operates in a cyclic group $G$ of prime order $q$ with generator $g$. The prover, who knows a secret value $x$, broadcasts $y = g^x$. The process begins with the prover selecting a random value $r$ and computing $t = g^r$, which is sent to the verifier. The verifier then issues a random challenge $c$. The prover responds by calculating $s = r + cx$ and sending $s$ back. The verifier verifies the validity by checking $g^s = t \cdot y^c$. This ensures the verifier gains no information about the secret $x$ while being convinced of the prover's knowledge of it. 
The Fiat-Shamir heuristic~\cite{fiat1986prove,kalai2017obfuscation,chen2021does} converts the interactive Schnorr protocol into a non-interactive version by replacing the verifier's random challenge with a deterministic one derived from a cryptographic hash function. In this non-interactive protocol, the prover computes the challenge $c = H(y \| t)$, where $H$ is a hash function, instead of waiting for a challenge from the verifier. The verifier re-computes $c$ as $c = H(y \| t)$ for the verification. This transformation removes the need for interaction.

\subsubsection{Common schemes}

We present four common ZKP schemes related to verifiable machine learning. 

\textbf{Sumcheck protocol}~\cite{lund1992algebraic} is an interactive protocol to sum a multivariate polynomial $f:\mathbb{F}^{\ell}\rightarrow\mathbb{F}$ with binary inputs: \textcolor{black}{$H=\sum_{b_1,b_2,...,b_\ell\in{0,1}}{f(b_1,b_2,...,b_\ell)}$}.
Directly summing requires $2^\ell$ computations according to the combinations of $b_i$. 
The sumcheck protocol enables the verifier $V$ to efficiently verify $H$ with the prover $P$ in $\ell$ rounds with proof size of $O(d\ell)$. 
In the first round, $P$ sends a uni-variate polynomial $g_1(x_1)=\sum_{b_2,...,b_\ell}f(x_1,b_2,...,b_\ell)$, $V$ checks whether $H=g_1(0)+g_1(1)$, and sends a random challenge $r_1$ to $P$.
Then in the $i$-th round, $P$ sends a uni-variate polynomial $g_i(x_i)=\sum_{b_{i+1},...,b_\ell}f(r_1,...,r_{i-1},x_i,b_{i+1},...,b_\ell)$, $V$ checks whether $g_{i-1}(r_{i-1})=g_i(0)+g_i(1)$, and sends a random challenge $r_i$ to $P$. 
In the $\ell$-th round, which is the last round, $P$ sends a uni-variate polynomial $g_\ell(x_\ell)=f(r_1,...,r_{\ell-1},x_\ell)$, $V$ checks whether $g_{\ell-1}(r_{\ell-1})=g_\ell(0)+g_\ell(1)$, and generates a random challenge $r_\ell$. 
With the access to $f(\cdot)$, $V$ accepts the proof if $g_\ell(r_\ell)=f(r_1,...,r_{\ell-1},r_\ell)$. 
In the sumcheck protocol, the prover $P$ eventually needs to expose the polynomial in order for the verifier $V$ to evaluate at these challenge points, which might leak knowledge.
By applying a zero-knowledge polynomial commitment following the framework in~\cite{wahby2018doubly}, the polynomial can be concealed and zero-knowledge can be achieved.
Furthermore, the interactive challenge process can be replaced by the Fiat-Shamir heuristic~\cite{fiat1986prove}, and the sumcheck protocol thus becomes a NIZK protocol.

\textbf{Quadratic arithmetic program (QAP)}~\cite{gennaro2013quadratic} is an effective coding method of circuit satisfaction (C-SAT) problem, by which the C-SAT problem can be reduced to divisibility problem between polynomials. 
The C-SAT problem refers to whether, for a given circuit, there exists an input such that the output of the circuit is 1.

\begin{myDef}
    A quadratic arithmetic program~\cite{gennaro2013quadratic} $\mathcal{Q}$ converted from arithmetic circuit $\mathcal{C}$ in field $\mathbb{F}$ is consisting of three sets of polynomials ${{u_i(x),\ v_i(x),w_i(x)}}_{i=0}^m$ and a target polynomial $t(x)$. 
    For public inputs and outputs $(c_1,...,c_\ell)$, the $\mathcal{Q}$ is satisfiable if and only if there exist coefficients $(c_{\ell+1},...,c_m)$ such that $t(x)$ divides $p(x)$, where
    $p(x)=(\sum_{i=1}^{m}{c_i\cdot u_i(x)})\cdot(\sum_{i=1}^{m}{c_i\cdot v_i(x)})-\sum_{i=1}^{m}{c_i\cdot w_i(x)}$.
\end{myDef}

Based on the QAP and the common reference string (CRS) model~\cite{blum1988non}, Gennaro et al.~\cite{gennaro2013quadratic} constructed zero-knowledge succinct non-interactive argument of knowledge (zk-SNARK).
In brief, a zk-SNARK scheme for relation $R$ is a quadruple of probabilistic polynomial algorithms $\mathrm{(Setup, Prove, Vfy)}$ such that: 

\begin{itemize}
    \item $(\sigma,\tau)\leftarrow\mathrm{Setup}(R):$ The \textit{Setup} takes a relation $R$ as input, outputs a CRS $\sigma$ and the trapdoor $\tau$.
    \item $\pi\leftarrow\mathrm{Prove}(R,\sigma,\phi,w):$ The \textit{Prove}, under relation $R$, takes a CRS $\sigma$ and $(\phi,w)\in R$ as input, outputs proof $\pi$.
    \item $0/1\leftarrow\mathrm{Verify}(R,\sigma,\phi,\pi):$ The \textit{Verify}, under relation $R$, takes a CRS $\sigma$, a statement $\phi$ and a proof $\pi$ as input, outputs 0 (reject) or 1 (accept).
\end{itemize}

\textbf{Inner product argument (IPA)}~\cite{bootle2016efficient} is an interactive protocol to prove the knowledge of the inner product of vectors $\vec{a}$ and $\vec{b}$ such that \textcolor{black}{$z=\vec{a}\cdot \vec{b} = \sum^n_{i=1}a_ib_i$}
with corresponding commitment $C=\sum_{i=1}^na_iG_i+\sum_{i=1}^nb_iH_i$. 
The core idea of IPA is to recursively reduce the length of the vector up to the scalar by random challenges in several rounds. 
With the Fiat-Shamir heuristic~\cite{fiat1986prove}, the IPA protocol can be non-interactive. 
In order to construct a zero-knowledge proof from IPA, the arithmetic circuit constraints between and among multiplication gates are first formalized by the Schwartz-Zippel Lemma~\cite{schwartz1980fast, zippel1979probabilistic} as the problem where the coefficients of particular terms in a polynomial are zero.
And then it is transformed into a statement of inner product form, which can be handled by IPA directly.

\textbf{MPC-in-the-head}~\cite{ishai2007zero} is an approach to constructing zero-knowledge proofs via secure multi-party computation (MPC) protocols.

\begin{myDef}
    A secure multi-party computation (MPC) protocol~\cite{canetti2000security} $\Pi_f$ allows $n$ participants $P_1,... ,P_n$ to compute a common output $y=f(x,w_1,r_1,...)$ based on the common input $x$, respective secret input $w_1,... ,w_n$ and random inputs $r_1,... ,r_n$ through $k$ rounds of interaction processes.
    Let the view of participant $P_i$ be $V_i=(x, w_i, r_i, M_i)$, where $M_i=(m_{i,1},...,m_{i,k})$ denotes the message received by $P_i$.
    Then the behavior of participant $P_i$ can be determined by the number of rounds $j$ and the current view $V_{i,j}=(x, w_i, r_i, (m_{i,1},...,m_{i,j}))$.
\end{myDef}

The prover simulates several participants to executes a multi-party computation protocol and saves the views of each participant in the process. 
The verifier can check these views through commitments to verify the correctness of the protocol execution process.

\subsubsection{Advantages of Zero-Knowledge Proof}

While various cryptographic techniques address verifiable machine learning to some extent, they each have practical limitations. Zero-knowledge proofs (ZKPs), however, not only ensure computational integrity but also prevent information leakage, making them well-suited for verifiable machine learning. Below, we outline the drawbacks of alternative cryptographic approaches.


\textbf{Secure Multi-Party Computation (MPC)} enables multiple parties to jointly compute a function without revealing their private inputs~\cite{zhao2019secure}. However, MPC requires multiple rounds of interaction and synchronized communication, increasing complexity. In contrast, many ZKPs are non-interactive, reducing network assumptions and communication overhead.


\textbf{Homomorphic Encryption (HE)} allows computations on encrypted data without prior decryption~\cite{acar2018survey}. HE schemes fall into three categories: partially homomorphic encryption (supporting only addition or multiplication), somewhat homomorphic encryption (allowing limited additions and multiplications), and fully homomorphic encryption (supporting arbitrary computations). While fully homomorphic encryption enables complex computations for ML computations, its prohibitive computational cost makes it impractical for machine learning. ZKPs, in comparison, offer significantly lower computational overhead, enhancing feasibility for VML.


\textbf{Differential Privacy (DP)} protects individual data privacy while preserving statistical utility~\cite{dwork2008differential}. However, DP does not provide verifiability. Unlike DP, ZKPs ensure both privacy and verifiability through their zero-knowledge properties.


\textbf{Trusted Execution Environments (TEE)} ensure secure code execution via isolated hardware~\cite{sabt2015trusted}. While TEE and ZKPs both guarantee correctness and privacy, TEE depends on a trusted hardware provider, introducing additional security assumptions and implementation costs. In contrast, ZKP security relies solely on cryptographic assumptions, avoiding extra hardware expenses.

Overall, ZKPs provide a balance of efficiency, privacy, and verifiability, making them a strong candidate for VML applications.


\subsection{Communication in ZKP-VML} 

The simplest approach to verify machine learning computations is for the performer to send all input and output data to the requester, who then re-executes the computation to verify the correctness. However, this method compromises data privacy and significantly increases both communication and computational costs for the requester. ZKP-VML addresses these issues by ensuring privacy and reducing communication overhead at the cost of increased computation on the performer’s side. 


\subsubsection{Communication Architecture}

Different DML frameworks lead to distinct communication architectures, each requiring tailored privacy and verifiability mechanisms. We summarize three common architectures for further analysis on communication.


\begin{figure}
    \centering
    \includegraphics[width=0.4\textwidth]{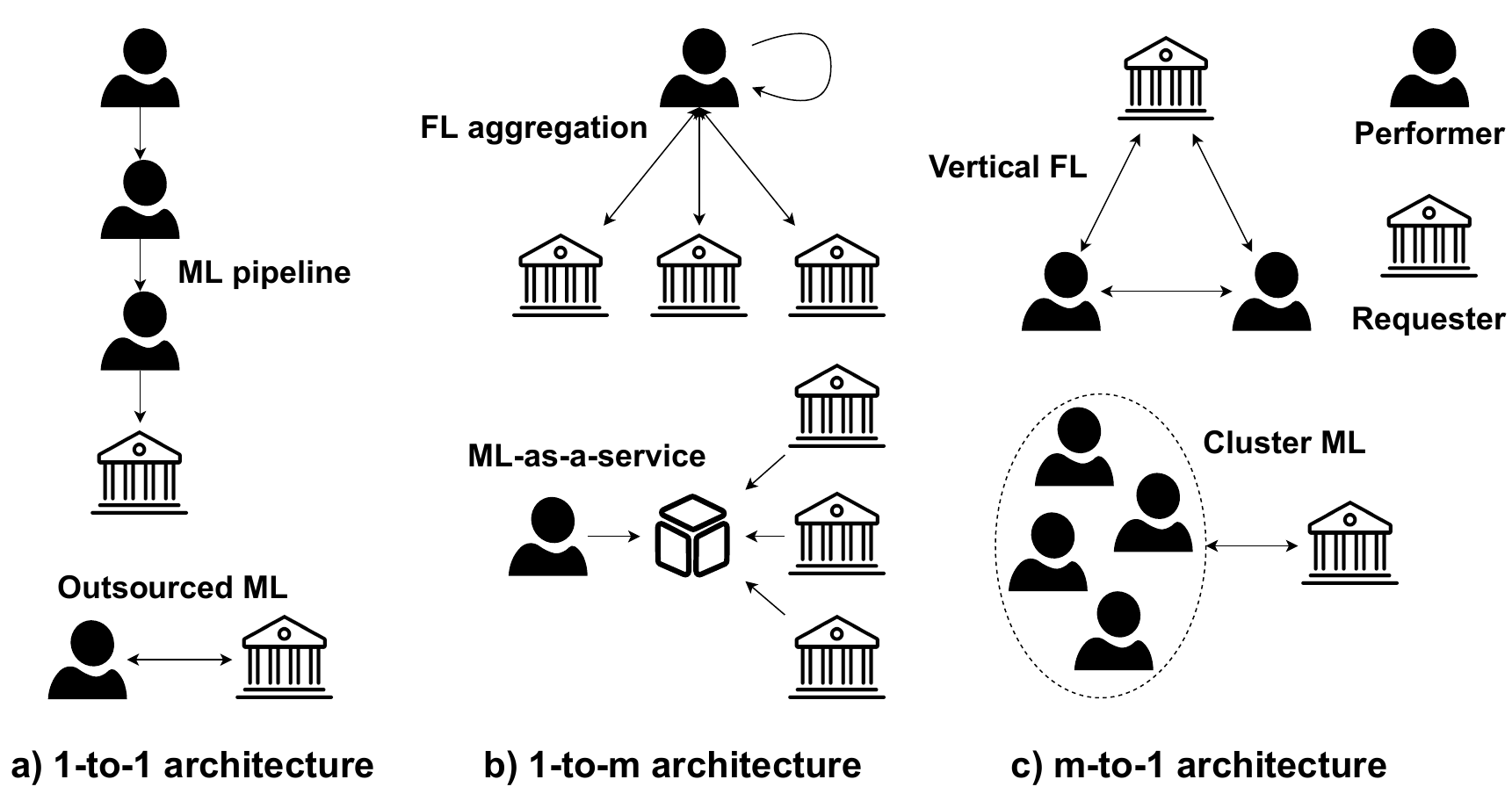}
    \caption{The architecture of different verifiable machine learning.} 
    \label{figarch}
\end{figure}



\textbf{1-to-1. }
This architecture contains only 1 requester and 1 performer. 
After receiving the ML task and public data, the performer can execute the computation locally and generate a proof.
This architecture is common in ML pipelines, outsourced ML, and various DML scenarios.
The requester can run the setup step if the trusted setup is needed for the ZKP protocol. 

\textbf{1-to-m. } 
This architecture contains 1 performer and multiple requesters. 
The performer has to prove the correctness of the computation to multiple requesters. 
To minimize redundant communication, non-interactive ZKPs can be adopted, which allows requesters to verify proofs independently once published.
The 1-to-m architecture may exist in FL aggregation, ML-as-a-service, and many other DML scenarios. 
The multiple requesters have to run the setup step collaboratively if the trusted setup is needed for the ZKP protocol. 
Further, the 1-to-m architecture involves extra privacy issues if there exists private data between different requesters. 
For example, in FL aggregation, the server has to provide privacy protection of each local model. 
While in ML-as-a-service, the server only has to generate a proof arguing for the performance of its model, and each user can check the proof at their use. 

\textbf{m-to-1. }
This architecture contains multiple performers and 1 requester. 
Each performer is responsible for one part of the ML task. 
Performers are asked not only generate proofs for their local computations, but also jointly prove the correctness of their interactions, ensuring end-to-end verification. 
This architecture is relevant to vertical FL, cluster ML, and similar scenarios.
The requester has to run the setup step for different kinds of computations within the task if the trusted setup is needed for the ZKP protocol. 
Further, the m-to-1 architecture involves extra privacy issues if there exists private data between different performers. 
For example, in vertical FL, the training data in each participant is private. 
While in cluster ML, all the node shares the training dataset and model parameters. 

\subsubsection{\textcolor{black}{Attackers in Communication}}
\textcolor{black}{
In the aforementioned DML communication architectures, both the requester and the performer can potentially act as adversaries, thereby compromising the security and privacy of the machine learning system. 
Specifically, a malicious requester typically targets data privacy, attempting to extract sensitive information from the performers, whereas a malicious performer primarily undermines data security by providing incorrect computational outputs. 
Several classical attack methods in machine learning can be adapted to exploit these vulnerabilities. 
}

\textcolor{black}{
When performers act as adversaries, they may intentionally provide incorrect results during computation. In machine learning, two major performer attacks are poisoning attacks and free-rider attacks.
\textbf{Poisoning attacks} degrade model performance by introducing incorrect patterns into the learning process, leading to misclassifications (e.g., assigning label $b$ to the sample in class $a$). Adversaries may inject mislabeled and poisoned data into the training dataset or directly manipulate model parameters, degrading the accuracy and reliability of the model delivered to the requester. 
\textbf{Free-rider attacks} allow adversaries to submit results and complete tasks without performing the required computations. These attacks often arise in collaborative scenarios, such as federated learning or cluster machine learning, where multiple performers execute the similar computational task for some rewards. Adversaries may copy and reuse results submitted by other performers, deceiving the requester into accepting these results, thereby stealing training rewards while minimizing computational costs.
}

\textcolor{black}{
When requesters act as adversaries, they may attempt to infer private data used by performers, compromising data privacy. In machine learning, two major requester attacks are membership inference attacks and reconstruction attacks.
\textbf{Membership inference attacks} seek to determine whether a specific sample is included in the training dataset of the model. These attacks exploit the observation that models generally exhibit higher confidence and accuracy on their training data than on unseen data, as the model has directly learned patterns and features from these samples. Adversaries may construct shadow models that replicate the behavior of the target model, or develop specialized metrics to evaluate the confidence levels of the model inference results. 
\textbf{Reconstruction attacks} aim to reconstruct private training data by leveraging model gradient updates. Since similar training data produce similar gradient updates, adversaries can iteratively refine candidate inputs until their gradients match those obtained from the performer. This process can reconstruct data similar to private training data, violating the privacy of performers.
}

\textcolor{black}{
Additionally, adversaries may also disrupt the ML task by refusing to respond or providing incorrect responses, leading to communication failures and task disruption.
Such attacks pose a greater threat in ML scenarios with additional computational and communication processes to ensure privacy. 
}



%
%

A list of key acronyms and abbreviations used throughout the paper is given in Table~\ref{Acronyms}.

\begin{table}[htbp]
\centering
\caption{List of Key Acronyms}
\label{Acronyms}
\begin{tabular}{c|c}
\hline
\thead{Acronyms} & \thead{Definitions} \\ \hline
\multirow{2}{*}{\textcolor{black}{ZKP-VML}} & \textcolor{black}{Zero-Knowledge Proof-based} \\ 
& \textcolor{black}{Verifiable Machine Learning} \\
ML & Machine Learning \\
\textcolor{black}{DML} & \textcolor{black}{Decentralized Machine Learning} \\
ZKP & Zero-Knowledge Proof \\
\multirow{2}{*}{zk-SNARK} & Zero-Knowledge Succinct \\ 
& Non-interactive Argument of Knowledge \\
QAP & Quadratic Arithmetic Problem \\
QPP & Quadratic Polynomial Problem \\
IPA & Inner Product Arguments \\
R1CS & Rank-1 Constraint System \\
\textcolor{black}{VOLE} & \textcolor{black}{Vector-Oblivious Linear Evaluation} \\
\textcolor{black}{MPC} & \textcolor{black}{Multi-Party Computation} \\
DP & Differential Privacy \\
HE & Homomorphic Encryption \\
FL & Federated Learning \\
NN & Neural Network \\
CNN & Convolutional Neural Network \\
\textcolor{black}{RNN} & \textcolor{black}{Recurrent Neural Network} \\
DT & Decision Tree \\
\textcolor{black}{LLM} & \textcolor{black}{Large Language Model} \\
\hline
\end{tabular}
\end{table}

\section{Zero-Knowledge Proof-based Verifiable Machine Learning\label{sec3}}

\subsection{Definition}

Considering that most of the verifiability issues in distributed machine learning exist among inference and training tasks, we give a pioneering definition of zero-knowledge proof-based verifiable machine learning (ZKP-VML), in order to provide an evaluation criterion for existing work.
We model ZKP-VML through participants, algorithms, and workflows.

\textcolor{black}{
There are two types of participants, the prover $P$ and the verifier $V$. 
$V$ delegates ML tasks (training or inference) to $P$, and expects $P$ to return the correct result. 
$P$ performs the required ML tasks to return the result, while generating proofs of its correctness, which is verified by $V$. 
Both $P$ and $V$ can be malicious, leading to incorrect results being submitted or private data being leaked. 
}


Gennaro, Gentry and Parno~\cite{gennaro2010non} provided a formal definition of outsourced computation with four algorithms. In order to provide an accurate and unambiguous description, we define the non-interactive verifiable machine learning modeled after that.

\begin{myDef}
    \textit{A verifiable machine learning scheme allowing a prover $P$ to trustfully execute given machine learning task $T$ with specific input data $D$ for verifier $V$ is a tuple of below algorithms:} 
    \begin{itemize}
        
        \item $(PK, VK)\leftarrow \textbf{Setup}(T, \lambda)$: By inputting the task function $T$ and a security parameter $\lambda$, the \textit{setup} algorithm generates a proving key $PK$ for the proof generation and a verification key $VK$ for the proof verification.
        
        \item $R\leftarrow \textbf{Execute}(T, D):$ By inputting the task function $T$, both the private and public input data of the prover $D = \{D_{pri}, D_{pub}\}$, the \textit{execution} algorithm perform the task $T$ and outputs the execution result $R$. 
        
        \item $\pi\leftarrow \textbf{Prove}(PK, D, R):$ By inputting the proving key $PK$, input data $D$ and execution result $R$, the \textit{proving} algorithm generates the corresponding proof $\pi$.
        
        \item $[0,1]\leftarrow \textbf{Verify}(VK, \pi, D_{pub}, R):$ By inputting the verification key $VK$, proof $\pi$, public input data $D_{pub}$ and the result $R$, the \textit{verification} algorithm checks whether the proof $\pi$ is valid. If the verification pass, the algorithm outputs a positive number not greater than 1, otherwise 0.


    \end{itemize}
\end{myDef}

\begin{figure}
    \centering
    \includegraphics[width=0.4\textwidth]{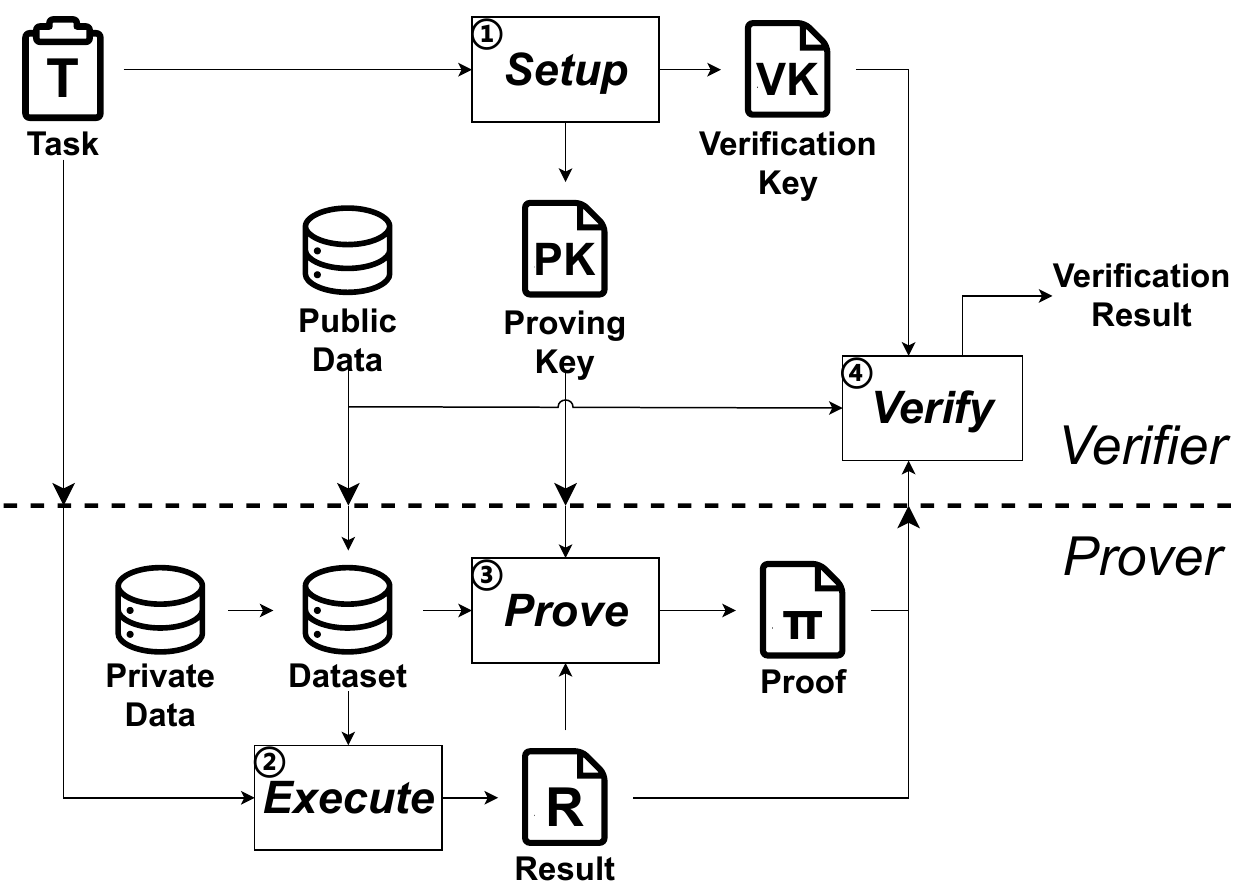}
    \caption{The workflow of verifiable machine learning.} 
    \label{figworkflow}
\end{figure}

A ZKP-VML scheme should be both correct and secure. 
More formally: 

\textit{Correctness.} If $P$ has honestly completed the task with the given input, then \textbf{Verify} outputs 1 (Accept) with the probability of 1. 
\begin{equation}
    \mathrm{Pr}\left[
    \begin{aligned}
    &(PK,VK)\leftarrow \textbf{Setup}(T,\lambda);\\
    &R\leftarrow \textbf{Execute}(T, D);\\
    &\pi\leftarrow \textbf{Prove}(PK, D, R):\\
    &\textbf{Verify}(VK, \pi, D_{pub}, R)>0
    \end{aligned}
    \right]=1
\end{equation}
    
\textit{Security.} If $P$ is malicious then \textbf{Verify} outputs 1 with a negligible probability. For any PPT adversary $\mathcal{A}$, 
\begin{equation}
    \mathrm{Pr}\left[
    \begin{aligned}
    &(PK,VK)\leftarrow \textbf{Setup}(T,\lambda);\\
    &R\leftarrow \textbf{Execute}(T, D);\\
    &(R',\pi')\leftarrow \mathcal{A}(PK,T):\\
    &R'\neq R \land \textbf{Verify}(VK, \pi', D_{pub}, R')>0
    \end{aligned}
    \right]\leq negl(\lambda)
\end{equation}

In addition, we should also consider the privacy of the proof verification process and the characteristic of machine learning computation results.

\textit{Partial Privacy.} If the private data $D_{pri}\neq\emptyset$.
    
\textit{Privacy.} If the scheme satisfies \textit{partial privacy} and further, leaks no information that contributes to the inference or reconstruction of the prover's private input $D_{pri}$ besides the verification result to $V$. For any PPT adversary $\mathcal{A}$, 
\begin{align}
    \mathrm{Pr}\left[
    \begin{aligned}
    &(PK,VK)\leftarrow \textbf{Setup}(T,\lambda);\\
    &d\in D_{pri} ;R\leftarrow \textbf{Execute}(T, D);\\
    &\pi\leftarrow \textbf{Prove}(PK, D, R):\\
    &1\leftarrow \mathcal{A}(VK,T,\pi,D_{pub},R,d)
    \end{aligned}
    \right]\notag\\
    =\mathrm{Pr}\left[
    \begin{aligned}
    &(PK,VK)\leftarrow \textbf{Setup}(T,\lambda);\\
    &d'\notin D_{pri} ;R\leftarrow \textbf{Execute}(T, D);\\
    &\pi\leftarrow \textbf{Prove}(PK, D, R):\\
    &1\leftarrow \mathcal{A}(VK,T,\pi,D_{pub},R,d')
    \end{aligned}
    \right]
\end{align}

\textit{Distinctness.} If there exist a task function $T$ and three distinct input data $D_1, D_2, D_3$, with their execution results and proofs respectively, satisfying $\mathbf{Verify}(VK,\pi_1,D_{pub},R_1)<\mathbf{Verify}(VK,\pi_2,D_{pub},R_2)<\mathbf{Verify}(VK,\pi_3,D_{pub},R_3)$. 



We will explain the meaning of the above algorithms and properties into further detail by providing a basic example workflow of a verifiable machine learning scheme as is shown in Fig.~\ref{figworkflow}. 
First, the verifier runs the \textbf{Setup} to generate the proving key $PK$ and the verification key $VK$ according to the machine learning task $T$. 
In this process, the machine learning computation process is transformed into some representation of the problem with zero-knowledge proofs, mostly in the form of arithmetic circuits. 
That is, the computational process to be proved is transformed into an equivalent arithmetic circuit representation. 
Thus the correctness of the computation can be verified by proving the C-SAT problem, i.e., that the output is obtained via the claimed circuit with some input. 
By sending the proving key $PK$, task $T$, and some necessary public data $D_{pub}$ to the prover $P$, the verifier can delegate the task to the prover. 
Thus the prover can \textbf{Execute} the task $T$ with the dataset $D$ to obtain the result $R$.
It is worth mentioning that $R$ does not refer to the computational result $\hat{R}$ of a pure machine learning task, such as the plaintext of the optimized model parameters in a training task. 
$R$ could be the encrypted, masked, or committed result, taking into account some additional privacy settings.
The proof $\pi$ arguing for the correctness of $R$ can be generated with \textbf{Prove}. 
Finally, the verifier can \textbf{Verify} whether the result $R$ is correct with the proof $\pi$. 
Another point worth noting is that the result returned by \textbf{Verify} is not 0 (reject) or 1 (accept), but a result within the interval from 0 to 1, where 0 means reject and any positive number within the interval means accept. 
This is due to that the computational results of machine learning tasks possess some non-binary additional properties, such as accuracy. 
Such multi-result verification algorithm can provide room for the scheme to evaluate of the accuracy or other properties of the results. 

As for the properties of verifiable machine learning, correctness guarantees that the result of an honest execution will always pass the verification, while security guarantees that a malicious result will be rejected with high probability. 
It is worth mentioning that, some of the verifiable machine learning schemes cannot fully satisfy the correctness and security, these schemes will misjudge the proof with an uncertain probability. 
We define them as \textit{partial correctness and security} in this paper.
Partial privacy means that in this machine learning task, there does exist some private data to the prover. 
These private data will not directly leak to the verifier during the verification due to the algorithm itself. 
Considering some privacy attacks on machine learning based on the computational results, such as the membership inference attack, data reconstruction attack, we further provide the definition of privacy. 
Privacy means that the computation and verification process provide no information that helps the verifier to identify whether a certain data is in the private dataset or not. 
Distinctness refers to the ability of a verifiable machine learning scheme to give a non-binary evaluation criterion more than acceptance or rejection of the computation results. 
Unlike other computations, the computational results of machine learning tasks are generally non-deterministic, and for some machine learning tasks, there is no absolute right and wrong about the result. 
For example, in model training, the verifier obtains two local models from different provers, both of which are computed correctly, how can the verifier further distinguish and evaluate these two models? 

\subsection{\textcolor{black}{Threat Model}} 

\textcolor{black}{
In the context of the ZKP-VML scenario defined above, we consider the threat model of the system in terms of the roles, capabilities, and objectives of the attacker. }

\textcolor{black}{
When the performer acts as an attacker, integrity attacks are executed. \textbf{Integrity attacks} introduce errors into the computation process, leading to incorrect results being returned to the requester. In multi-round machine learning tasks, integrity attacks can be further classified as single-round or multi-round. A single-round integrity attack occurs when the attacker manipulates results in one specific round while behaving normally in others, primarily to introduce random errors. In contrast, a multi-round integrity attack involves selectively poisoning a subset of rounds, enabling more subtle and strategic manipulation of results. Attackers may also collaborate with other malicious executors to manipulate deliberate errors. Notably, integrity attacks encompass any deliberate submission of incorrect results. Common integrity attacks include poisoning attacks and free-rider attacks.
For integrity attacks, ZKPs enable efficient verification of the performer’s local computation, allowing the requester to verify with minimal computational overhead, whether the submitted results align with the claimed computation process. This prevents attackers from returning incorrect results. Additionally, ZKPs provide further insight into specific properties within the computation to detect poisoned data used in the computational process, mitigating the risk of an attacker using malicious inputs to generate poisoned results through correct computation processes. In multi-round tasks involving multiple performers, the increased complexity of computation and communication introduces additional security risks. To mitigate these risks, supplementary cryptographic techniques such as HE and MPC may be necessary. 
}

\textcolor{black}{
When the requester assumes the role of an attacker, privacy attacks are conducted. \textbf{Privacy attacks} steal information about the private data used in ML tasks, leading to varying degrees of privacy leakage. The data available to the attacker is limited to the submitted computation results and the communication contents during the interaction. In multi-performer scenarios, the malicious requester may collude with certain performers to gain additional data, further facilitating privacy attacks. Consequently, performers can also act as attackers if they have access to transcriptions. Common privacy attacks include membership inference attacks and reconstruction attacks. 
For privacy attacks, it is crucial to ensure that the original ML framework does not expose the performer’s private data. Cryptographic techniques like DP and HE can protect computation results, making it difficult for attackers to infer or reconstruct private inputs, while maintaining the usability of computation results. Since ZKP-VML relies on ZKPs for verification, it does not introduce additional privacy risks, even if a malicious requester colludes with certain performers.
}

\subsection{Challenges\label{sec_cha}}

Based on the above mentioned workflow, we clarify the challenges that a ZKP-based verifiable machine learning scheme faces.
Dividing the above process by algorithm, which is the action of parties, the challenges are mainly present in \textit{Setup}, \textit{Prove}, and \textit{Verify}. 
These additional algorithms are designed to introduce verifiability to machine learning, and therefore also introduce additional challenges.

The first challenge is the \textit{Generalizability}, which lies in the \textit{Setup} step. 
Due to the nature of the training data, model parameters, etc., the vast majority of computations involved in machine learning are floating-point computations. 
However, as a cryptographic technique, zero-knowledge proofs work on the group of finite fields, which are the integers. 
Therefore, the floating-point computational process cannot be directly mapped to the group without accuracy loss. 
Moreover, zero-knowledge proofs can only support additive and multiplicative operations, making it difficult to represent complex non-linear activation functions in neural networks. 
Thus the first challenge is about how to transform the computational process of machine learning tasks into a zero-knowledge proof problem and representation. 
This challenge can be divided into two sub-problem, one is float-integer conversion problem, and another is non-linear computation conversion problem. 
However, this challenge does not serve as the main research direction of ZKP-VML, even though each ZKP-VML scheme needs to face and address these problems. 
Existing ZKP-VML schemes tend to implement simpler solutions or existing schemes, such as directly scaling up the floating-number and then truncating them, or using existing mapping methods.

The second challenge is the \textit{Efficiency}, which lies in the \textit{Prove} step. 
Machine learning is a computationally intensive application. 
For the neural networks commonly used today, the number of parameters are all at least in the millions. 
And the time complexity of zero-knowledge proofs is basically related to the circuit size. 
Such massive computations with millions of neurons can lead to the large size of arithmetic circuits, which in turn leads to the unaffordable computational cost for zero-knowledge proofs. 
To visualize this problem, we provide a real example. 
Zero-knowledge succinct non-interactive argument of knowledge~\cite{gennaro2013quadratic} (zk-SNARK) scheme is one of the most commonly used type of zero-knowledge proof schemes, and Groth16~\cite{groth2016size} is the one of the most commonly used of zk-SNARKs. 
For the neural network model VGG16~\cite{simonyan2014very} with 138M parameters, the proof generation time reaches a staggering 10 years. 
Also the size of the generated proving key and verification key exceeds 1,000 TB. 
The second challenge is about how to efficiently generate the proofs for the computational process of machine learning tasks. 
This challenge is the main research direction of ZKP-VML. 
Most ZKP-VML schemes work on improving the proving efficiency in different ways to reduce the additional burden introduced by ZKP, making the scheme more practical. 
The technical routes of existing work include: 
1. Optimizing the circuit representation. 
Which involves how to reduce the size of the circuit of the equivalent representation for the given computational process. 
2. Improving the efficiency of verification. 
Which involves how to improve the efficiency of proof generation and verification at the cost of acceptable security errors for the given computational process. 

The third challenge is the \textit{Evaluability}, which lies in the \textit{Verify} step. 
Zero-knowledge proofs can be used to prove the correctness of the computational process of a machine learning task, avoiding the result without honest computations submitted by lazy or malicious participants. 
However, this is not an adequate to defense against attacks and dishonest behaviours. 
Considering participants without an honest dataset, they can still perform the machine learning task and honestly generate valid proofs for the results. 
It is difficult for the verifier to detect such malicious behaviour, for the malicious result comes with a valid proof, arguing for the "honest" execution process. 
This issue is equally important for the verifiability. 
Thus the third challenge is about how to evaluate the result beyond the verifiability. 
There exists several technical routes, aiming at identify different properties of the submitted result, such as the accuracy, the integrity, or the fairness of the model. 
Thus a filter can be designed to distinguish between malicious and benign results, and this filter can be implemented with zero-knowledge proofs within a privacy-preserving manner. 

\subsection{\textcolor{black}{Applications in Communication Networks}}

\textcolor{black}{
The relationship between ZKP-VML and communication networks is close and various. 
In distributed systems, communication networks serve as the foundation for ZKP-VML, with the structure and efficiency of the network significantly impacting the overall performance of ZKP-VML. 
Consequently, improving communication efficiency and reducing communication overhead are among the key challenges for ZKP-VML. 
Moreover, the evolution of communication networks has introduced new tasks and scenarios for ZKP-VML. 
Semantic communication, a novel communication paradigm, aims to enhance communication efficiency, reduce transmission redundancy, and better support intelligent applications by conveying the meaning or intent of information rather than the traditional symbols or data themselves~\cite{luo2022semantic}. 
Compared to conventional communication systems, semantic communication emphasizes the understanding, reasoning, and contextual relevance of information~\cite{cheng2024goal}. 
While semantic communication offers significant support for machine learning, it also introduces new security challenges. 
Malicious adversaries could exploit semantic communication to generate misleading information, deceiving the receiver~\cite{hu2023robust}, or launch privacy attacks when participants sharing background knowledge~\cite{luo2023encrypted}. 
This presents a new application scenario for ZKP-VML: \textit{verifying the correctness of semantic communication without disclosing additional information}~\cite{lin2023blockchain}. 
In real-world distributed machine learning scenarios, participants often lack mutual knowledge and trust, making them more vulnerable to malicious actors. 
Adversaries may infiltrate the system and disrupt the machine learning task through malicious actions. 
Therefore, another key application of ZKP-VML is to \textit{provide identity authentication for participants while ensuring privacy protection}~\cite{dwivedi2022privacy}, confirming their ability to engage in the machine learning task. 
Furthermore, the results of these machine learning tasks can be leveraged to optimize resource allocation in communication networks~\cite{kawamoto2023traffic}, thereby enhancing communication efficiency and robustness~\cite{rodrigues2024smart}.
}

\section{\textcolor{black}{Overview of ZKP-VML Research Status}\label{sec4}}

\textcolor{black}{In this section, we provide a comprehensive overview of the existing ZKP-VML schemes. }
First of all, we provide the scope of the ZKP-VML schemes covered in this paper. 
We try to cover, as far as possible, all ZKP-VML schemes until June 2024, i.e., schemes that use zero-knowledge proofs to provide privacy and verifiability for machine learning computations. 
Our survey mainly relies on google scholar and dblp to search for relevant literature. 
Beyond this, we also cover a small number of schemes that fail to sufficiently address privacy, for example SafetyNets~\cite{ghodsi2017safetynets}, as an introduction to the issue of ZKP-VML. 
It is worth mentioning that several kinds of schemes are excluded from our coverage for their limited contribution to the ZKP-VML issue: 
\begin{enumerate}

    \item General zero-knowledge proof schemes, such as DIZK~\cite{wu2018dizk}, zk-Authfeed~\cite{wan2022zk}, which have not been specifically optimized for ML computations, can be used in many other scenarios. 
    
    \item Schemes not directly using ZKPs, such as Drynx~\cite{froelicher2020drynx}, Guo et al.~\cite{guo2020secure}, GOPA~\cite{sabater2022accurate}, which just include zero-knowledge proofs as a sub-module, rather than directly providing verifiability of ML computations.
    For example, Guo et al. utilize a ZKP protocol to prove the correctness of the encryption. 
    
    \item Schemes with limited confidence, such as Ju et al.~\cite{ju2021efficient}, Ghaffaripour et al.~\cite{ghaffaripour2021mutually}, which lack sufficient security and experimental analysis to demonstrate the feasibility of their schemes.

    
\end{enumerate}



\subsection{Timeline}

\begin{figure*}[htbp]
    \centering
    \includegraphics[width=1\textwidth]{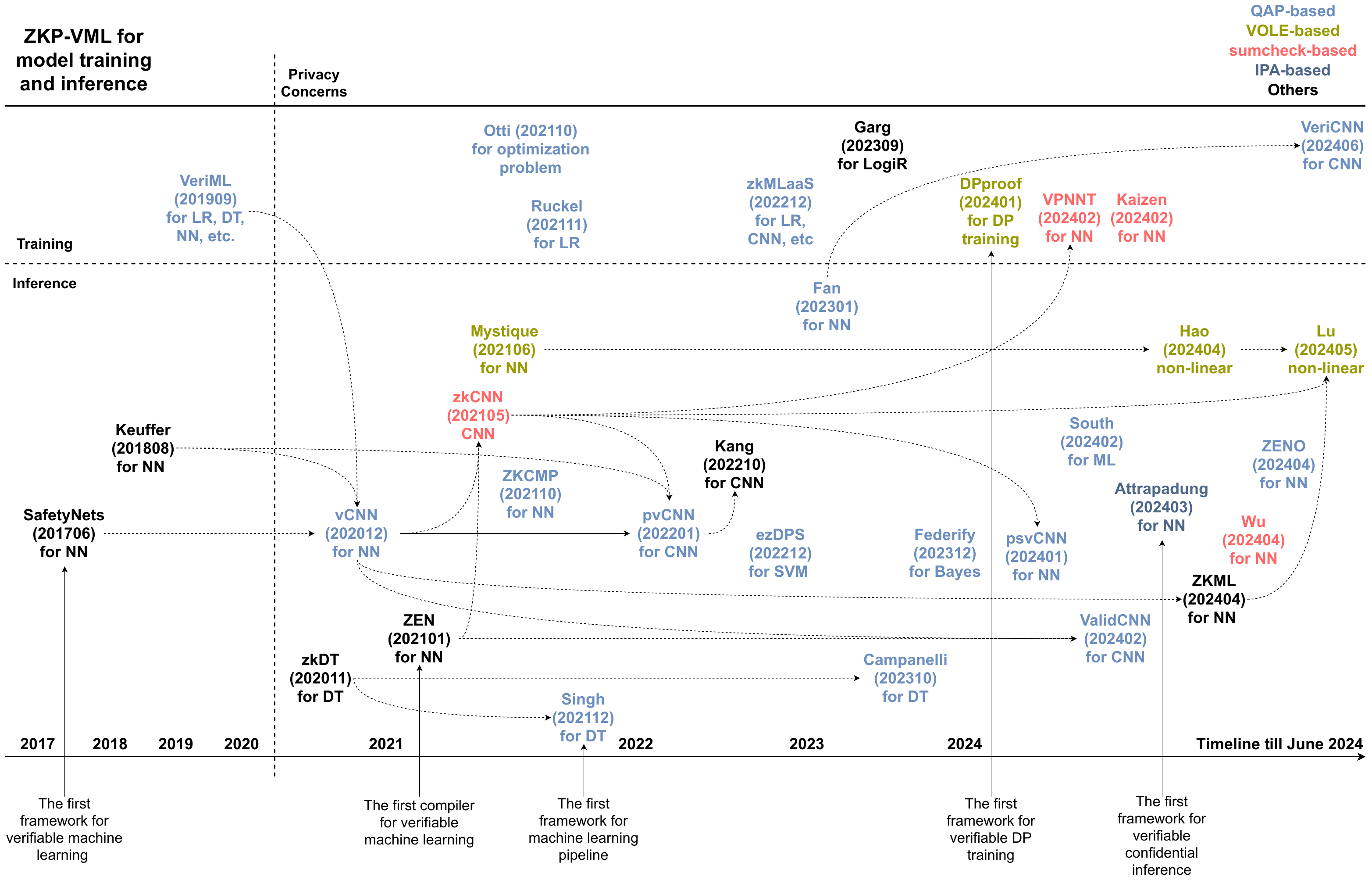}
    \caption{Timeline of existing work in the field of ZKP-VML for training and inference. 
    The dash line from scheme A to B indicates that B is shown to be more advanced than A under certain conditions by theoretical or experimental analysis. 
    The solid line from schemes A to B indicates that B is inspired by A. 
    For brevity, we omit the dash line from A to C, when A points to both B, C and B points to C. \label{timeline}}
\end{figure*}

\begin{figure*}[!htbp]
    \centering
    \includegraphics[width=1\textwidth]{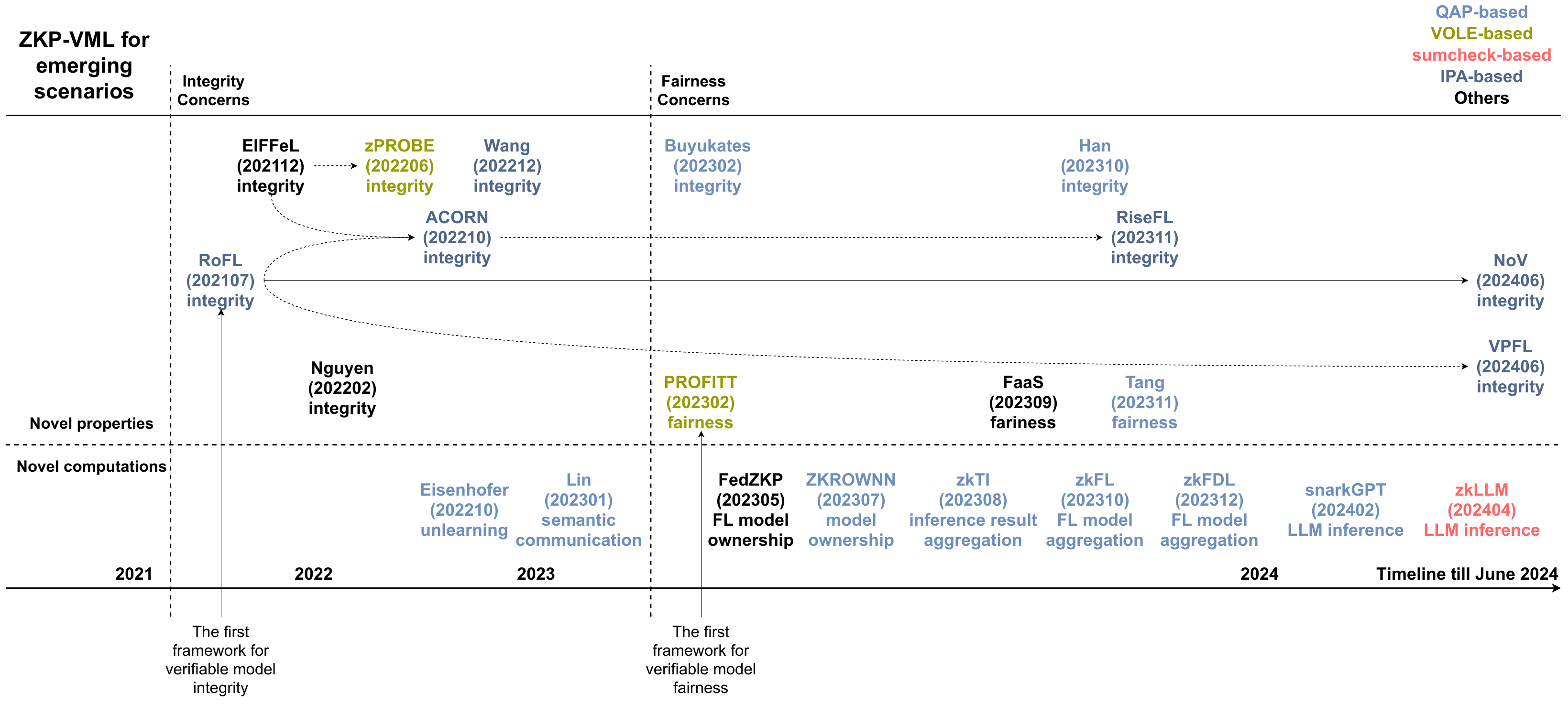}
    \caption{Timeline of existing work in the field of ZKP-VML for emerging scenarios and computations. 
    The dash line from scheme A to B indicates that B is shown to be more advanced than A under certain conditions by theoretical or experimental analysis. 
    The solid line from schemes A to B indicates that B is inspired by A. 
    For brevity, we omit the dash line from A to C, when A points to both B, C and B points to C. \label{timeline2}}
\end{figure*}

In order to visualize the research history of ZKP-VML, we have organized the existing schemes of each category into a timeline according to their initial appearance, as shown in Fig.~\ref{timeline} and~\ref{timeline2}, respectively. 
We follow the time when the scheme was first made public, rather than the time when it was formally published by a conference or journal. 

In 2017, Ghodsi et al. first propose the concept and scheme Safetynets~\cite{ghodsi2017safetynets} for verifiable machine learning. 
Although Keuffer et al.~\cite{keuffer2018efficient} and Zhao et al.~\cite{zhao2021veriml} apply zero-knowledge proofs to verify the correctness of machine learning computations, the assumption that the verifier owns both the model parameters and the training dataset led to the scenario with limited privacy. 
The privacy in VML is not considered till 2020, the presence of zkDT~\cite{zhang2020zero}, where the model parameters are private to the prover. 

QAP-based schemes are widely used for constructing ZKP-VML systems due to their applicability in transforming any additive or multiplicative computation into equivalent arithmetic circuits, simplifying the proof generation process for ML computations. However, QAP-based ZKP schemes can be costly under some situations. As a result, recent efforts have explored alternative ZKP backends for ZKP-VML, such as vector-oblivious linear evaluation-based (VOLE) and sumcheck-based schemes.

From 2021, researchers have also begun to focus on properties and computations other than the correctness in training and inference. 
To address the model poisoning issues in federated learning, Lycklama et al.~\cite{lycklama2023rofl} check the model integrity through ZKP to filter out malicious local models without compromising model privacy. 
To ensure that the decisions given by the model are fair, Shamsabadi et al.~\cite{shamsabadi2022confidential} check the model fairness through ZKP to ensure the model satisfies several given fairness definitions without compromising model privacy. 
In addition, ZKP-VML schemes have been proposed for addressing emerging machine learning computations, such as verifiable machine unlearning, model ownership, and large language model inference.

\subsection{Properties}

\begin{table*}[!htbp]
\centering
\caption{Existing Work Related to Zero-Knowledge Proof-based Verifiable Machine Learning for Training and Inference. }
\label{overview}
\resizebox{\textwidth}{!}{
    \begin{threeparttable}
\begin{tabular}[]{|c|c|c|cccc|}
\hline

\multirow{2}{*}{Scheme} & \multirow{2}{*}{ML Scenario} & \multirow{2}{*}{ZKP System} & \multicolumn{4}{c|}{\thead{Properties}} \\

 & & & \thead{Correctness} & \thead{Security} & \thead{Privacy} & \thead{Distinctness} \\ \hline

ZEN~\cite{feng2021zen} & \thead{verifiable inference for NN} & R1CS & \CIRCLE & \CIRCLE & \CIRCLE & \Circle \\ \hline

Mystique~\cite{weng2021mystique} & \thead{verifiable inference for NN} & VOLE-based & \CIRCLE & \CIRCLE & \CIRCLE & \Circle \\ \hline

Hao et al.~\cite{haoscalable} & \thead{verifiable inference for NN} & VOLE-based & \CIRCLE & \CIRCLE & \CIRCLE & \Circle \\ \hline

Lu et al.~\cite{lu2024efficient} & \thead{verifiable inference for NN} & VOLE-based & \CIRCLE & \CIRCLE & \CIRCLE & \Circle \\ \hline \hline\

Safetynets~\cite{ghodsi2017safetynets} & \thead{verifiable inference for NN} & \textbf{--} & \CIRCLE & \CIRCLE & \Circle & \Circle \\ \hline



zkCNN~\cite{liu2021zkcnn} & \thead{verifiable inference for CNN} & sumcheck-based & \CIRCLE & \CIRCLE & \CIRCLE & \Circle \\ \hline

Wu et al.~\cite{wu2024confidential} & \thead{verifiable inference for ML\\with confidential computation} & sumcheck-based & \CIRCLE & \CIRCLE & \CIRCLE & \Circle \\ \hline

VPNNT~\cite{duan2024verifiable} & \thead{verifiable training for NN} & sumcheck-based & \CIRCLE & \CIRCLE & \CIRCLE & \Circle \\ \hline

Kaizen~\cite{abbaszadeh2024zero} & \thead{verifiable training for NN} & sumcheck-based & \CIRCLE & \CIRCLE & \CIRCLE & \Circle \\ \hline

vCNN~\cite{lee2024vcnn} & \thead{verifiable inference for CNN} & QAP-based & \CIRCLE & \CIRCLE & \CIRCLE & \Circle \\ \hline

pvCNN~\cite{weng2023pvcnn} & \thead{verifiable inference for CNN} & QAP-based & \CIRCLE & \CIRCLE & \CIRCLE & \Circle \\ \hline 

ZENO~\cite{feng2024zeno} & \thead{verifiable inference for NN} & QAP-based & \CIRCLE & \CIRCLE & \CIRCLE & \Circle \\ \hline

Fan et al.~\cite{fan2023validating} & \thead{verifiable inference for CNN} & QAP-based & \CIRCLE & ? & \CIRCLE & \Circle \\ \hline

ValidCNN~\cite{fan2024validcnn} & \thead{verifiable inference for CNN} & QAP-based & \CIRCLE & ? & \CIRCLE & \Circle \\ \hline

VeriCNN~\cite{fan2024vericnn} & \thead{verifiable training for CNN} & QAP-based & \CIRCLE & ? & \CIRCLE & \Circle \\ \hline

psvCNN~\cite{fan2024psvcnn} & \thead{verifiable inference for CNN} & QAP-based & \CIRCLE & ? & \CIRCLE & \Circle \\ \hline

Campanelli et al.~\cite{campanelli2024lookup} & \thead{verifiable inference for DT} & QAP-based & \CIRCLE & \CIRCLE & \CIRCLE & \Circle \\ \hline

Singh et al.~\cite{singh2022zero} & \thead{verifiable ML pipeline and\\verifiable inference for DT} & QAP-based & \CIRCLE & \CIRCLE & \CIRCLE & \Circle \\ \hline

\textcolor{black}{South et al.~\cite{south2024verifiable}} & \textcolor{black}{\thead{verifiable inference for ML}} & \textcolor{black}{QAP-based} & \textcolor{black}{\CIRCLE} & \textcolor{black}{\CIRCLE} & \textcolor{black}{\CIRCLE} & \textcolor{black}{\Circle} \\ \hline

\textcolor{black}{Federify~\cite{keshavarzkalhori2023federify}} & \textcolor{black}{\thead{verifiable FL training}} & \textcolor{black}{QAP-based} & \textcolor{black}{\CIRCLE} & \textcolor{black}{\CIRCLE} & \textcolor{black}{\CIRCLE} & \textcolor{black}{\Circle} \\ \hline

ezDPS~\cite{wang2022ezdps} & \thead{verifiable ML pipeline and\\verifiable inference for SVM} & QAP-based & \CIRCLE & \CIRCLE & \CIRCLE & \CIRCLE \\ \hline

Kang et al.~\cite{kang2022scaling} & \thead{verifiable inference for CNN} & Other (Plonk-based) & \CIRCLE & \CIRCLE & \CIRCLE & \Circle \\ \hline

ZKML~\cite{chen2024zkml} & \thead{verifiable inference for NN} & Other (Plonk-based) & \CIRCLE & \CIRCLE & \CIRCLE & \Circle \\ \hline

Keuffer et al.~\cite{keuffer2018efficient} & \thead{verifiable inference for NN} & QAP-based \& sumcheck-based & \CIRCLE & \CIRCLE & \LEFTcircle & \Circle \\ \hline

Attrapadung et al.~\cite{attrapadung2024privacy} & \thead{verifiable inference for CNN\\with confidential computation} & QAP-based \& IPA-based & \CIRCLE & \CIRCLE & \CIRCLE & \Circle \\ \hline

Garg et al.~\cite{garg2023experimenting} & \thead{verifiable training for LogiR} & QAP-based \& MPCitH & \CIRCLE & \CIRCLE & \CIRCLE & \Circle \\ 
\hline

DPproof~\cite{shamsabadi2024confidential} & \thead{verifiable DP training for ML} & VOLE-based & \CIRCLE & \CIRCLE & \CIRCLE & \Circle \\ \hline 

zkDT~\cite{zhang2020zero} & \thead{verifiable inference for DT} & Other (RS codes) & \CIRCLE & \CIRCLE & \CIRCLE & \Circle \\ \hdashline \hdashline

Otti~\cite{angel2022efficient} & \thead{verifiable training for LP, SDP, SGD} & QAP-based & \CIRCLE & \LEFTcircle & \CIRCLE & \Circle \\ \hline

Ruckel et al.~\cite{ruckel2022fairness} & \thead{verifiable training for LR} & QAP-based & \CIRCLE &\LEFTcircle & \CIRCLE & \Circle \\ \hline

VeriML~\cite{zhao2021veriml} & \thead{verifiable training for ML} & QAP-based & \CIRCLE & \LEFTcircle & \LEFTcircle & \Circle \\ \hline

zkMLaaS~\cite{huang2022zkmlaas} & \thead{verifiable training for ML} & QAP-based & \CIRCLE &\LEFTcircle & \LEFTcircle & \Circle \\ \hline

ZKCMP~\cite{zhou2021zero} & \thead{verifiable inference for NN} & QAP-based / sumcheck-based & \CIRCLE & \LEFTcircle & \CIRCLE & \Circle \\ \hline

\end{tabular}
        \begin{tablenotes}
            \footnotesize
            \item[1] $\Circle$ denotes that the requirement is not satisfied; 
            \item[2] $\LEFTcircle$ denotes that the requirement is partially satisfied; 
            \item[3] $\CIRCLE$ denotes that the requirement is fully satisfied;
            \item[4] ? denotes that the satisfaction of this requirement is doubtful. 
            \textcolor{black}{
            \item[5] the double dash line divide the scheme with full security and partial security according to the categorization. 
            }
        \end{tablenotes}
    \end{threeparttable}
}
\end{table*}

\begin{table*}[!htb]
\centering
\caption{Existing Work Related to Zero-Knowledge Proof-based Verifiable Machine Learning for Emerging Scenarios and Computations}
\label{overview2}
\resizebox{\textwidth}{!}{
    \begin{threeparttable}
\begin{tabular}[]{|c|c|c|cccc|}
\hline

\multirow{2}{*}{Scheme} & \multirow{2}{*}{ML Scenario} & \multirow{2}{*}{ZKP System} & \multicolumn{4}{c|}{\thead{Properties}} \\

 & & & \thead{Correctness} & \thead{Security} & \thead{Privacy} & \thead{Distinctness} \\ \hline

RoFL~\cite{lycklama2023rofl} & \thead{local model integrity for FL} & IPA-based & \LEFTcircle & \LEFTcircle & \CIRCLE & \LEFTcircle \\ \hline

ACORN~\cite{bell2023acorn} & \thead{local model integrity for FL} & IPA-based & \LEFTcircle & \LEFTcircle & \CIRCLE & \LEFTcircle \\ \hline

RiseFL~\cite{zhu2023secure} & \thead{local model integrity for FL} & IPA-based & \LEFTcircle & \LEFTcircle & \CIRCLE & \LEFTcircle \\ \hline

NoV~\cite{xing2024no} & \thead{local model integrity for FL} & IPA-based & \LEFTcircle & \LEFTcircle & \CIRCLE & \LEFTcircle \\ \hline

VPFL~\cite{ma2024vpfl} & \thead{local model integrity for FL} & IPA-based & \LEFTcircle & \LEFTcircle & \CIRCLE & \LEFTcircle \\ \hline

Wang et al.~\cite{wang2022privacy} & \thead{local model integrity for FL} & IPA-based & \LEFTcircle & \LEFTcircle & \CIRCLE & \LEFTcircle \\ \hline

EIFFeL~\cite{roy2022eiffel} & \thead{local model integrity for FL} & Other (SNIP) & \LEFTcircle & \LEFTcircle & \CIRCLE & \LEFTcircle \\ \hline

zPROBE~\cite{ghodsi2023zprobe} & \thead{local model integrity for FL} & VOLE-based & \LEFTcircle & \LEFTcircle & \CIRCLE & \LEFTcircle \\ \hline

Nguyen et al.~\cite{nguyen2022preserving} & \thead{local model integrity for FL} & MPCitH & \LEFTcircle & \LEFTcircle & \CIRCLE & \LEFTcircle \\ \hline 

Han et al.~\cite{han2023kick} & \thead{local model integrity for FL} & QAP-based & \LEFTcircle & \LEFTcircle & \LEFTcircle & \LEFTcircle \\ \hline

Buyukates et al.~\cite{buyukates2023proof} & \thead{local model integrity for FL} & QAP-based & \LEFTcircle & ? & \LEFTcircle & \CIRCLE \\ \hline \hline

Tang et al.~\cite{tang2023privacy} & \thead{model fairness for ML} & QAP-based & \CIRCLE & \CIRCLE & \CIRCLE & \LEFTcircle \\ \hline

FaaS~\cite{toreini2023verifiable} & \thead{model fairness for ML} & Other (1-out-of-n ZKP) & \CIRCLE & \Circle & \CIRCLE & \LEFTcircle \\ \hline

PROFITT~\cite{shamsabadi2022confidential} & \thead{model fairness for DT} & VOLE-based & \CIRCLE & \CIRCLE & \CIRCLE & \CIRCLE \\ \hline \hline

zkFL~\cite{wang2024zkfl} & \thead{verifiable model aggregation for FL} & QAP-based & \CIRCLE & \Circle & \LEFTcircle & \Circle \\ \hline

zkFDL~\cite{ahmadi2024zkfdl} & \thead{verifiable model aggregation for FL} & QAP-based & \CIRCLE & \CIRCLE & \Circle & \Circle \\ \hline \hline

zkTI~\cite{liu2023evaluate} & \thead{verifiable result aggregation for crowdsoucing inference} & QAP-based & \CIRCLE & ? & \CIRCLE & \Circle \\ \hline \hline

FedZKP~\cite{yang2023fedzkp} & \thead{verifiable model ownership for FL} & Sigma protocol & \CIRCLE & \CIRCLE & \LEFTcircle & \Circle \\ \hline

ZKROWNN~\cite{sheybani2023zkrownn} & \thead{verifiable privacy model ownership} & QAP-based & \LEFTcircle & \LEFTcircle & \CIRCLE & \Circle \\ \hline \hline

Eisenhofer et al.~\cite{eisenhofer2022verifiable} & \thead{verifiable machine unlearning} & QAP-based & \CIRCLE & \CIRCLE & \CIRCLE & \Circle \\ \hline \hline

zkLLM~\cite{sun2024zkllm} & \thead{verifiable LLM inference} & sumcheck-based & \CIRCLE & \CIRCLE & \CIRCLE & \Circle \\ \hline

snarkGPT~\cite{ganescu2024trust} & \thead{verifiable LLM inference} & QAP-based & \CIRCLE & \CIRCLE & \CIRCLE & \Circle \\ \hline \hline

\textcolor{black}{Lin et al.~\cite{lin2023blockchain}} & \textcolor{black}{\thead{verifiable semantic communication}} & \textcolor{black}{QAP-based} & \textcolor{black}{\CIRCLE} & \textcolor{black}{\CIRCLE} & \textcolor{black}{\CIRCLE} & \textcolor{black}{\Circle} \\ \hline

\end{tabular}
        \begin{tablenotes}
            \footnotesize
            \item[1] $\Circle$ denotes that the requirement is not satisfied; 
            \item[2] $\LEFTcircle$ denotes that the requirement is partially satisfied; 
            \item[3] $\CIRCLE$ denotes that the requirement is fully satisfied;
            \item[4] ? denotes that the satisfaction of this requirement is doubtful. 
        \end{tablenotes}
    \end{threeparttable}
}
\end{table*}

We classify the 55 ZKP-VML schemes into two categories according to their application scenarios. 
The first category consists of 33 schemes on the verifiability of inference and training processes in machine learning, as shown in Table~\ref{overview}, and the second categories consists of 22 schemes on the verifiability of some emerging scenarios and computation processes in machine learning, as shown in Table~\ref{overview2}. 
In this section we focus on analyzing how the properties of ZKP-VML are achieved. 

\textbf{Correctness and Security.} 

\textcolor{black}{
Building on the definition of ZKP-VML provided in the previous section, correctness refers to the ability of an honest prover to be verified correctly, while security refers to the detection of malicious attackers during the verification process. 
Considering attackers with varying capabilities and objectives in different scenarios, we introduce the concept of \textit{partial security}. 
Partial security refers to situations where the scheme does not offer the level of security defined in the previous section to counter all possible threats, but remains secure within the specific context in which it is applied. 
This partial security arises from the adoption of certain optimization techniques, such as sampling. 
Both correctness and security are ensured through the generation of proofs of the computational process using existing ZKP schemes.
}
Thus, under the completeness and soundness of their used ZKP schemes, honest proofs and results will pass the verification, while malicious proofs and results will not. 
\textcolor{black}{
However, some schemes improve the efficiency by applying some approximation and relaxing the original constraints, resulting in varying degrees of security loss. 
For example, Otti and Ruckel et al. employ approximated results with higher computational efficiency in place of the exact computational outcomes, demonstrating that the approximations are close to the actual results. 
This optimization means that the scheme cannot guarantee that the computed results originate from the given input and the prescribed computational process. 
It can only ensure that the results fall within an acceptable range. 
This optimization approach may render the scheme vulnerable to free-rider attacks.
}
VeriML, zkMLaaS, and ZKCMP adopt the sampling strategy. 
By randomly sampling a small portion from the original set for the verification, the efficiency can be improved in several orders of magnitude. 
However, this strategy cannot detect single forged proof, where the malicious participant aiming at submitting a wrong model with a forged proof in only one round of training. 
Besides, several schemes (Fan~\cite{fan2023validating}, ValidCNN~\cite{fan2024validcnn}, VeriCNN~\cite{fan2024vericnn}, psvCNN~\cite{fan2024psvcnn}, Buyukates et al.~\cite{buyukates2023proof}, zkTI~\cite{liu2023evaluate}) compromise the security due to the incorrect implementation of ZKP schemes. 
Groth16~\cite{groth2016size} requires a trusted setup, where the setup step need to be executed by a trusted third party to guarantee the backdoor generated in the setup to be discarded. 
However, in these schemes, the prover runs the setup step. 
With the backdoor generated, the prover can forge valid proofs for any result, threatening the security of the scheme. 
\textcolor{black}{
For schemes that introduce additional properties or computational types into ZKP-VML, their security warrants further discussion. 
Regarding the issue of model integrity, since the underlying principle relies on observations rather than formal proofs, both correct and incorrect samples have a very small probability of being misclassified, which leads to the \textit{partial correctness}. 
This can result in a loss of both correctness and security. 
}
ZKROWNN~\cite{sheybani2023zkrownn} suffers from a similar problem, the watermark embedding and extraction method adopted, DeepSigns~\cite{darvish2019deepsigns}, has no security proof, but experimental verification shows that the method is functional. 
FaaS~\cite{toreini2023verifiable} and zkFL~\cite{wang2024zkfl}, on the other hand, are less secure due to their weak security assumptions. 
In FaaS, participants are assumed to be honest in providing inference results. 
In zkFL, a malicious server can provide incorrect aggregation results through replay attacks, but is not considered in the threat model.

\textbf{Privacy.} 
\textcolor{black}{
According to the definition provided in the previous section, privacy refers to the protection of the local data used by the performer in a machine learning task. 
Given the varying capabilities of adversaries, privacy is also extended to \textit{partial privacy}. 
One of the key advantages of distributed machine learning is its ability to safeguard data privacy, allowing data to be used directly for model training without exposing it. 
Partial privacy means that during the execution of the machine learning task, the local data used is not directly exposed to the requester. 
However, considering attacks such as model reconstruction and membership inference, an attacker may still infer or reconstruct the local dataset used by the performer through the trained model, which could compromise the data privacy. 
In contrast, complete privacy refers to the situation in which, even if an attacker gains access to all the data exchanged during the machine learning task, no additional information about the private data can be obtained.
Most schemes satisfy privacy by generating proofs using existing zero-knowledge proof protocol within a privacy-preserving ML framework. 
}
Zero-knowledge means that the verifier should learn nothing from the prover except for the validity of the statement being proved. 
In other words, any information the verifier gains by interacting with an honest prover could be learned independently without access to the prover. 
However, there are some schemes (Keuffer et al.~\cite{keuffer2018efficient}, VeriML~\cite{zhao2021veriml}, zkMLaaS~\cite{huang2022zkmlaas}) that do not take privacy as an important consideration. 
In their scenarios, the data required for the computation (dataset, model parameters) are provided by the verifier, while the only private data for the prover are the auxiliary inputs (e.g., hyperparameters), thus the privacy they offer is limited. 
For scenarios involving model aggregation, some schemes (Han et al.~\cite{han2023kick}, Buyukates et al.~\cite{buyukates2023proof}, zkFL~\cite{wang2024zkfl}, zkTI~\cite{liu2023evaluate}) require the client to provide the plaintext of its local model. 
Considering potential model reconstruction and membership inference attacks, these scheme also compromise privacy to some extent. 
zkFDL~\cite{ahmadi2024zkfdl}, on the other hand, directly treats all the inputs and outputs of the aggregation as public data. 
Compared to ZKROWNN~\cite{sheybani2023zkrownn}, FedZKP~\cite{yang2023fedzkp} does not consider the privacy of the watermark itself.

\textbf{Distinctness.} 
\textcolor{black}{
Distinctness is a property used to assess the quality of the results produced by the performer. 
Through ZKPs, ZKP-VML schemes can verify whether the results provided by the performer are derived from the promised input through the specified computational process. 
However, machine learning task results often exhibit variations in quality, with models trained on lower-quality data typically contributing less than those trained on higher-quality data. 
Furthermore, adversaries may also degrade the model performance by low-quality but valid training. 
\textit{Distinctness} refers to the ability of the scheme to provide additional evaluation metrics, beyond simple verification, to distinguish the quality of the results presented by the performer. 
}
The majority of existing work does not consider judgment criteria other than the correctness of the computation process, but this is one of the properties that distinguishes machine learning from other computations. 
Considerations of model integrity and fairness enrich the evaluation criteria. 
Furthermore, PROFITT~\cite{shamsabadi2022confidential} proves both the correctness and the fairness of the trained DT model. 
ezDPS~\cite{wang2022ezdps} generates proof of accuracy besides the proof of correctness and Buyukates et al.~\cite{buyukates2023proof} evaluates the contribution of local models besides the model integrity. 

\section{Technique Route Analysis\label{sec5}}

\textcolor{black}{In this section, we aim to provide a systematic analysis of the existing ZKP-VML schemes by categorizing the schemes based on the specific problems they address and the technical routes they adopt. 
In addition, to clarify the contributions and possible shortcomings of the ZKP-VML work covered, each scheme is analyzed in detail. }
The specific categorization is as follows: 
\begin{itemize}
    \item Transforming machine learning to zero-knowledge proofs. 
    \item Improving the efficiency of zero-knowledge proofs. \textcolor{black}{
    \begin{itemize}
        \item with full security . 
        \item with partial security. 
    \end{itemize}}
    \item Introducing additional properties or computations to ZKP-VML. \textcolor{black}{
    \begin{itemize}
        \item by introducing integrity. 
        \item by introducing fairness.
        \item by verifying model aggregation.
        \item by verifying model ownership.
        \item by verifying crowdsourcing inference.
        \item by verifying machine unlearning. 
        \item by verifying LLMs. 
        \item by verifying Semantic Communication. 
    \end{itemize}}
\end{itemize}

\subsection{Transforming Machine Learning to Zero-Knowledge Proofs}

This is the first step in ZKP-VML, which is also known as the arithmetization or quantization problem. 
Specifically, there are two main barriers, namely the \textit{floating-point number} and \textit{non-linear function}. 
Most machine learning computations involve floating-point numbers, while ZKPs typically handle integers in groups. Simply scaling and truncating floating-point numbers to integers for ZKP proofs can lead to significant accuracy loss in the ML computations. Thus the research question becomes \textit{How to prove floating-point computation with zero-knowledge proof systems while minimizing the impact on precision and accuracy.}
Non-linear activation functions provide neural networks with powerful representational ability, making them one of the most commonly used machine learning models, and able to handle most of the machine learning tasks.
Since they are based on arithmetic circuits, most ZKP systems can only represent linear addition and multiplication operations. 
There are several simple ways to bridge this gap, such as using linear activation functions like $y=x^2$, or dividing the non-linear function into pieces so that it is linear within each piece. 
However, the former one will decrease the accuracy of the neural network, and the latter one will introduce additional proof costs. 
Thus the research question becomes \textit{How to prove non-linear computation with zero-knowledge proof systems while minimizing the impact on accuracy and the additional cost.} 

These two questions are faced by almost all ZKP-VML schemes, yet the solution for most of them is to utilize simple or existing schemes. 
For example, in VeriML~\cite{zhao2021veriml}, each input is constrained to have at most $\ell$ bits of decimal points, thus each input can be scaled to integer by simply multiplying $2^\ell$. 
The non-linear computation within the activation function is approximated by using the Remez method with polynomials. 
zkCNN~\cite{liu2021zkcnn} employs the approach proposed by Jacob et al.~\cite{jacob2018quantization}, which involves transforming a real number $a$ into an integer $q$ using a real number $L$ and an integer $Z$ such that $a=L(q-Z)$. 
zkCNN computes ReLU function using bit-decomposition, which increasing the proof cost. 

Next, we will introduce several schemes specifically aiming at transforming ML computations to ZKPs. 
Compared to previous solutions, these schemes not only improve the performance and efficiency, but also consider a wider problem in ML and ZKPs, allowing future ZKP-VML schemes to simply integrate them as modules. 

Feng et al.~\cite{feng2021zen} present ZEN, a compiler to Rank-1 Constraint System (R1CS) constraints. 
ZEN consists of R1CS friendly quantization and stranded encoding of R1CS constraints. 
In this quantization algorithm, ZEN avoids the costly bit-decompositions caused by two complex operations, comparison and division by two optimization methods, sign-bit grouping and remainder-based verification. 
By stranded encoding, ZEN encodes several low-precision unsigned integers in quantized neural networks as finite field elements on an elliptic curve, reducing the number of constraints compared to the previous one-to-one encoding approach. 
Based on these optimization methods, ZEN can reduce the R1CS constraint by 5.43$\sim$22.19$\times$ when generating verifiable neural network inferences compared to a general integer-arithmetic-only neural network baseline~\cite{jacob2018quantization}, which is adopted by zkCNN and other schemes.  

Weng et al.~\cite{weng2021mystique} propose Mystique, an efficient conversion solution between ZKPs and ML, providing efficient conversion between arithmetic/boolean values, committed/authenticated values and fixed-point/floating-point values. 
Mystique is built on the VOLE-based ZKP protocol QuickSilver~\cite{yang2021quicksilver}. 
The conversion between arithmetic/boolean aims at switching circuit types to improve efficiency based on specific computations, especially for the non-linear functions in the ML scenario. 
In the MPC setting, this conversion can be addressed with extend doubly-authenticated bits (edaBits)~\cite{escudero2020improved} by converting authenticated shares between arithmetic and Boolean circuits. 
Mystique expands this method into zero-knowledge manner, constructing zk-edaBits. 
The second conversion between committed/authenticated values aims at allowing publicly committed data to be simply used in zero-knowledge proof scheme, providing more convenient privacy support. 
The third conversion between fixed-point/floating-point values aims at solving the inconsistency between floating-point numbers used in machine learning algorithms and fixed-point number used in cryptographic algorithms. 
To achieve this, Mystique designs a pair of encoding and decoding methods for supporting IEEE-754 single-precision floating-point number. 
In addition, the scheme is also optimized for matrix multiplication in terms of reducing proof cost. 
It is remarkable that the above solutions are integrated into Rosetta~\cite{Code_Mystique}, a privacy-preserving framework based on TensorFlow~\cite{abadi2016tensorflow}, which means that developers can simply call these interfaces and ignore the cryptographic details involved. 

Hao et al.~\cite{haoscalable} propose a VOLE-based ZKP framework for non-linear functions based on table lookup. 
Traditionally, non-linear functions to be proved have to be converted into Boolean circuit via bit decomposition techniques, causing $O(\log p)$ multiplication complexity in the prime field $\mathbf{F}_p$. 
In this framework, a public table that stores all input-output pairs of the non-linear function is pre-computed by both prover and verifier. 
To reduce the cost caused by large table size, ZKP for read-only memory access (ROM)~\cite{yang2023two} protocol is used to prove that a value read from a committed memory table. 
Besides, proofs for comparison and truncation are constructed as building blocks of the protocol for the transformation of non-linear operations including exponential, division and square root. 
Ultimately, non-linear functions commonly used in machine learning, such as ReLU and Softmax, can be built and proved on these operations. 
The floating-point number is handled using the conversion method proposed by Mystique~\cite{weng2021mystique}. 
Compared to Mystique~\cite{weng2021mystique}, Hao's scheme achieves 50$\times$ to 179$\times$ runtime improvement. 

Lu et al.~\cite{lu2024efficient} raise a ZKP framework for neural network based on efficient proofs for non-linear layers. 
Compared to previous works representing non-linear layers with the costly bit decomposition, they convert the non-linear relations into range and exponent relations, reducing the number of constraints in the circuit. 
Lu et al. designed efficient VOLE-based~\cite{weng2021wolverine} range proof and lookup proof to constrain the primitive operations for non-linear layers, including max, sign, right shift, round and other operations. 
Thus the non-linear layers in CNN and TF including ReLU, MaxPooling, Softmax, GELU can be proved by these primitive constraints. 
The floating-point number is handled using the conversion method proposed by Gholamiet al.~\cite{gholami2022survey}. 
Experiments show that both the range proof and lookup proof outperform existing schemes~\cite{bunz2018bulletproofs,setty2024unlocking,gabizon2020plookup}. 
Compared to existing work such as Mystique~\cite{weng2021mystique}, Hao et al.~\cite{haoscalable}, zkCNN~\cite{liu2021zkcnn} and ZKML~\cite{chen2024zkml}, this scheme performs better on different kinds of non-linear layers and convolutional and transformer neural networks, including the GPT-2~\cite{radford2019language} with 117 million parameters. 

\textbf{Sum up.} 
For the floating-point conversion problem, the general idea is about how to map floating-point numbers with high accuracy into a smaller range. 
ZEN compresses the mapping range by strand encoding, while Mystique handles IEEE-754 floating-point numbers, which improves the ease of use of the solution.
For non-linear operation problems, a common optimization idea is to describe the non-linear operation with simple operations with smaller proof cost, such as range proofs, table lookup proofs, etc. 
Mystique optimizes the conversion between arithmetic circuits and Boolean circuits to reduce the proof cost of bit-decomposition. 
In addition, more schemes utilize the VOLE-based ZKP system~\cite{yang2021quicksilver,weng2021wolverine} due to their efficiency and suitability for constructing the required proofs, such as table lookup and range proofs.

\subsection{Improving the Efficiency of Zero-Knowledge Proofs}

Considering the additional computational cost introduced by ZKP, in order to make the ZKP-VML scheme more practical, many works have focused on how to improve the efficiency and reduce the cost of proofs. 
\textcolor{black}{We categorize these schemes into two types, \textit{improvements with full security} and \textit{improvements with partial security}. }
The former one does not harm the soundness of the original ZKP system.
The latter one adds a non-negligible probability for the adversary, providing additional possibilities for invalid proofs to pass the verification. 
\textcolor{black}{Furthermore, to provide a more systematic analysis from a technical perspective, we will conduct a detailed analysis and presentation of the existing schemes based on the types of ZKPs employed and the specific application scenarios.}

\subsubsection{With Full Security}
\textcolor{black}{\textbf{For schemes adopting the sumcheck protocol, }}
Ghodsi et al.~\cite{ghodsi2017safetynets} raise SafetyNets, an interactive proof protocol for verifiable execution of a class of neural networks.
It is worth mentioning that SafetyNets is the first scheme on VML, although the concept of zero knowledge is not involved in this scheme. 
In SafetyNets, both the inputs and the model are given by the verifier, so there is no privacy issue. 
In SafetyNets, the efficient verification of neural network computation is achieved by designing GKR~\cite{goldwasser2015delegating} protocols for matrix multiplication. 
By randomly selecting a point $C_{i,j}$ in the matrix $C$ and representing its computation process in the form of sumcheck protocol (i.e. $C_{i,j} = \Sigma_{k\in \{0, 1, \ldots, n\}} A[i,k]B[k,j]$), a protocol with computational complexity $O(n^2)$ for both the prover and verifier can be obtained
But unfortunately, for activation functions and pooling layers, SafetyNets can only support specific quadratic activation functions and sum pooling, making it not practical enough. 
However, it still provides an idea to efficient verification by constructing GKR protocol for specific computations. 

Liu et al.~\cite{liu2021zkcnn} propose zkCNN, a zero knowledge proof scheme for convolutional neural networks based on GKR protocol~\cite{goldwasser2015delegating}.
It is worth mentioning that GKR protocol is a non zero-knowledge interactive protocol, but can be converted to zero-knowledge by using a zero-knowledge polynomial commitment~\cite{wahby2018doubly} and non-interactive by the Fiat-Shamir heuristic~\cite{fiat1986prove}. 
zkCNN improves the efficiency by optimizing the proof cost of convolutional computation in CNN.
Liu proposed a new GKR protocol for checking the computation of the fast fourier transform (FFT). 
FFT can be used to improve the computational efficiency of the convolution, thus verifying the convolutional computation process using the FFT is faster than directly verifying the original one. 
Further, by verifying the original convolution in the form of $\overline{U}=\overline{X}*\overline{W}=IFFT(FFT(\overline{X})\odot FFT(\overline{W}))$ with sumcheck protocol for FFT and Inverse FFT (IFFT), the overall prover time can be reduced to surprisingly $O(n^2)$, which is even faster than computing the convolution, with $O(\log^2n)$ proof size and verifier time. 
In addition, to improve the performance of GKR on CNN, they also proposed several improvements and generalizations. 
According to the experiment, compared to vCNN~\cite{lee2024vcnn} and ZEN~\cite{feng2021zen}, zkCNN is 11.2$\times$ faster and 213$\times$ faster on LeNet, respectively. 

Wu et al.~\cite{wu2024confidential} introduce a confidential and verifiable delegation scheme based on ZKP and MPC. 
In this scheme, the verifier has all the machine learning models and data samples, and these data are private to the verifier. 
By using a secret sharing scheme, the verifier can distribute shares of these data to several prover, for their confidential computation through a MPC protocol. 
To prove the correctness of the computation on the server, Wu proposed GKR protocols tailored for MPC matrix multiplication over shared values, which improves its efficiency on neural networks. 
Compared to another ZKP scheme for distributed secrets~\cite{ozdemir2022experimenting}, the performance of Wu’s protocol is 88$\times$ faster on 3-layer MLP network and 74.8$\times$ faster on LeNet. 

Duan et al.~\cite{duan2024verifiable} propose VPNNT, a ZKP framework for verifiable neural network training based on sumcheck protocol. 
Duan et al. introduce custom gates for several binary and unary matrix operations for neural networks. 
The non-linear layers are represented by R1CS and further converted into custom gates as well. 
Thus the computation of the training process can be expressed in the matrix form and proved by these ZKP building blocks for matrix operations. 
To efficiently verifying the claims containing the same structure, Duan et al. combine multiple claims of the same matrix using the multi-linear extension. 
By introducing randomness into these claims, the prover can prove multiple claims at once within an acceptable soundness error. 
Compared to ZKP system Virgo~\cite{zhang2020transparent} and CNN-specialized ZKP system zkCNN~\cite{liu2021zkcnn}, the prover time of VPNNT is 1.16$\times$ to 158.4 $\times$ faster. 

Abbaszadeh et al.~\cite{abbaszadeh2024zero} present Kaizen, a framework for verifiable training on deep neural network. 
Traditionally, proofs are required for each round of training, where the proof size is linear to the number of training rounds, leading to the large proof size. 
To overcome this, Kaizen leverages recursive proof composition, also refers to incrementally verifiable computation (IVC), where the proof for each round training arguing for the current correct computation and a valid proof arguing for the previous correct computation. 
Thus the proof size will be independent to the training rounds.
To efficiently handle the training process, Abbaszadeh proposed sumcheck proofs for gradient descent, where both the linear and non-linear operations are processed following existing methods. 
To build the IVC on the proofs of gradient descent, an aggregation scheme for multivariate polynomial commitments is designed to reduce the cost on verifying the polynomial commitments. 
To further improve the performance, several sumcheck-specific optimizations are also deployed. 
Compared to other IVC schemes including Fractal~\cite{chiesa2020fractal}, Halo~\cite{bowe2019recursive} and Nova~\cite{kothapalli2022nova}, Kaizen achieves 43$\times$ faster prover time and 224$\times$ less prover memory usage. 

\textcolor{black}{
\textbf{Sum up.}
The sumcheck protocol is an interactive proof protocol in which, during the final round of verification, the verifier accesses the value of the target polynomial at the challenge point. 
As such, the protocol is not inherently zero-knowledge. 
However, by employing a commitment scheme to protect the target polynomial, additional information leakage can be prevented, thus achieving zero-knowledge properties. 
Therefore, when using the sumcheck protocol as a proof protocol, it is crucial to ensure that the scheme maintains the desired zero-knowledge guarantees. 
Such protocols primarily focus on how to transform the target computation into a form that is verifiable via the sumcheck protocol, and then efficiently verify the target computation.
}

\textcolor{black}{\textbf{For schemes adopting the QAP-based protocol, }}
Lee et al.~\cite{lee2024vcnn} introduce vCNN, a framework for verifiable convolutional neural network based on zk-SNARKs.
Lee extends the original quadratic arithmetic program (QAP) to quadratic polynomial program (QPP), and constructs the QPP-based zk-SNARKs to prove the convolutional computation.
QPP-based zk-SNARKs involve assigning a polynomial value to each wire, allowing for the expression of 1-D convolution computation using a single multi-gate in the arithmetic circuit. 
\textcolor{black}{As a result, the number of multi-gates required for proving convolution between two matrices $X$ and $W$ of size $n\times n$ and $w\times w$ is reduced from $O(n^2w^2)$ to $O(n^2+w^2)$. }
For the pooling and activation layers, vCNN still retains the QAP-based zk-SNARK. 
And the proofs about the continuity between the two proofs are generated by the commit-and-prove SNARK (CP-SNARK) to connect the adjacent layers.
In this way, the QAP-based and QPP-based zk-SNARKs proves the correctness of the intra-layer computation, while the CP-SNARK proves the continuity of the computation of each layer, and finally the vCNN proves the correctness of the entire convolutional neural network computation.
Theoretically, vCNN has a certain improvement in proving time compared to schemes such as SafetyNets~\cite{ghodsi2017safetynets}, VeriML~\cite{zhao2021veriml}, and Embedded proof~\cite{keuffer2018efficient}. 
Meanwhile, experimental results show that vCNN is 18000 times more efficient on the VGG16 model compared to the original zero-knowledge proof scheme Groth16~\cite{groth2016size}, the latter of which takes more than a decade to generate a proof.

Inspired by vCNN, Weng et al.~\cite{weng2023pvcnn} further propose pvCNN, also a framework for verifiable convolutional neural network.
The main innovation of this paper is the circuit representation method, which proposes a zk-SNARKs scheme based on quadratic matrix program (QMP) based on the QPP-based method proposed by vCNN.
By further expanding the representation capability of the wire in the circuit from array to matrix, pvCNN reduces the size of the circuit by reducing the number of multiplication gates in convolutional operation, thereby improving efficiency. 
In addition, since the neural network is layered, multiple proofs for different inputs of the same CNN layer can be aggregated into one proof with SnarkPack~\cite{gailly2022snarkpack}. 
vCNN split the neural network into prior and later parts. 
The prior net is locally computed with HE to protect the data privacy, while the later net is delegated to be computed in plaintext. 
Thus the computational overhead of fully homomorphic encryption and privacy requirements can be balanced. 
In terms of performance, the scheme is compared theoretically with SafetyNets, zkCNNs, vCNNs, etc., and outperforms the above schemes in terms of proving time for convolution.
And the experimental results also show that QMP-based zk-SNARKs has higher efficiency than the QAP-based for convolution operations. 

Feng et al.~\cite{feng2024zeno} design ZENO, an optimizer for zk-SNARK-based verifiable neural network inference. 
Traditionally, zk-SNARK protocols are designed for scalar type, which makes it complicated to implement zk-SNARK to NN with intensive tensor computations. 
Thus Feng et al. proposed ZENO circuit as an efficient intermediate representation between NN layers and constraints. 
ZENO circuit can minimize the number of addition gates for dot product by aggregating these addition gate into one, which reduces the computational complexity from $O(n^2)$ to $O(n)$. 
Based on this design, ZENO circuit for fully connected, convolution and pooling layers can be extended. 
Another key insight is that the privacy of data in zk-SNARK can be further exploited to reduce the number of constraints.
For example, the product of a public tensor and a private tensor of length $n$ requires $n+1$ constraints.
Whereas if the public tensor is considered as the coefficients in a linear combination, only $1$ constraint is required. 
Such either private feature or private weights situation is common in zero-knowledge proof-based machine learning. 
Further, several optimization methods based on parallel workload and computation reuse are also proposed. 
Compared to several existing zk-SNARK framework including Arkworks~\cite{arkwork}, Bellman~\cite{bellman} and Ginger~\cite{ginger}, ZENO achieves up to 8.5$\times$ end-to-end speedup for NN. 
ZENO can reduce the proof time for VGG16 from 6 minutes to 48 seconds. 

Fan et al.~\cite{fan2023validating} also focus on convolutional computation, converting the computations therein into a simple arithmetic expression in matrix form. 
For the convolution layer, the 3D convolution is represented using a 2D matrix by the im2col method~\cite{chellapilla2006high}, which in turn transforms the convolution calculation into the equivalent matrix multiplication.
The pooling layer also uses the im2col method, which reduces the 3D data to a 2D representation. 
The activation functions ReLU and Softmax are also expressed in the form of matrix multiplication. 
Where ReLU is represented as the input matrix multiplied by a matrix with elements 0 or 1, and Softmax is represented as the output matrix multiplied by a vector of summations of exponents.
Further, all the matrix computation in the convolutional neural network can be optimized by the Freivalds's algorithm~\cite{freivalds1977probabilistic}, which greatly increases the efficiency of setup and proof generation. 

Beside of this, Fan et al. also design three schemes for verifiable CNN training or inference, namely ValidCNN~\cite{fan2024validcnn}, veriCNN~\cite{fan2024vericnn} and psvCNN~\cite{fan2024psvcnn}. 
Similar to their previous work, these three schemes leverage im2col to convert the convolution into matrix multiplication, represent different layers as matrix multiplication and apply Freivalds algorithm to reduce the proof overhead from all these matrix multiplications. 
Specifically, there are some differences between these four schemes. 
veriCNN adopts Winograd~\cite{lavin2016fast} to accelerate the matrix multiplication after im2col conversion and a matrix multiplication verification scheme~\cite{korec2014deterministic} based on Freivalds. 
psvCNN faces a scenario where the server is required to prove the correctness of the prediction performed on a server cluster. 
The computation between each convolutional kernel is independent, thus psvCNN can split the original CNN into independent tasks for parallel execution and proof generation among the cluster. 
However, these four works suffer from a same significant security vulnerability, where the prover runs the setup algorithm of its zk-SNARKs to generate the common reference string for the proof generation and verification. 
It is well known that zk-SNARK schemes, such as Groth16~\cite{groth2016size}, which is widely adopted by Fan et al., requires a trusted setup. 
Because the setup algorithm outputs a trapdoor which can be used to forge proofs for arbitrary statement. 
Thus the setup should be run by another party than the prover, and the trapdoor should be discard once generated. 

Campanelli et al.~\cite{campanelli2024lookup} focus on table lookup proofs, extending the vector lookups to matrix lookups, based on which a zero-knowledge decision tree accuracy scheme is proposed. 
Compared to previous work of lookup proofs cached quotients (cq) ~\cite{eagen2022cq}, Campanelli proposed cq$^+$, which not only brings zero-knowledge to cq, but even introduces no additional prover computation with shorter proof size. 
With the KZG commitments~\cite{kate2010constant} for matrix, cq$^+$ extends the vector lookups to matrix lookups. 
By encoding decision tree models as matrices, the evaluation of the DT is represented as locating the row containing the correct leaf and the input vector matches all the constraints, which can be proved with the matrix lookup proofs. 
For zero-knowledge DT accuracy, this scheme achieves the improvement that the prover time complexity is independent of the size of the decision tree. 
Compared to zkDT~\cite{zhang2020zero}, this scheme reduces the prover time by one order of magnitude and the verifier time by two order of magnitude. 
The adopted ZKP backend is Lunar~\cite{campanelli2021lunar}. 

Singh et al.~\cite{singh2022zero} present a zk-SNARK-based verifiable scheme for decentralized AI pipelines, containing a privacy-preserving verification scheme for decision tree inference. 
The distributed AI pipeline assigns the different steps of data collating, model training, and using the model to make predictions to independent actors such as data owner, model owner, and model consumer.
Compared with zkDT, this scheme avoids costly hashing operations by changing the way of representing and committing to the decision tree.
And it further reduces the number of multiplication gates in the arithmetic circuit by improving the access method in the arithmetic circuit to reduce the access cost of different operations in the prediction path verification. 
Which also reduces the cost for proving data operations in AI pipeline, such as the inner-join or filter during the data curation. 
For the decision tree inference task, the complexity of the circuit generated by this scheme is ten times better than that of zkDT~\cite{zhang2020zero}. 

\textcolor{black}{
South et al.~\cite{south2024verifiable} proposed a framework for guaranteed inference, which not only verifies the model's performance but also ensures that the inference results are generated by the specified model. 
This framework comprises two key components. 
First, for a given model, the prover is required to generate proofs of the inference results on a test dataset to demonstrate the model’s performance. 
Second, when a user queries the model for inference on a given sample, the prover must generate a proof of the inference process. 
The model is bound to the hash of its weights, ensuring that the user can verify that a high-performance model is used for the inference task. 
To address the additional computational overhead associated with proof generation, the framework adopts a "trust but verify" strategy, whereby the prover submits the proofs after a delay following the inference result. 
However, this strategy does not sufficiently mitigate the computational cost. 
To prevent the potential leakage of the test dataset during the model performance verification and simplify the verification, the prover can aggregate all inference proofs into a single proof. 
To do so, the prover can design a specific circuit, which verifies all the proofs on the test dataset and output some designed metrics, such as different accuracy measures. 
}

\textcolor{black}{
Keshavarzkalhori et al.~\cite{keshavarzkalhori2023federify} introduced Federify, an on-chain verifiable federated learning framework. 
In this framework, participants can encrypt and generate proofs for their locally trained models using HE and ZKPs, respectively, while servers collaboratively decrypt the aggregated global model. 
The verification and aggregation of local models are carried out through a smart contract deployed on the blockchain. 
However, Federify is constrained by its reliance on the naive Bayes classifier, the large size of proof files, significant gas costs, and the lack of sufficient optimization, which together limit the scalability and practical utility of the framework.
}

Wang et al.~\cite{wang2022ezdps} introduce ezDPS, a zero-knowledge ML pipeline with high accuracy of inference under Spartan~\cite{setty2020spartan}. 
ezDPS is the first considering to generate proofs for the ML pipeline scenario including Discrete Wavelet Transformation (DWT)~\cite{vonesch2007generalized} for preprocessing, Principal Components Analysis (PCA)~\cite{wold1987principal} for feature extraction and Support Vector Machines (SVM)~\cite{chung2003radius} for classification. 
To efficiently represent these algorithms as circuits, ezDPS applies 2 previously proposed gadgets and 4 specially designed gadgets including exponent, greaterthan, maximum/minimum and absolute gadget. 
Further, ezDPS improves the efficiency of proof generation for DWT and PCA through random linear combination. 
To ensure that the inference is reliable, a zero-knowledge proof of accuracy (zkPoA) is generate to guarantee the high performance of the given SVM, arguing whose accuracy is at least $\delta$. 
Experiments are carried on UCR-ECG, LFW, and CIFAR-100 dataset. 
Compared to the Spartan as the baseline, ezDPS achieves up to 1842$\times$ faster proving time. 

\textcolor{black}{
\textbf{Sum up.}
QAP-based ZKP protocols may have potential security vulnerabilities, with improper implementations allowing adversaries to easily forge proofs. 
Some QAP-based protocols are built upon the Common Reference String (CRS) model, and a subset of these rely on a trusted setup, such as the widely used Groth16~\cite{groth2016size} protocol. 
In these protocols, the setup phase not only generates the proving and verification keys but also produces toxic waste, which can be exploited to forge proofs. 
Consequently, the setup must be performed by a trusted party, ensuring that this toxic waste is discarded properly. 
Therefore, the design of such schemes should prevent the prover from conducting the trusted setup alone, as a malicious prover could potentially exploit the toxic waste to forge proofs. 
QAP-based protocols primarily focus on efficiently representing specific computational processes using arithmetic circuit structures. 
For instance, vCNN extends the wire representation in circuits to directly express polynomials, thereby reducing the number of multiplication gates. 
Similarly, the ezDPS protocol designs gadgets to combine and represent specific computations.
}

\textcolor{black}{\textbf{For schemes adopting the Plonk-based protocol, }}
Kang et al.~\cite{kang2022scaling} tailor the ImageNet-scale model MobileNet v2~\cite{sandler2018mobilenetv2} with the zero-knowledge proof scheme halo2~\cite{halo2book} to obtain an ImageNet-scale zk-SNARK circuit.
Verifying the division operation in the circuit is expensive, to solve this problem, two optimizations are applied to the Plonkish arithmetization~\cite{gabizon2019plonk} used by halo2~\cite{halo2book}.
In linear layers (convolutional layers, residual connection layers, fully-connected layers), two custom gates are designed to reduce the cost. 
As for the non-linear layers (ReLU, softmax), the lookup argument is applied to reduce the representation cost of division. 
Compared with existing schemes including ZEN, vCNN, pvCNN, zkCNN, this scheme improves the proving time on MobileNet by at least ten times. 

Chen et al.~\cite{chen2024zkml} propose ZKML, a verifiable inference framework for realistic machine learning models, based on the halo2~\cite{halo2book} proving system. 
To support different operations in ML, ZKML designed several gadgets in four categories, namely shape, arithmetic, pointwise non-linear and specialized operations.
These gadgets can be used to form 43 ML layers, which can be divided into linear, arithmetic, activation layer and softmax. 
Also, an optimizer is also designed for converting the ML model into an optimized circuit layout. 
The optimizer will generate various layouts and select the optimized one by cost estimation. 
The experiment shows that ZKML can support inference on a distilled GPT-2~\cite{sanh2019distilbert} with 1 TB of RAM within 66 minutes. 

\textcolor{black}{
\textbf{Sum up.}
As a relatively novel zero-knowledge proof protocol, Halo2~\cite{halo2book} has gained increasing attention due to its strong computational performance. 
As a result, Halo2 is often employed in scenarios with heavy computational demands, such as tasks involving models with larger parameters.
}

\textcolor{black}{\textbf{For schemes adopting hybrid ZKP protocols, }}
Keuffer et al.~\cite{keuffer2018efficient} propose a hybrid embedded proof scheme for verifiable computation combining GKR protocol and zk-SNARKs. 
The goal is to achieve a balance between efficiency and usability by first processing individual functions with efficient verifiable computation schemes (EVCs, such as the GKR protocol, which are more efficient but can only handle relatively simple computations), and then processing sequences of functions with general purpose verifiable computation schemes (GVCs, such as zk-SNARKs, which are less efficient but can handle more kinds of computations).
A neural network can be thought of as consisting of several functions, where the correctness of the computation of each function is guaranteed by the GKR protocol, which means that several proofs of the GKR protocol are generated. 
And zk-SNARKs not only prove the continuity of inputs and outputs between functions, but also verify the proofs generated by each function.
Eventually, it generates a total proof.
By verifying the total proof, the verifier can know whether all the specific computations proved by GKR protocol pass or not.
Experimentally, it is shown that the embedded proof scheme has twice as good proving time on two-layer neural networks compared to the scheme using only zk-SNARKs.
However, although the scheme claims that it protects the privacy of the provers' inputs, this is irrelevant in the case where the functions and inputs are known to the verifier. 

Attrapadung et al.~\cite{attrapadung2024privacy} raise a confidential and verifiable CNN inference framework that protects the privacy of both the input data and models from different parties. 
There are two parties, namely $P_1$ and $P_2$, where $P_1$ holding the private model parameters and $P_2$ holding the private data. 
They hope to jointly compute an output and generate corresponding proof arguing for its correctness. 
Arithemetic black-box abstraction (ABB)~\cite{damgaard2003universally} allows parties to perform field arithmetic without explicitly knowing the values, which is leveraged to constructing the confidential CNN inference, and realized through SPDZ~\cite{damgaard2012multiparty}, a MPC protocol based on HE. 
A new collaborative zk-SNARK based on Bulletproofs is constructed following the notion of collaborative zk-SNARK~\cite{ozdemir2022experimenting}, allowing parties to jointly generate Bulletproofs without revealing their secrets for arithmetic circuits. 
Although this scheme achieves the input confidentiality, there is still space for efficiency improvement, for its prover time doubles that of the plain Groth16~\cite{groth2016size}.

Garg et al.~\cite{garg2023experimenting} propose an efficient ZKP protocol for logistic regression (LogiR) training. 
In this work, Grag combined zk-SNARKs with MPC-in-the-head (MPCinH) to balance the overhead in proof generation time and proof size. 
To prove the correctness of the training without revealing the input dataset and output model, a MPCinH protocol is employed for the LogiR training. 
However, the MPCinH protocol needs a trusted pre-processing phase to generate correlated randomness and secret shares of data for the training, thus zk-SNARK is employed to prove the honest execution. 
Besides, some checks of views are also implemented by verifying the corresponding zk-SNARK proofs. 
Customized MPC protocol and packed secret sharing scheme can further reduce the communication overhead. 
Finally, the proof size is reduced to $O(N)$, even though the total computation is $O(DN)$, where $N$ is the size of the dataset and $D$ is the size of the sample. 

\textcolor{black}{
\textbf{Sum up.}
The hybrid use of different ZKP protocols arises from the fact that various protocols exhibit different performance when applied to different types of computations. 
For complex computations, specialized ZKP protocols are often insufficient, while more general-purpose protocols can efficiently represent the computation, improving proof efficiency. 
In contrast, for simpler computations, general-purpose protocols may introduce unnecessary computational overhead, while specialized protocols can generate proofs more efficiently, enhancing computational performance. 
When employing multiple ZKP protocols, it is essential to ensure consistency between them to prevent attackers from using incoherent data to forge separate proofs for different protocols.
}


Shamsabadi et al.~\cite{shamsabadi2024confidential} present Confidential-DPproof, a ZKP framework for differentially private training, especially for the DP-SGD algorithm~\cite{abadi2016deep}. 
Confidential-DPproof is the first framework for ZKP-based verifiable DP training including the privacy budget $\epsilon$. 
The training dataset is committed using the information-theoretic message authentication codes (IT-MACs)~\cite{franzese2021constant}. 
Considering the randomness involved in the training process, a $\Sigma$-protocol is utilized to generate the unbiased randomness seed. 
Each computation included in the DP-SGD is represented in circuits and encoded and proved by the EMP toolkit~\cite{wang2016emp}. 
Experiment shows that it takes 100 hours for Confidential-DPproof to train a model achieving 91\% accuracy on CIFAR-10 in a DP manner. 

Zhang et al.~\cite{zhang2020zero} propose zkDT, a verifiable zero-knowledge proof scheme for decision tree prediction and accuracy.
For a decision tree model, to verify the output, a prior commitment to the decision tree by the prover is required, and then the prover proves the validity of the prediction path to the verifier.
Whereas converting each comparison on the prediction path into an arithmetic circuit is very expensive, to improve efficiency, the authors reduce the generation cost of proofs by inserting designed sibling nodes on the prediction paths.
And the proof is generated with Aurora~\cite{ben2019aurora}, unlike most other ZKP-VML schemes.

\subsubsection{With Partial Security}
Angel et al.~\cite{angel2022efficient} introduce Otti, a compiler for zkSNARKs that focuses on optimization problems including linear programming (LP), semi-definite programming (SDP), and a broad class of stochastic gradient descent (SGD) instances, which are often used in the training of neural networks. 
Otti can compile programs written in a subset of C that describe optimization problems into rank-1 constraint satisfiability (R1CS). 
Otti's idea is to avoid proving the solving process by proving the optimality of the solution, constructing a non-deterministic checker from the certificate of optimality, and then compiling this checker into R1CS. 
For the LP and SDP problems, Otti proves optimality by using the properties of the primal and dual solutions to the optimization problem. 
For the SGD problem, Otti proves optimality by showing that the gradient at the solution has certain properties.
With the Spartan proof system, Otti can prove the optimality of the solution in zero-knowledge within 100 ms, which is four orders of magnitude faster than existing methods. 

Ruckel et al.~\cite{ruckel2022fairness} design a scheme for zero-knowledge verification of linear regression.
The optimization idea is to use an approximate solution that is less computationally expensive and prove how close that approximate solution is to the true solution with some range proofs, which is derived from DIZK~\cite{wu2018dizk}. 
Compute the inverse of a matrix is relative expensive.
To improve the efficiency of the inverse matrix computation in the model update, the author ensures the correctness of the computation by proving the proximity of the computed inverse matrix to the true inverse matrix through a series of range constraints on the norm. 
Besides, differential privacy is also applied to protect the updated local weights in order to achieve stronger privacy. 
Although the scheme is relatively complete in terms of process, the limitations of the optimization method lead to the narrow application scenarios. 

\textcolor{black}{
\textbf{Sum up.}
These schemes compromise security by relying on approximate results. While they ensure that the outcomes remain within an acceptable range, they cannot verify the correctness of the computational process, leaving them susceptible to free-rider attacks. Consequently, when using approximate results to reduce computational burden, it is essential to assess the security requirements of the application scenario.
}

Zhao et al.~\cite{zhao2021veriml} introduce VeriML, a framework for integrity and fairness in outsourced machine learning.
Since several iterations of the same process are performed during training, VeriML chooses to use several iteration rounds of the training process as a challenge for verification, thus reducing the cost of proving.
By storing the input and output of some iterations in advance during the training process and committing to them, the provers can retrieve to the specified iterations and generate proofs of their computational processes as requested by the verifier.
VeriML supports a total of six machine learning models, including linear regression, support vector machines, and neural networks.
For each kind of model, VeriML also proposes some small optimizations for improving the proving efficiency.
In addition, VeriML uses an on-chain protocol to protect the privacy of the trained models for fair trading of the models.

Huang et al.~\cite{huang2022zkmlaas} raise zkMLaaS, a verifiable scheme for machine learning as a service (MLaaS), which focuses on handling volume issue of input data with the random sampling idea.
Proof costs are proportionally reduced by randomly selecting and challenging committed epochs and iterations, which is similar to VeriML.
As for the convolution operation in CNNs, the optimization idea is similar to Fan et al. The im2col algorithm~\cite{jia2014caffe} is applied to convert the convolution into matrix multiplication and the Freivalds'  algorithm~\cite{motwani1996randomized} is further utilized to reduce the overhead of matrix multiplication.
Compared to simply using zk-SNARKs directly, zkMLaaS saves approximately 273$\times$ the proof generation runtime. 

Zhou et al.~\cite{zhou2021zero} present a zero knowledge contingent model payments (ZKCMP) for training neural networks. 
ZKCMP aims at paying for qualified trained models. 
The seller who possesses the trained model sends the encrypted model to the buyer. 
Buyer sends a test dataset with a threshold for the accuracy, on which seller evaluates their model and generate zero-knowledge proofs, arguing that the accuracy exceeds the given threshold. 
After the buyer confirms the proof is verified, the seller can redeem the payment by sending the valid decryption key. 
The above zero-knowledge proofs can be generated with either zk-SNARKs~\cite{ben2013snarks} or Libra~\cite{xie2019libra}. 
To reduce the proof cost in zk-SNARKs, ZKCMP sampling a portion of the constraints for proof generation and verification. 
To convert the computation of inference and encryption into Libra, several sub-circuits are constructed and combined. 
To guarantee the privacy of the intermediate value between sub-circuits, zkVPD~\cite{zhang2017vsql} is adopted for connection. 
Experiments show that for 10,000 images, zk-SNARKs' proof time is nearly 6000s, while Libra's proof time is nearly 1,000 seconds, although Libra's proof is larger. 

\textcolor{black}{
\textbf{Sum up.}
Similarly, schemes that rely on sampling for verification also compromise security. 
Although they can ensure the correctness of most proofs, they cannot detect a single forged proof. 
This makes these schemes susceptible to single-point attacks. 
Consequently, when using sampling-based verification, careful attention should be given to the security implications within the specific application scenario and possible threats. 
}

\subsection{Introducing Additional Properties or Computations into ZKP-VML} 

\begin{figure*}
    \centering
    \subfigure[ZKP-VML in Model Integrity. Different distribution between benign and malicious model.]{
        \includegraphics[width=0.25\textwidth]{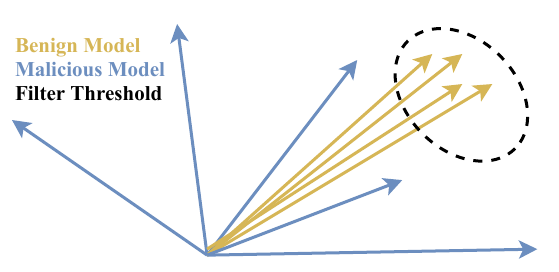}
        \label{figintegrity}
    }
    \subfigure[\textcolor{black}{ZKP-VML in Model Aggregation.}]{
        \includegraphics[width=0.3\textwidth]{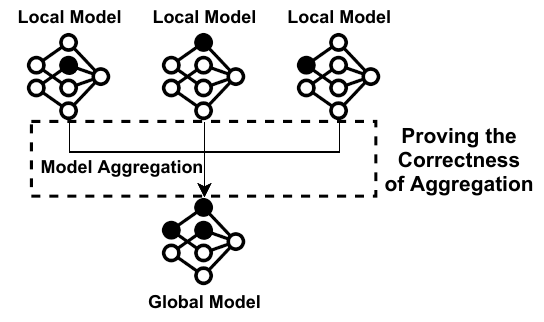}
        \label{figaggregation}
    }
    \subfigure[\textcolor{black}{ZKP-VML in Crowdsourcing Inference.}]{
        \includegraphics[width=0.3\textwidth]{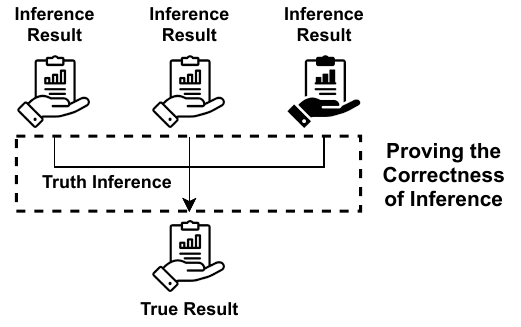}
        \label{figcrowdsourcing}
    }
    \subfigure[\textcolor{black}{ZKP-VML in Model Ownership.}]{
        \includegraphics[width=0.3\textwidth]{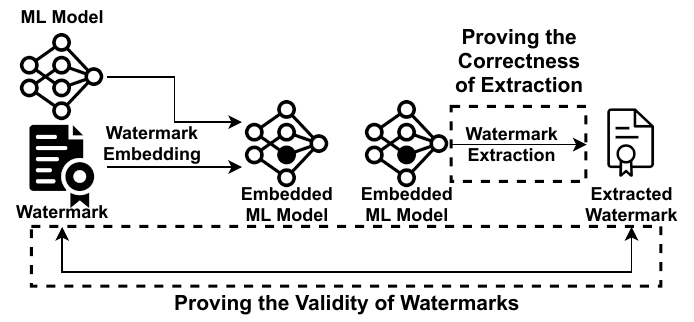}
        \label{figownership}
    }
    \subfigure[\textcolor{black}{ZKP-VML in Machine Unlearning.}]{
        \includegraphics[width=0.3\textwidth]{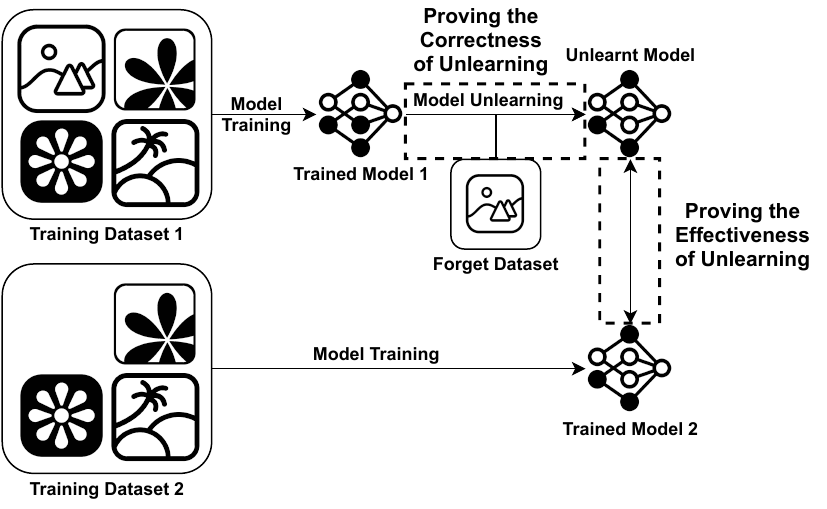}
        \label{figunlearning}
    }
    \subfigure[\textcolor{black}{ZKP-VML in Semantic Communication.}]{
        \includegraphics[width=0.3\textwidth]{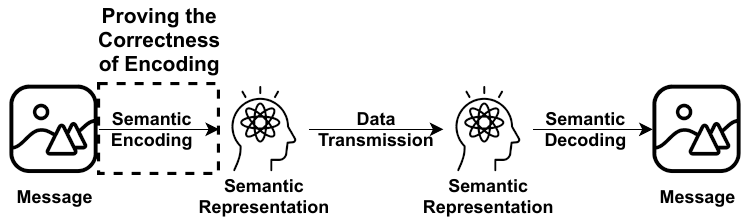}
        \label{figsemantic}
    }
    
    \caption{\textcolor{black}{ZKP-VML in Different Scenarios for Additional Verifiability. }}
    \label{addproperties}
\end{figure*}

\textcolor{black}{
To clarify how ZKP-VML ensures verifiability in these scenarios and its importance, we present ZKP-VML in various scenarios for additional verifiability in Fig.~\ref{addproperties}.
}

\subsubsection{\textcolor{black}{Integrity}}
Integrity refers to whether the model is well-formed or not. 
Usually the integrity of the model can be compromised by model poisoning attacks. 
For example, in federated learning, an attacker may submit a malicious model generated from a poisoning attack to the aggregation, to degrade the performance of the global model. 
A common observation~\cite{cao2020fltrust} is that, if forming the parameters of the model into a vector, then benign models follow a similar distribution of magnitude and direction, while malicious models not, as shown in Fig.~\ref{figintegrity}. 
The malicious models can be identified by abnormal magnitude or direction, thus the integrity of models can be ensured. 
The checking process of the model magnitude or direction can be achieved through zero-knowledge proofs to ensure the privacy of the model and corresponding data.
However, since the checking method is just based on a common observation and there is no proof or theorem to support it, so most of these schemes do not possess provable correctness and security.
The goal of such ZKP-VML schemes is to detect malicious models while protecting the privacy of the submitted models. 
This type of schemes usually requires a corresponding secure aggregation protocols for complete model privacy protection. 

Lycklama et al.~\cite{lycklama2023rofl} propose RoFL, a secure federated learning framework focusing on the robustness. 
RoFL applies adaptive $L_2$ and $L_\infty$ norm to bound the model updates, aiming at identifying malicious updates and ensuring the robustness of the federated learning. 
Provers are required to generate range proofs with Bulletproof~\cite{bunz2018bulletproofs}, arguing that their submitted models meet the norm bound.
In order to reduce the proof burden, RoFL optimizes the proofs for $L_2$ and $L_\infty$ norms, respectively. 
For the $L_\infty$ norm, the verifier randomly selects and verifies a subset of all parameters, for it only takes one failure to identify the attack. 
For the $L_2$ norm, RoFL applies random subspace learning~\cite{li2018measuring} to reduce the size of model parameters and further the proofs. 
RoFL further improves the proof efficiency by applying an optimistic continuation strategy for federated learning.
Where the model verification is run after the model aggregation, thus the verification of round $i$ can be run in parallel with the training of round $i+1$. 
If any verification fails, the server can abort the training and reset the model to the last valid round. 
Although RoFL adds more than ten times the extra computation time compared to basic federated learning, it provides a defense against most data and model poisoning attacks. 

Bell et al.~\cite{bell2023acorn} introduce ACORN, a secure aggregation protocol preserving the integrity of model updates. 
Similar to RoFL, ACORN applies $L_0$, $L_2$ and $L_\infty$ norm bound checks and Bulletproofs for the submitted models. 
Nevertheless, to improve the efficiency of range proofs, ACORN adopts random projection~\cite{gentry2022practical} and several optimizations for range proofs~\cite{groth2005non}. 
Compared to RoFL~\cite{lycklama2023rofl} and EIFFeL~\cite{roy2022eiffel}, ACORN has lower client communication and computation complexity. 

Zhu et al.~\cite{zhu2023secure} raise RiseFL, a robust federated learning system with secure aggregation. 
RiseFL proposes a probabilistic $L_2$ norm check method to detect attacks while improving the efficiency using the Bulletproofs.
Rather than compute the $L_2$ norm of the whole model, RiseFL randomly samples $k$ vectors following a chi-square distribution to compute the $L_2$ norm. 
By using a batch verification strategy, RiseFL reduces the number of group exponential operations from $O(d)$ to $O(d/\log d)$. 
The norm bound check is implemented through Bulletproofs. 
Besides of $L_2$ norm, RiseFL can also support sphere~\cite{steinhardt2017certified}, cosine similarity~\cite{bagdasaryan2020backdoor} and Zeno++~\cite{xie2019zeno++} with simple modifications. 
According to the problem of secure aggregation of verified inputs (SAVI) defined by EIFFeL, RiseFL relaxes the input integrity, while achieving the same level of input privacy. 
In terms of performance, RiseFL is up to 28$\times$, 53$\times$ and 164$\times$ faster with 100K model updates compared to ACORN, RoFL and EIFFeL, respectively. 

Xing et al.~\cite{xing2024no} design NoV, a federated learning framework for fully input integrity for federated learning. 
Inspired by RoFL~\cite{lycklama2023rofl}, NoV applies Bulletproof~\cite{bunz2018bulletproofs} for $L_2$ norm range proofs. 
To further enhance the defense performance against model poisoning attacks, especially for projection gradient descent (PGD) attack~\cite{madry2017towards}, which is one of the strongest first order attack, NoV adopts a model filter combining hybrid and layer-wise strategy. 
For each local model, both of the $L_2$ norm of the whole model and the cosine similarity at each layer are checked. 
In order to provide more robustness for the federated learning, NoV designs a secure aggregation protocol that can  recognize Byzantine attackers within the aggregation process. 
Experiments have demonstrated that NoV has better defense performance against PGD attack compared to FLTrust~\cite{cao2020fltrust}, CosDefense~\cite{yaldiz2023secure} and RoFL~\cite{lycklama2023rofl}.

Ma et al.~\cite{ma2024vpfl} present VPFL, a ZKP scheme for training verifiability by a third party in federated learning. 
VPFL defines a complex system model for federated learning, including five different roles and the bulletin board. 
VPFL leverages Merkle commitment tree for local data integrity check, enabling data owner to check whether their data are correctly stored by the clients. 
The correctness of the training result, which is the local model parameters, is constrained by a predefined range and implemented through Bulletproof~\cite{bunz2018bulletproofs}. 
Compared to several existing schemes for verifiable aggregation and RoFL~\cite{lycklama2023rofl}, VPML performs better in both the computational complexity of the server and client. 

Wang et al.~\cite{wang2022privacy} raise a robust federated learning framework with DP. 
To provide privacy protection in federated learning, each local model will be masked with a Gaussian noise. 
To provide further privacy protection, a mask-based secure aggregation protocol~\cite{bonawitz2017practical} is also introduced. 
Before to be updated to the server, each client is required to clip the local model according to the given $L_2$ norm bound and generate a zero-knowledge range proof with Bulletproof~\cite{bunz2018bulletproofs}, arguing that the $L_\infty$ of the clipped local model falls in the given bound range. 
Each local model is bind with an ElGamal commitment, which is also used to verify the consistent of the mask and compute the aggregated global model. 
The experiment compares the performance of the global model under different parameter settings, but lack of assessment of the efficiency.

Roy et al.~\cite{roy2022eiffel} propose EIFFeL to support arbitrary model integrity checks in federated learning. 
The author formalized the problem as the secure aggregation of verified inputs (SAVI), requiring (1) integrity check of each local update, (2) only well-formed updates are aggregated (3) only the final aggregated result is released.  
EIFFeL employs secret-shared non-interactive proofs (SNIP)~\cite{corrigan2017prio}, a light-weight ZKP system designed for a multi-verifier setting where the private data is distributed or secret-shared among the verifiers. 
In EIFFeL, each client is required to distribute the proof and shares of its private update among clients. 
By verifying and aggregating the valid shares, each client can aggregate the share of the global update. 
With these shares of global update, the server can verify and reconstruct the global update with valid shares. 
To improve the efficiency, EIFFeL replaces the origin share reconstruction method with a robust probabilistic reconstruction based on Reed-Solomon error correcting codes~\cite{shu1983Error}, to improve the performance at the cost of a small probability of failure. 
Further, EIFFeL employes several crypto-engineering optimizations, such as random projection~\cite{nelson2020sketching}, to reduce both the computation and communication costs. 
Compared to RoFL, EIFFeL can support multiple model checking methods, including $L_2$ bound, norm bound~\cite{sun2019can}, Zeno++~\cite{xie2019zeno++} and cosine similarity~\cite{bagdasaryan2020backdoor}. 
Compared to the baseline scheme BREA~\cite{so2020byzantine}, EIFFeL not only provides additional privacy protection, but also improves the efficiency by 2.5$\sim$18.5$\times$. 

Ghodsi et al.~\cite{ghodsi2023zprobe} design zPROBE to defense Byzantine attacks in federated learning. 
The server randomly clusters the clients into several clusters. 
Then each cluster is required to submit a cluster global model through zPROBE's secure aggregation protocol. 
By accessing these cluster model in plaintext, the server can compute and publish the threshold from the median of these models, which is dynamic and can be automatically changed. 
Then each client is required to prove that their masked local model meets the given threshold by using a VOLE-based ZKP system~\cite{weng2021wolverine}. 
Further, to improve the efficiency of zPROBE, a probabilistic optimization is employed to randomly sample some parameters in the local model for the check. 
Experiment shows that zPROBE achieves lower computation complexity compared to both BREA~\cite{so2020byzantine} and EIFFeL~\cite{roy2022eiffel} and higher accuracy compared to EIFFeL. 

Nguyen et al.~\cite{nguyen2022preserving} detect backdoor attacks in federated learning with zero-knowledge proofs. 
A pruning-based backdoor defense method~\cite{liu2018fine} is applied, where a neuron was considered backdoored if pruning a neuron with lower activation had no effect on the main task performance. 
Each client is required to submit the proof of their model passes this detect method. 
The zero-knowledge proof is implemented through a MPC-in-the-head system~\cite{ishai2007zero}. 

Han et al.~\cite{han2023kick} introduce an aggregation protocol with verifiable model detection. 
However, all the clients are required to submit their local model in plaintext to the server, which may lead to privacy attacks. 
The server will execute model detection and aggregation, and generate corresponding zero-knowledge proofs for this process. 
To detect malicious update, this framework deploy cross-round and cross-client detection. 
The cross-round detection checks the cosine similarity between the local model and both the local model and global model in the previous round. 
If any of the similarities are under the given threshold, the cross-client detection will be triggered. 
In the cross-client detection, the $L_2$ distance to the previous global model of each local model is computed as the score. 
The server fits a normal distribution $\mathcal{N}(\mu, \sigma)$ based on the scores of all clients and removes local models with score exceeds $\mu+\lambda\sigma$, where $\lambda$ is a pre-defined threshold. 
To verify the correct execution of the detection mechanism at the server, zk-SNARK proofs, PERCHs~\cite{bitansky2012extractable} are generated by the server, allowing any client to verify the model removal and aggregation.
It is worth mentioning that PERCH does not require a trusted setup, so the proofs generated by server can be verified by any client. 
To reduce the cost of the detection, only the second-to-the-last layer is involved in the detection. 
Compared to existing defense method including m-Krum~\cite{blanchard2017machine}, Foolsgold~\cite{fung2020limitations} and RFA~\cite{pillutla2022robust}, this scheme provides the best defense performance. 
For the CIFAR-100 task running on ResNet56, the proof per round is generated in 100s and verified in 3ms. 

Buyukates et al.~\cite{buyukates2023proof} perform malicious detection and contribution evaluation of local models in federated learning. 
To evaluate the contribution of each local model, a randomized leave-one-out (RLOO) method is proposed. 
Specifically, the contribution of local model $i$ is evaluated by comparing the accuracy between the global model and global model without aggregating local model $i$. 
The larger difference indicates larger contribution. 
The malicious detection method is similar to Han et al.~\cite{han2023kick} and corresponding proofs are generated. 
Therefore this scheme also suffers from the limited privacy issue. 
At the end of the federated learning, the server will evaluate the contribution of each model round-by-round, and generated corresponding proofs. 
All these proof are generated under Groth16~\cite{groth2016size}. 
To improve the proof efficiency, both optimized gadgets for non-linear operations and Freivald's algorithm~\cite{freivalds1977probabilistic} for matrix multiplication verification are introduced. 
For the MLP model, the proof time of contribution evaluation is more than 400s, and the proof time of model detection and aggregation is more than 170s. 
Furthermore, due to the utilization of Groth16, this system requires a trusted setup to guarantee the validity of zero-knowledge proofs, however this is not clearly described.

\textbf{Sum up.} Most of these schemes following the existing detection methods on $L_2$, $L_\infty$ norm and cosine similarity, with zero-knowledge range proofs to check the model integrity without direct access. 
Therefore, the focus of most schemes is not on the detection performance on malicious models, but on the efficiency improvement for range proofs. 
Variously, Nguyen et al. adopts a pruning-based strategy~\cite{liu2018fine} to detect possible malicious and poisoned neurons. 
Meanwhile, because these proofs are centered around the model rather than the computational process, the proof cost is much lower than those checking the training process.
Since the effectiveness of these detection methods is not provable, but based on experimental observations~\cite{cao2020fltrust}, the correctness and security of these schemes can be compromised. 
For example, in practice, a very small percentage of benign and malicious models will be misjudged, although this may not impact the integrity of the final result.
Unlike other schemes, instead of using ZKP schemes for clients to prove the integrity of their model, Han et al.~\cite{han2023kick} and Buyukates et al.~\cite{buyukates2023proof} place the ZKP schemes on the server side, to prove that the detection is correctly carried out. 
Where the privacy is compromised for the server has full access to the model from each client. 

\subsubsection{\textcolor{black}{Fairness}}
Fairness refers to that models should make decisions that are unbiased and equitable across different groups of people. 
There are several key aspects in fairness, including demographic parity~\cite{dwork2008differential}, equalized odds~\cite{hardt2016equality}, equal opportunity~\cite{hardt2016equality}, etc., which can be measured by the evaluation result on specific input dataset. 
Model providers need to demonstrate that their models are fair. 
For example, in the scenario of outsourced inference, the service provider needs to prove that the model it uses for the inference service is fair. 
By generating proofs arguing for both the correctness of the evaluation result and the fairness measurement, the prover can convince the verifier that the provided model is fair. 
The privacy of the evaluation input data can be protected by the zero-knowledge. 

Tang et al.~\cite{tang2023privacy} design a framework for verifiable fairness of machine learning models based on zk-SNARKs. 
In this framework, there are four kinds of fairness definition involved, namely demographic parity, equalized odds, equal opportunity and disparate impact, each of which has a formulation expression. 
In this system, a regulator for fairness is presented. 
This regulator will give out test data, ask the server to provide the evaluation result on that test data, and prove the correctness of that evaluation result by zero-knowledge proof. 
Based on the evaluation result, the regulator audits the model fairness and generate corresponding zero-knowledge proofs for the audition process. 
The fairness measurements are computed based on model statistical metrics, which is computed from confusion metrics. 
The non-linear operations involved in the quantization are represented with bit-decomposition. 
Finally, the client can verify the fairness of the model from public commitments and proofs, without accessing to the plaintext of model or audition dataset. 
The framework adopts SnarkPack~\cite{bunz2021proofs} to aggregate multiple proofs into a single proof to reduce the overhead of the verifier. 
Experiments show that for a LogiR model,  the proving time is in the hundreds of seconds and the verification time is in the milliseconds. 

Toreini et al.~\cite{toreini2023verifiable} propose Fairness-as-a-Service (FaaS), a framework for verifiable fairness in machine learning systems. 
In FaaS, the fairness of models are verified through three fairness metrics for demographic parity, equalized odds and equal opportunity. 
The server is required to provide the encrypted properties of each test data for the fairness evaluation. 
1-out-of-8 zero-knowledge proofs~\cite{cramer1994proofs} are generated for arguing that the encrypted properties are well-formed, rather than the correctness of these properties themselves. 
After receiving the encrypted properties, the regulator can obtain statistical information of data properties and thus determines whether the model that outputs these data is fair. 
However, the server is assumed to be honest, leading to a very limited security and verifiability. 
Experiment shows that for a dataset of size 3166, the whole scheme runs in more than 15 hours. 

Shamsabadi et al.~\cite{shamsabadi2022confidential} present Confidential-PROFITT, a ZKP-friendly decision tree training algorithm as well as the corresponding specialized ZKP protocol, which is derived from a vector-oblivious linear evaluation (VOLE)~\cite{weng2021wolverine}. 
Confidential-PROFITT is the first framework for ZKP-based verifiable DT training. 
Traditionally, the DT is trained by recursively splitting the training dataset. 
To find the best split, Shamsabadi proposed an optimization problem that maximizes the accuracy gain and upper bounds the unfairness gain. 
Rather than proving each training step in the training process, Confidential-PROFITT proves that the given DT model is a valid and fair result on the training dataset, the fairness refers to demographic parity and equalized odds. 
The ZKP protocol is used to prove that with the given committed trained DT model and training dataset, each committed path is correct, the sensitive attribute passes are balance, and the unfairness gain of each node is below the threshold. 
Further, this framework can be expanded to random forests by running a secure coin-flipping before the training. 
Experiment shows that it takes less than 2 minutes for Confidential-PROFITT to prove the fairness of a trained DT. 

\textbf{Sum up.}
The fairness of the model is defined according to several metrics. 
Thus such schemes center around the model, first prove the correctness of the evaluation result on a specific dataset, and then prove that the result satisfies the defined fairness metrics. 
These fairness metrics are usually easy to calculate and do not involve complex non-linear operations, so there is no additional performance enhancement method beyond traditional ZKP-VML schemes for fairness.

\subsubsection{\textcolor{black}{Model Aggregation}}


In federated learning, the local model submitted by clients are aggregated by the server to obtain the global model. 
\textcolor{black}{This type of ZKP-VML schemes aims at guarantee the correctness of the aggregation computation as shown in Fig.~\ref{figaggregation}. }

Wang et al.~\cite{wang2024zkfl} introduce zkFL for verifiable model aggregation in federated learning. 
To guarantee the correctness of average aggregation under a malicious server, each client updates the local model with corresponding Pedersen commitment and signature. 
The server is required to prove the correctness of aggregation alone with the validity of the commitment and signature. 
However, the plaintext of local models are sent to the server for aggregation, leading to the potential privacy leakage. 
Moreover, the security can be harmed by the server reusing previous local model for aggregation. 
Further, the experiment shows that both the time and communication overhead of zkFL are high, making a aggregation scheme not so practical. 
    
Ahmadi et al.~\cite{ahmadi2024zkfdl} propose zkFDL for verifiable aggregation in federated learning. 
However, due to its assumption of public input and output, The utility of the scheme is also in doubt. 

\textbf{Sum up.}
ZKP may not be a proper choice for model aggregation. 
The computations involved in model aggregation contain large number of inputs and outputs with relatively simple computation process, which can be handled with simple additive homomorphic commitment schemes. 
Introducing ZKPs into model aggregation not only increase no additional privacy protection, but also increases unnecessary computational cost, which is unacceptable and unsuitable for federated learning. 

\subsubsection{\textcolor{black}{Crowdsourcing Inference}}


Crowdsourcing inference refers to the scenario where an inference task is outsourced by the user to a cluster of workers with their own models. 
Each worker will generate an inference result with their own model. 
The cluster center will collect these inference results, from which the center can deduce and return a reliable result to the user. 
\textcolor{black}{This type of ZKP-VML scheme aims at arguing that the returned result is indeed computed from some reliable truth inference algorithms as shown in Fig.~\ref{figcrowdsourcing}. }

Liu et al.~\cite{liu2023evaluate} design zkTI for crowdsourcing truth inference. 
In crowdsourcing truth inference, the server is required to distribute the tasks to all clients for their inference. 
After collecting the inference results, the server runs a truth inference algorithm to deduce the true result and evaluate the quality of results from clients.
The truth inference algorithm can prevent clients from submitting malicious results, thus ensuring the quality of the final results. 
To prove that the final result is reliable, the prover convert the truth inference algorithm into circuits, and generate proofs for the computation process with zk-SNARKs. 
Specifically, both Groth16~\cite{groth2016size} and Spartan~\cite{setty2020spartan} can be implemented. 
Experiments show that zkTI shows 1.5-4$\times$ improvement over the baseline~\cite{xu2020catch}, depending on the ZKP backend and parameters. 
But the needed trusted setup for Groth16 is also not clearly described. 

\textbf{Sum up.}
Crowdsourcing inference is an emerging application scenario for ZKP-VML. 
Similar to model aggregation, crowdsourcing inference also involves gathering results and computations among which. 
Differently, crowdsourcing inference works on smaller input and output size with more complicated computations, which broaden the space for applying ZKP. 

\subsubsection{\textcolor{black}{Model Ownership}}


Model ownership refers to the ownership and intellectual property of the model, especially where the value of data and ML models increasing. 
\textcolor{black}{For the ZKP-VML scenario, there are two verification tasks for model ownership as shown in Fig.~\ref{figownership}. 
One task is for the owner to prove the correctness of the watermark extraction process. 
If the watermark extraction process can be proved, then the ownership can be ensured without reveal the extracted watermark. 
Another task is for the owner to prove the validity of a extracted watermark. 
If the owner can prove the validity of the extracted watermark under some fixed commitment, then the ownership can also be proved without leaking the original watermark. 
Once some entity can prove that they indeed own this model by adding a verifiable watermark to the model, then the illegal use of the model can be prevented, and the intellectual property can be protected. }

Yang et al.~\cite{yang2023fedzkp} propose FedZKP for verifying the ownership of the global model in FL. 
Each client generates a pair of private and public input via exact learning parity with noise (xLPN)~\cite{jain2012commitments}, where the public input is sent to the server. 
The server constructs a watermark based on the public input from all clients and sends this watermark to all clients. 
Each client embeds this watermark into the batch normalization (BN) layer during their local training following existing method~\cite{li2022fedipr}. 
When the user wants to verify the watermark in the global model, first the user extracts the watermark from the global model and compare whether the watermark is similar enough to the one constructed from the public inputs of the clients. 
Then the client can generate a zero-knowledge proof that it has a secret corresponding to the public input through the Sigma protocol~\cite{damgaard2002sigma}. 
Experiments prove that FedZKP can effectively defend against fine-tuning~\cite{simonyan2014very} and pruning~\cite{see2016compression} attacks on AlexNet and Resnet18. 
In addition, FedZKP runs with additional time cost in tens of seconds and communication cost in a few MBs.
    
Sheybani et al.~\cite{sheybani2023zkrownn} raise ZKROWNN for verifying the model ownership while preserving the privacy of the watermark. 
The watermarks are embedded and extracted on specific layers following DeepSigns~\cite{darvish2019deepsigns}. 
ZKROWNN leverages zero-knowledge proofs, especially Groth16~\cite{groth2016size}, to prove the correctness of the embedded watermark during the extraction process. 
To generate proofs for the extraction algorithm, ZKROWNN provides different sub-circuits for each computation. 
For example, for convolution, the three-dimensional convolution is reduced to one-dimensional vector. 
The Sigmoid function is represented by a polynomial approximation. 
ReLU and hard thresholding are represented by piecewise functions. 
Besides, Freivald's algorithm~\cite{freivalds1977probabilistic} is introduced to improve the efficiency for proving matrix multiplication. 
Experiments shows that ZKROWNN can complete the proof and verification in tens of seconds and a few KBs of communication. 

\textbf{Sum up.}
Depending on the different types of entities and privacy protection needs, the application of ZKP in model ownership scenarios can be various. 
Multi-entities creates additional verifiability challenges for watermark generation and embedding. 
Whether the privacy protection of the watermark itself is needed is also variable. 

\subsubsection{\textcolor{black}{Machine Unlearning}}


Machine unlearning refers to remove specific training data from the trained model. 
\textcolor{black}{For the ZKP-VML scenario, there are two verification tasks for machine unlearning as shown in Fig.~\ref{figunlearning}. 
One task is for the performer to prove the correctness of the model unlearning process. 
If the model unlearning process can be proved, then the unlearnt model can be ensured without reveal additional information about the data to be forgotten. 
Another task is to prove the effectiveness of an unlearnt model. 
If the unlearnt model can be proved similar enough to a model trained without the data to be forgotten, then the unlearning can also be proved without leaking additional information. }

Common unlearning approaches include retraining the model using a dataset without specific data, and fine-tuning the trained model to remove the effects of specific data. 
The importance of machine unlearning increases with the wide use of ML models and concerns about data privacy.  

Eisenhofer et al.~\cite{eisenhofer2022verifiable} propose a ZKP-based verifiable machine unlearning framework. This iterative scheme supports both verifiable learning and unlearning, where clients can request the addition or removal of data from the training dataset in each round. Learning and unlearning are modeled as separate algorithms, converted into circuits, and verified via zk-SNARKs. The server maintains the system state through hash records of the learning and unlearning datasets to ensure the correctness of client requests. For learning requests, the server proves: 1. Correct computation of the training process. 2. Correct updating of the learning hash record. 3. No modification to the unlearning hash record. For forgetting requests, in addition to proving the correctness of computation and hash maintenance, it is also verified that the deleted data is no longer in the training dataset. The framework uses Spartan for zero-knowledge proofs and Poseidon, a ZKP-friendly hash function. In the experiments, three machine unlearning methods are implemented within this framework, namely, retraining-based, optimization-based~\cite{warnecke2021machine}, and amnesiac~\cite{graves2021amnesiac}. For a linear regression model, proof generation and verification times for one unlearning request range from 1-5 minutes and 0.4-2 minutes, respectively.


\textbf{Sum up.}
Similar to the training process, the unlearning process also involves training data and ML models. 
Nevertheless, the differences in adopted algorithms bring additional optimization possibilities for machine unlearning. 
Besides, not only the unlearning process needs verifiability, the verification of the unlearning effect can also be handled by ZKP in a privacy-preserving manner. 

\subsubsection{\textcolor{black}{Large Language Model}}
Compared to the neural network focused and studied by the majority of ZKP-VML schemes, large language models (LLMs) introduce massive data and computations, as well as additional computation types, such as the attention mechanism and the transformer. 
How to handle the additional computation types and further optimize the proof efficiency for the increased burden are the main concerns of such ZKP-VML schemes. 

Sun et al.~\cite{sun2024zkllm} propose zkLLM, a ZKP-VML framework for LLM inference. 
To address the non-linear and non-arithmetic operations, Sun et al. proposed \textit{tlookup}, a more efficient table lookup proof especially for the tensor-based structure based on sumcheck protocol. 
This method proves that elements of one tensor are contained within another, enabling the verification of non-arithmetic operations such as ReLU, layer normalization, and GELU.
On which \textit{zkAttn} is further built to deal with the attention mechanism and softmax function within. 
zkAttn avoids the division in softmax by translation invariance and decomposes the result of softmax into a summation of several numbers, applying tlookup to prove each. 
For matrix multiplication in attention, zkAttn uses Safetynets~\cite{ghodsi2017safetynets}, and for component-wise proofs, it draws on zkCNN~\cite{liu2021zkcnn} techniques.
Experiments with the OPT, LLaMa-2 models, and the C4 dataset demonstrate that zkLLM supports models 10× larger and achieves proof times 50× faster than Kang~\cite{kang2022scaling}. 
For the 13B-parameter LLaMa-2 model, zkLLM delivers proof in under 15 minutes, with a proof size of 188KB and verification within 5 seconds.

Ganescu et al.~\cite{ganescu2024trust} raise snarkGPT to guarantee that the given output by the LLM is indeed inferred on the specific model. 
To achieve which, snarkGPT generates proofs arguing for the correctness of the inference computation with the commitment of the used LLM. 
To apply the ZKP system to GPT-2, table lookup proof is also adopted to deal with the transformer. 
snarkGPT is implemented with EZKL~\cite{ezkl}, however, the experiment shows a large space for further improvement. 

\textbf{Sum up.} 
The ZKP-VML schemes for LLMs make a further step. 
The proving and optimization methods of ZKP-VML for NNs are not able to meet the massive computation burden in LLM. 
Advanced optimization ideas and methods are necessary. 
Nevertheless, these advanced methods for LLM may also contribute to the ZKP-VML schemes for NNs and other models. 

\subsubsection{\textcolor{black}{Semantic Communication}} 


\textcolor{black}{
Semantic communication is an emerging paradigm focused on transmitting the "meaning" of information rather than raw data. The sender uses a semantic encoder to extract key features from the raw data and convert it into a semantic representation. For example, in image transmission, a deep learning model may extract high-level features. The receiver then demodulates the signal to recover the semantic representation and decodes it into the original data or an approximation. In the context of ZKP-VML, as shown in Fig.~\ref{figsemantic}, the sender must prove that the semantic representation is accurately derived from the original message, ensuring the integrity of the transmitted data.
}

\textcolor{black}{
Lin et al.~\cite{lin2023blockchain} proposed a secure semantic communication framework based on blockchain and ZKP to address the issue where attackers may send malicious data with similar semantic content but different actual data. 
In this framework, edge devices perform spatial transformations (e.g., bilinear interpolation) on semantic data, such as images, before transmitting it to a Virtual Service Provider (VSP). 
These transformations serve to obscure the image and increase the distinction between adversarial and genuine samples in terms of semantic similarity. 
ZKPs are used to record and verify the correctness of these transformations, ensuring that the data remains untampered during transmission. 
The VSP uses smart contracts on the blockchain to validate the legality of the semantic data transformation process. 
In this scheme, ZKPs provide a method for verification without revealing the actual data that has been transformed. 
}

\textcolor{black}{
\textbf{Sum up.} 
The application of ZKP-VML in semantic communication represents a further deep integration of ZKP-VML with communication networks. 
Although research in this area is still limited, this work highlights the potential value of ZKP-VML in the context of semantic communication. 
}

\section{Optimization Method Analysis\label{sec6}}

In this section we summarize and categorize existing optimization methods in ZKP-VML schemes. 
Basically, in a ZKP-VML scheme, the prover generates proofs arguing the correctness of their computational process, and the verifier can check whether the proof is valid. 
Efficiency optimization is a key part of making the ZKP-VML scheme practical.
Optimizations within this process can be broadly categorized into verification-based and generation-based. 
Verification-based optimization is about to changing the verification object, i.e., the proof to be verified is not directly arguing the correctness of the complete computational process. 
Generation-based optimization refers to changing the way that proofs are generated, i.e., designing proof structures or generation algorithms with lower computation and communication costs, especially for a specific target type of computation. 

\subsection{Verification-based Optimization}

Within the verification-based optimization, there are two main types: 1. \textit{Embed} and aggregate the proofs. 2. \textit{Sample} a small portion of the proofs for verification. 

For the first type, since the number of claims to be proved is not reduced, and the proof \textit{embedding} and aggregation are performed by the prover, the optimization will focus more on the communication and computational burden on the verifier. 
Keuffer et al.~\cite{keuffer2018efficient} first generate sumcheck proofs $\pi_1$ arguing for the correctness of the machine learning computational process. 
To embed these proofs, a zk-SNARK proof $\pi_2$ is generated, arguing for the valid verification process of $\pi_1$. 
Thus, by verifying only $\pi_2$, the verifier can check the whole computation, and the communication and verification cost of $\pi_1$ can be reduced. 
Similarly, Garg et al.~\cite{garg2023experimenting} prove the validity of views in the MPCitH proofs by generating zk-SNARK proofs, which also reduces the communication and verification cost of verifying the views. 
Abbaszadeh et al.~\cite{abbaszadeh2024zero} adopt incrementally verifiable computation, where each proof arguing for not only the current round training process, but also the correctness of the previous proof.
Thus there will be only one proof in the end, which optimizes the proof size to be independent to the training rounds. 

For the second type, by \textit{sampling} and verifying a small portion of the proofs, the verifier can still detect a certain percentage of malicious behavior with a very high probability. 
For example, considering there are 30k fraud proofs in 100k proofs, the verifier only need to randomly verify 14 proofs to detect the misbehaviour with the probability higher than 99\%. 
Although the soundness error can be extremely small, it still introduce a non-negligible probability. 
Both Zhao et al.~\cite{zhao2021veriml} and Huang et al.~\cite{huang2022zkmlaas} leverage the random sampling on proofs, i.e., randomly select a small portion of proofs for generation and verification. 
Similarly, Zhou et al.~\cite{zhou2021zero} sample a small portion of the constraints in the computational process for the prover to generate proofs for. 
These sampling significantly reduces the proof generation and verification overhead, but also introduce a non-negligible soundness error. 

\textcolor{black}{
The optimization of embedding and aggregating proofs do not reduce the computational complexity for the prover. 
In fact, they may slightly increase it, as the prover has to do the extra embedding and aggregation. 
However, these techniques significantly alleviate the bandwidth burden, making them more suitable for scenarios with stringent bandwidth requirements. 
Among these methods, the embedding proof and incremental computation verification techniques are particularly easy to migrate and implement to other schemes. 
On the other hand, the approach of verifying a subset of rounds is more applicable to scenarios involving a large number of training iterations. 
By sampling only a small number of rounds for verification, the efficiency of the scheme can be improved by several orders of magnitude. 
However, it is important to note that this sampling optimization may compromise the security of the scheme to some extent, as it renders the system vulnerable to single-round attacks. 
Therefore, this approach is better suited to scenarios where the verification workload is a priority.
}

\subsection{Generation-based Optimization}

For the generation-based optimization, we categorize it into: 1. \textit{Tailor} existing proof protocol to the target computation. 2. \textit{Change} the proof route according to the properties of the target computation. 3. \textit{Prune} the additional privacy burden. 

The majority of existing work \textbf{tailors} the proof protocol for proving the target computation. 
This part can be further categorized by their different target computations. 

For the \textit{matrix multiplication}, we consider a 2-D matrix multiplication between $A$ and $B$ both of size $n\times n$ as $C=A\times B$. 
The complexity of input and output is $O(n^2)$ and the computational complexity is $O(n^3)$. 

Ghodsi et al.~\cite{ghodsi2017safetynets} construct sumcheck protocol for matrix multiplication proof. 
\textcolor{black}{First, the matrix multiplication is transformed into a sumcheck problem through a multilinear extension, and then the correctness of the polynomial at random points is progressively verified using a recursive sum-check protocol. }
By randomly selecting a point $C_{i,j}$ in the matrix $C$ and representing its computation process in the form of sumcheck protocol (i.e. $C_{i,j} = \Sigma_{k\in \{0, 1, \ldots, n\}} A[i,k]B[k,j]$), \textcolor{black}{a protocol with prover complexity of $O(n(n+d))$ and verifier complexity of $O(n^2)$ can be obtained,} while $4\log(n)$ elements need to be exchanged in the process. 
\textcolor{black}{
For the matrix multiplication proved with sumcheck protocol, Duan et al.~\cite{duan2024verifiable} further combine the computations containing the same matrices by multi-linear extension, enabling the prover to prove multiple claims at once with an acceptable soundness error. 
}

Weng et al.~\cite{weng2021mystique} adopt Freivald's algorithm to reduce the proof cost for matrix multiplication. 
Rather than check the equation $A\times B=C$ directly, a random challenge vector $t$ of length $n$ is introduced. 
The original equation can be checked by multiplying a challenge vector as $A\times(B\cdot t)=C\cdot t$, \textcolor{black}{with prover complexity of $O(n^2)$, verifier complexity of $O(n)$, and constant proof size $O(k\log q)$, where $k$ is secure parameter and $q$ is the big prime. }

Wu et al.~\cite{wu2024confidential} construct sumcheck protocol for the matrix multiplication in the MPC protocol to achieve the confidential computation, whose prover time is $O(n^2)$ and verifier time is $O(\log n)$, with $O(\log n)$ proof size. 


For the \textit{matrix convolution}, we consider a 2-D convolution between two matrices $X$ and $W$ of size $n\times n$ and $w\times w$ ($n\gg w$) as a $(n-w+1)\times (n-w+1)$ matrix $U=X*W$.
The complexity of input and output is $O(n^2+w^2)$ and the computational complexity is $O(w^2(n-w+1)^2)=O(w^2(n-w)^2)$.

Liu et al.~\cite{liu2021zkcnn} construct sumcheck protocol for convolution. 
\textcolor{black}{Convolution can be accelerated by FFT, then verifying FFT-based convolution is also more efficient than verifying convolution directly. }
zkCNN designs a sumcheck protocol for verifying the FFT, which is a matrix-vector multiplication of size $n$, to reduce the original prove complexity from $O(n^2)$ to $O(n)$. 
Further, by verifying the original convolution in the form of $\overline{U}=\overline{X}*\overline{W}=IFFT(FFT(\overline{X})\odot FFT(\overline{W}))$ with sumcheck protocol for FFT and Inverse FFT (IFFT), the overall prover time can be reduced to $O(n^2)$, with $O(\log^2n)$ proof size and verifier time.

Lee et al.~\cite{lee2024vcnn} extend the original QAP-based zk-SNARK to QPP-based zk-SNARK. 
This innovative approach involves assigning a polynomial value to each wire, allowing for the expression of 1-D convolution computation using a single multi-gate in the arithmetic circuit. 
As a result, the number of multi-gates required for proving convolution is reduced to $O(n^2+w^2)$.
\textcolor{black}{Built on this, vCNN offers a QPP-based zk-SNARK where the prover complexity is $O(n^2+w^2)$, the verifier complexity is $O(n^2)$, along with a constant size of proof. }

Weng et al.~\cite{weng2023pvcnn} further extend the QPP-based zk-SNARK to QMP-based zk-SNARK, where each wire representing a matrix. 
Compared to the QPP-based, QMP-based zk-SNARK performs better on proving batched convolution. 
\textcolor{black}{For batch convolution operations involving $M$ matrices $W$ of size $w\times w$ and $Mn^2$ matrices $X$ of size $n\times n$, pvCNN transforms them into one matrix multiplication of size $Mn^2\times Mn^2$, which leverages a prover complexity of $O(M^2n^4)$ and verifier complexity of $O(Mn^2)$ with a QMP-based zk-SNARK. 
In contrast, under similar conditions, vCNN has a proof complexity of $O(M\cdot Mn^2\cdot(w^2+n^2))$. }

Feng et al.~\cite{feng2024zeno} proposed ZENO circuit to efficiently represent tensor in zk-SNARKs. 
\textcolor{black}{By leveraging the commutative property of addition gates and minimizing the number of which, the computation complexity of dot product of length $n$ can be reduced from $O(n^2)$ to $O(n)$.}
Thus the prover complexity of convolution is $=O(w(n-w)^2)$, verifier complexity is $O(n^2+w^2)$. 

Both Huang et al.~\cite{huang2022zkmlaas} and Fan et al.~\cite{fan2023validating,fan2024psvcnn,fan2024validcnn,fan2024vericnn} leverage im2col~\cite{jia2014caffe} to convert convolution into matrix multiplication. 
\textcolor{black}{
Where the size of input matrix is $(n-w+1)^2 \times w^2$ and size of kernel matrix is $w^2\times 1$.
The proofs of these matrix multiplications are further optimized through the Freivald's algorithm. 
Thus the prover complexity is $O(w^2(n-w)^2)$, the verifier complexity is $(n-w)^2$, the proof size is constant. }

For the \textit{decision tree}, we consider a binary decision tree $T$ with height $h$, which has $N$ nodes and an attribute set size of $d$.
To prove the inference of a decision tree, it is necessary to demonstrate the existence of a valid prediction path within this decision tree.
For a decision tree with a height of $h$ and nodes containing $d$ attributes, proving a decision path that includes $h$ calculations has a complexity of $O(d)$ per comparison. 
Therefore, the prover computational complexity for inference is $O(hd)$, for accuracy with $n$ samples is $O(nhd)$.

Zhang et al.~\cite{zhang2020zero} insert designed sibling nodes to the DT to reduce the proof cost caused by the comparison. 
The zkDT reduce the proof complexity to $O(d+h)$ by constructing an additional input permutation $\bar{a}$, which is based on the sorting of attributes used in the decision path. 
Verifying the permutation relationship between $a$ and $\bar{a}$ adds only $O(d)$ extra computational complexity.
Overall, during the commitment phase, the prover needs to perform $O(N)$ hash computations, and the size of the proof circuit is $O(d+h)$. 
With the Aurora system, for DT inference, the computational complexity of the proof generation is $O((d+h) \log(d+h))=O(d \log d)$, the verification complexity is $O(d+h)=O(d)$, and the proof size is $O(\log^2(d+h))$. 
For DT accuracy, the prover computation complexity is $O(nd\log(nd))$ with $O(N)$ hashes, the verification time is $O(nd)$ and proof size of $\log^2(nd)$. 

Singh et al.~\cite{singh2022zero} propose consistent memory access proof for efficient proof of dataset operations and DT models. 
For a batch of $n$ input samples with $d$ dimensions, this scheme reduces the cost of each sample from $O(d \log h)$ in zkDT to $O(d+h)$ through consistent memory access. 
Consequently, the size of the entire proof circuit is reduced to $O(N+n(d+h+wh))$, where $w$ is the bit-width of the feature values. 
In comparison, under the same conditions, the circuit size of zkDT is $O(c(H)N + n(d \log h + hw))$, where $c(H)$ represents the size of the hash circuit. 
For the DT accuracy, the circuit complexity is $O(N+hn(1+w+d))$. 
The prover complexity of backend ZKP is $O(C\log|C|)$, with constant size of proof and $O(|io|)$ verifier complexity, where $C$ represents the circuit and $io$ represents the input and output.

Campanelli et al.~\cite{campanelli2024lookup} proposed efficient matrix lookup proofs for DT accuracy. 
By converting the DT model into a matrix, the proof of DT evaluation can be transformed into proof of locations in matrix and proof of satisfaction of corresponding constraints.
With the efficient matrix lookup proof, the prover complexity can be reduced to $O(nd\log(nd))$, the verifier complexity can be reduced to $O(n)$, with a constant size of proofs. 

For the \textit{non-linear operations} in various neural networks, there are several representation methods. 

One intuitive way is to directly transform the non-linear operations. 
Weng tt al.~\cite{weng2021mystique} reduce the non-linear operation proof cost by providing efficient conversion between arithmetic and boolean circuit to reduce the bit-decomposition cost. 
Duan et al.~\cite{duan2024verifiable} adopt the GINGER~\cite{setty2012taking} to convert the non-linear operations into R1CS and further proved with Spartan~\cite{setty2020spartan}. 

Some researchers also build specialized proofs for non-linear operations. 
Feng et al.~\cite{feng2021zen} optimize the proof cost for both comparison and division operations through sign-bit grouping and remainder-based verification. 
Kang et al.~\cite{kang2022scaling} adopt table lookup for the division in non-linear operations. 
Hao et al.~\cite{haoscalable} reduce the proof cost for non-linear operation cost by providing table lookup proofs. 

Building custom gadgets is also a popular trick. 
Lu et al.~\cite{lu2024efficient} build gadgets based on range proof and table lookup proof for constructing constraints for non-linear layers in CNN and TF. 
Chen et al.~\cite{chen2024zkml} design plenty of gadgets for constructing the whole machine learning model including the non-linear operations. 
Wang et al.~\cite{wang2022ezdps} apply several gadgets to represent the SVM algorithms, including the radial basis function (RBF) kernel method. 

For some special computations, depending on the properties of their computation process or results, it is possible to construct specialized methods for optimizing the proof efficiency by \textbf{changing} the verification route.

Ruckel et al.~\cite{ruckel2022fairness} exploit the properties of inverse matrix in LR. 
Instead of proving the inverse computation, it can be proved that the proximity of the claimed matrix to the true inverse matrix. 
If the approximation error is below the threshold, then the claimed matrix is considered as a valid inverse matrix. 

Angel et al.~\cite{angel2022efficient} exploit the properties of solutions to optimization problems in Otti. 
For LP and SDP problem, Otti proves that the primal solution is a feasible solution to the primal problem, the dual solution is a feasible solution to the dual problem, and the duality gap is zero. 
For SGD problem, Otti proves that at $z$ the gradient estimates $\nabla f_i(z)$ have some property. 
By constructing proof methods checking the optimality of the results, proofs for the complete computational procedure can be eliminated. 

Shamsabadi et al.~\cite{shamsabadi2022confidential} exploit the properties of trained DT models and corresponding training dataset. 
Instead of proving the whole training process, it can be proved that the DT model is balance on the given dataset, thus the given DT model is a valid training result. 

A novel idea is to consider and \textbf{prune} the burden that data privacy imposes on ZKP-VML. 

Weng et al.~\cite{weng2021mystique} consider the conversion between committed and authenticated values and reduce the proof cost incurred by converting publicly committed value to the private authenticated values in the ZKP system through an efficient conversion method. 

Feng et al.~\cite{feng2024zeno} consider the proof cost caused by the excessive privacy. 
In practice, there are a fair number of cases where either the features or weights are private. 
One naive design is to generate one constraint for each multiplication in the dot product of features and weights, which incurs unnecessary constraints.
By designing specific proof patterns, the number of constraints and proof cost for multiplication with unnecessary privacy can be reduced. 

\textcolor{black}{
\textbf{Sum up.} 
Tailoring existing proof protocols to the target computation for specific optimizations represents a primary approach in ZKP-VML optimization. 
This method enables the maximization of computational and communication efficiency with minimal sacrifice in security. 
For matrix multiplication, commonly used optimization techniques include the sumcheck protocol and the application of Freivald's algorithm~\cite{freivalds1977probabilistic}. 
In the case of convolution, one approach is to employ the FFT algorithm to accelerate the convolution process, followed by the use of ZKPs to verify the accelerated computation. 
Another approach is to optimize the matrix representation within ZKPs to reduce the number of multiplication gates, thus enhancing the efficiency. 
However, while transforming convolution into a matrix form using im2col~\cite{jia2014caffe} and Freivald's algorithm allows for further optimization, this transformation introduces additional computational overhead, limiting the overall performance improvement. 
For decision trees, the primary focus of optimization lies in how to represent a decision tree in a proof-friendly manner. 
For nonlinear operations, a common optimization strategy is to construct basic gadgets and then combine these gadgets to represent the nonlinear operations, thereby minimizing the accuracy loss caused by approximation techniques. 
Optimizations based on altering the verification route are more dependent on the specific computation, making them difficult to transfer to other computations. 
Pruning the additional computational burden introduced by data privacy protection presents an interesting approach. 
ZKPs offer robust privacy protection. 
However, such protection may not be necessary for certain application scenarios. 
By relaxing this privacy requirement, computational overhead can be fine-tuned to optimize unnecessary operations.}

\textcolor{black}{
To provide a more intuitive comparison of the improvements in computational and communication efficiency brought about by various optimization strategies, we selected a subset of these methods and compared their performance in Table \ref{E&C}. 
Specifically, we focused on problems that can be quantitatively assessed, such as matrix multiplication, convolution, and decision tree accuracy. 
These problems exhibit clear computational complexity and input-output scale. 
In contrast, non-linear operations, even when considering specific computations, lack intuitive computational complexity, making it challenging to compare the performance improvement of different optimization methods. 
Furthermore, we selected optimization methods that allow for quantitative comparison, such as those based on tailored approaches. 
The performance of embedding methods depends on the selected proof scheme, while sample-based methods are influenced by the number of verification rounds chosen by the verifier. 
The effectiveness of changing and pruning methods depends on the optimization problem at hand. 
Since the performance improvements of the aforementioned methods typically rely on the specific problem context, it is difficult to provide a straightforward analysis or comparison. 
Therefore, we have excluded these methods from our comparative analysis. 
}

\begin{table*}[htbp]
    \centering
    \textcolor{black}{\caption{Computational \& Communication Complexity on Different Optimizations}}
    \label{E&C}
    \begin{tabular}{|c|c|c|c|c|}
    \hline
        \thead{Scheme} & \thead{Optimization\\Method} & \thead{Targeted\\Computation} & \thead{Efficiency\\(Prover / Verifier)} & \thead{Communication\\(Proof Size)}\\ \hline
        
        \thead{SafetyNets~\cite{ghodsi2017safetynets}, VPNNT~\cite{duan2024verifiable}} & \thead{Tailored sumcheck protocol} & \multirow{4}{*}{\thead{Matrix\\Multiplication\\$O(n^3)$}} & $O(n(n+d))/O(n^2)$ & $O(\log n)$ \\ \cline{1-2} \cline{4-5}
        \thead{Wu et al.~\cite{wu2024confidential}} & \thead{Tailored sumcheck protocol \& MPC} & & $O(n^2)/O(\log n)$ & $O(\log n)$ \\ \cline{1-2} \cline{4-5}
        \thead{Mystique~\cite{weng2021mystique},\\Buyukates et al.~\cite{buyukates2023proof}} & \thead{Freivalds~\cite{freivalds1977probabilistic}} &  & $O(n^2)/O(n)$ & $O(1)$ \\ \cline{1-5} 
        
        \thead{zkCNN~\cite{liu2021zkcnn}} & \thead{Tailored sumcheck protocol \& FFT} & \multirow{5}{*}{\thead{Convolution\\$O(w^2(n-w)^2)$}} &  \thead{$O(n^2)/O(\log^2n)$} & $O(\log^2n)$ \\ \cline{1-2} \cline{4-5} 
        \thead{vCNN~\cite{lee2024vcnn}} & \thead{Tailored representation of vector} &  & $O(n^2+w^2)/O(n^2)$ & $O(1)$ \\ \cline{1-2} \cline{4-5}  
        \thead{pvCNN~\cite{weng2023pvcnn}} & \thead{Tailored representation of matrix} & & $O(n^2+w^2)/O(n^2)$ & $O(1)$ \\ \cline{1-2} \cline{4-5} 
        \thead{ZENO~\cite{feng2024zeno}} & \thead{Tailored representation of tensor} & & $O(w(n-w)^2)/O(n^2)$ & $O(1)$ \\ \cline{1-2} \cline{4-5} 
        \thead{Fan et al.~\cite{fan2023validating,fan2024psvcnn,fan2024validcnn,fan2024vericnn},\\zkMLaaS~\cite{huang2022zkmlaas}} & \thead{Freivalds~\cite{freivalds1977probabilistic} \& im2col~\cite{jia2014caffe}} &  & $O(w^2(n-w)^2)/O(n^2)$ & $O(1)$ \\ \cline{1-5} 
        
        \thead{zkDT~\cite{zhang2020zero}} & \thead{Tailored representation of trees} & \multirow{3}{*}{\thead{Decision Tree\\$O(nhd)$}} & \thead{$O(nd\log nd)+O(N)/O(nd)$} & $O(\log^2nd)$ \\ \cline{1-2} \cline{4-5}
        \thead{Singh et al.~\cite{singh2022zero}} & \thead{Tailored representation of trees \\ \& 
        consistent memory access} & & \thead{$O(nd\log nd)/O(nd)$} & $O(1)$ \\ \cline{1-2} \cline{4-5}
        \thead{Campanelli et al.~\cite{campanelli2024lookup}} & \thead{Tailored representation of trees\\ \& table lookup} & & \thead{$O(nd\log nd)/O(n)$} & $O(1)$ \\ \hline 
        
    \end{tabular}
\end{table*}

\section{Challenges and Future Research Directions\label{sec7}}

\textcolor{black}{Based on the analysis of existing work, in this section, we explore the challenges and future research directions of ZKP-VML in communication networks. }

The main development direction of ZKP-VML revolves around enabling participants to verify ML computations in a zero-knowledge manner with minimal additional costs. 
This goal encompasses two key aspects: 
(1) enhancing the practical applicability of ZKP-VML and (2) enriching its content and properties. 
The limitations to the practical applicability of ZKP-VML primarily stem from two factors.

First, the verifiability offered by ZKPs is achieved at the expense of additional computational costs. 
Therefore, to make ZKP-VML more practical, improving the computational efficiency of proofs is essential, minimizing the overhead between ZKP-VML and existing ML frameworks. 
Additionally, the lack of development in user-friendly toolkits also degrades the adoption of ZKP-VML. 
As machine learning continues to evolve, numerous emerging ML scenarios and computational paradigms have arisen. 
To broadly incorporate verifiability in ML, supporting these new scenarios and computational types in ZKP-VML is crucial. 
Furthermore, integrating ZKPs with other cryptographic techniques can further enrich security attributes and properties in ML.

\subsection{Computational Efficiency}

Enhancing efficiency is a crucial research direction in ZKP-VML, as reducing the computational burden of ZKPs can significantly improve their practicality. There are three key avenues to advance the efficiency of ZKP-VML schemes:


1) \textbf{Proof System Design:} Tailoring proof systems to specific machine learning computations can reduce proof costs. This approach has been adopted by many existing schemes, wherein specialized zero-knowledge proof schemes are designed for specific computation types or model structures to improve efficiency. Future research can develop dedicated ZKP schemes for models with special properties, such as graph neural networks (handling graph structures), transformers (addressing stacked encoders/decoders and self-attention mechanisms), and recurrent neural networks (focusing on recurrent structures and time-series features). Additionally, exploring alternative zero-knowledge proof systems beyond QAP-based solutions, like Groth16~\cite{groth2016size}, could offer performance benefits in different computational scenarios.


2) \textbf{Specialized Hardware:} Using specialized hardware can alleviate the computational burden of generating ZKPs. Hardware accelerators, such as pipeline designs~\cite{zhang2021pipezk}, FPGAs for number-theoretic transforms~\cite{zhao2022hardware}, and GPUs for ZKP computation~\cite{ma2023gzkp,lu2023cuzk}, have shown potential in boosting efficiency in ZKP. \textcolor{black}{Integrating these acceleration strategies into existing ZKP-VML frameworks could create a comprehensive architecture for enhancing the operational efficiency of ZKP-VML systems, crucial for real-world deployment. }
    

3) \textbf{Balance with Privacy:} Achieving a balance between security \& Privacy with efficiency can help reduce overall system costs. In some scenarios, not all computations need to be verified, allowing for selective verification of a minimal subset of rounds in multi-round training processes. Techniques like range proofs can relax equality constraints in verification to improve efficiency. \textcolor{black}{Additionally, fine-tuning privacy protection requirements and avoiding unnecessary privacy measures can reduce computational overhead. Addressing these trade-offs between security and efficiency is key to the practical applicability of ZKP-VML. }


\subsection{Real-World Development}

Currently, ZKP-VML is still in development, with most implementations at the demo stage to showcase their performance. For practical deployment, it is crucial to ensure interoperability with existing solutions, allowing developers to seamlessly integrate privacy features without major modifications to their workflows. Key considerations include ease of integration, performance impact, and compatibility with various hardware and software environments. This will foster broader adoption and facilitate the real-world application of ZKPs across industries. 
\textcolor{black}{A key focus area is the development of privacy-preserving computing frameworks that support ZKP-VML, such as Rosetta~\cite{wagh2019securenn}. Currently, mainstream frameworks offer limited support for ML and ZKPs, often requiring extensive expertise to construct ZKP-VML systems. Making ZKP-VML more user-friendly and integrable into these frameworks will significantly expand its applicability. }

Additionally, standardization and collaboration are essential for advancing ZKP-VML. Academic, industrial, and open-source collaborations can drive the development of standardized protocols, efficient libraries, and benchmarking tools~\cite{modulus2023cost}. \textcolor{black}{For instance, benchmarking tools tailored to mainstream ZKP protocols can help quantify performance disparities, guiding users toward more efficient solutions and improving overall system efficiency. Such initiatives will foster knowledge sharing, accelerate innovation, and support the adoption of ZKP-VML as a reliable privacy-preserving tool in machine learning applications. }

\subsection{Novel Scenarios}

The evolving ML technologies and expanding ML scenarios offer increased opportunities for incorporating verifiability into ML through ZKPs.
Currently, the vast majority of research is limited to the traditional training and inference, which represent the most fundamental ML scenarios.
In DML, there are many scenarios where verifiability is worth exploring, such as proving model ownership and demonstrating model fairness. 
\textcolor{black}{
By identifying specific evaluation criteria and ML tasks in various scenarios, such as model watermark verification and model integrity verification, and converting them into forms that can be proved using ZKP, it becomes possible to extend ZKP-VML to new ML scenarios. }
ZKPs provide verifiability for these scenarios while simultaneously protecting data privacy.
The verifiability in these scenarios have not yet been fully investigated. 
Beyond these scenarios, there are numerous undeveloped ML scenarios that could benefit from verifiability, such as the deepfake detection, model interpretability and quantum machine learning.
Introducing verifiability for additional scenarios often involves new computational processes, such as calculating model watermarks or assessing model fairness. 
Since these processes and algorithms differ from standard training and inference, they present entirely new efficiency challenges for the proofs. 
Moreover, for more complex DML scenarios, relying solely on ZKP may not fully meet the security requirements. 
Therefore, it is also necessary to consider integrating other cryptographic techniques to enhance and enrich the security properties of ZKP-VML.

\subsection{Various Properties}

In future research, the complex combination with different cryptographic and techniques can be considered to enrich security properties of ZKP-VML.

By integrating cryptographic techniques related to data privacy protection, ZKP-VML can further enhance its capability to safeguard data privacy.
For instance, by combining differential privacy and homomorphic encryption techniques, ZKP-VML can achieve stronger privacy protection properties. 
Differential privacy can enhance the privacy protection level of data, preserving the data privacy during computations and interactions process beyond the proof generation and verification.
Through homomorphic encryption, verifiers can protect computation tasks by enabling provers to perform tasks using homomorphic computations on encrypted data.
This process enables the execution of computations without disclosing the plaintext content, thereby safeguarding the privacy of the verifier. 
Besides, by combining with blockchain, ZKP-VML can be utilized to build trustless and decentralized federated learning systems, which is a possible learning paradigm in the future. 
Blockchain and the smart contracts deployed on it ensure the decentralization of the system, while ZKP-VML constrains the training behaviors of participants without compromising the data privacy.

With the application of additional techniques, it becomes crucial to consider how newly introduced features might pose potential threats to the integrity or privacy of the original ZKP-VML scheme. 
As the system becomes more complex and incorporates additional functionalities, it is essential to conduct thorough security assessments to ensure that the integrity and privacy guarantees of the ZKP-VML scheme remain robust and uncompromised in a dynamic and evolving environment. 
This includes evaluating the impact of new features on data privacy, ensuring data integrity, and identifying potential vulnerabilities that could be exploited by malicious actors. 

\subsection{\textcolor{black}{Communication Network}}

\textcolor{black}{
The ZKP-VML approach faces several open issues in communication networks, especially regarding scalability, and integration with wireless technologies. These issues include:
}

\textcolor{black}{
1) Scalability: As machine learning models and data grow, the computational and communication overhead of zero-knowledge proofs can become a bottleneck. 
Besides, ZKP-VML are still relatively slow for real-time applications.
This is particularly relevant in real-time communication scenarios, such as 5G/6G networks where large amounts of data are transmitted.
Future research can explore efficient communication paradigm and protocol to support complex proof generation and verification. 
Hybrid cryptographic techniques, such as combining homomorphic encryption with ZKPs, could mitigate some of the performance bottlenecks in privacy-preserving computations.
}

\textcolor{black}{
2) Integration with communication scenarios: There is a growing interest in incorporating machine learning into communication networks, such as using ML for signal detection, channel estimation, and network optimization. 
ZKP-VML could help ensure the privacy and integrity of ML computations in these contexts. 
However, integrating ZKPs with existing wireless communication protocols remains a challenge, particularly in ensuring compatibility with both centralized and decentralized systems. 
Exploring dedicated proof systems for specific wireless communication tasks and optimizing proof schemes for real-time operation could address both latency and scalability concerns. 
At the same time, communication networks introduce new application scenarios and ML tasks for ZKP-VML, such as semantic communication. 
On one hand, semantic communication introduces new verifiable ML tasks for ZKP-VML, including novel encoding-decoding processes and semantic data verification. 
On the other hand, semantic communication may also contribute to enhancing the communication and computational efficiency of ZKP-VML. 
These are promising directions for further research.
}

\section{Conclusion\label{con}}
This paper provides a comprehensive review of zero-knowledge proof-based verifiable machine learning (ZKP-VML). 
\textcolor{black}{
Firstly, we provide an introduction to DML, ZKPs, and communication networks, emphasizing how ZKPs can address the security challenges inherent in DML. 
We then formally define ZKP-VML, including its underlying algorithms and key properties, while also discussing associated challenges and applications.
Next, we offer a comprehensive overview of the existing schemes, detailing the research timeline in this area and how the properties of ZKP-VML are realized across different approaches. 
For each scheme, we analyze the technical route adopted, presenting a well-structured categorization. Furthermore, we examine the optimization method utilized within these schemes, comparing their performance in terms of computational and communication complexity. 
}
Finally, we discuss the challenges and future research directions in ZKP-VML. 
We believe that this work will inspire further research into the field of ZKP-VML.


%



\section*{Acknowledgment}

This work is supported by National Key Research and Development Program of China under the grant 2023YFF0905302, National Natural Science Foundation of China (NSFC) under the grant No. 62172040, No. 62372020, No. 72031001, and ``Pioneer" and ``Leading Goose" R\&D Program of Zhejiang (No. 2024C01073). 

The first author is supported by China Scholarship Council (CSC) under the grant No. 202206030042. 

\ifCLASSOPTIONcaptionsoff
  \newpage
\fi



%



\bibliographystyle{IEEEtran}
\bibliography{IEEEabrv,IEEEbib}

%








\end{document}